\documentclass{article}

\usepackage{arxiv}

\usepackage[utf8]{inputenc} 
\usepackage[T1]{fontenc}    
\usepackage{booktabs}       
\usepackage{microtype}      
\usepackage{lipsum}	    
\usepackage{graphicx}
\usepackage{doi}

\usepackage[]{subcaption}
\usepackage{amsmath,amsthm,amssymb,amsfonts,bm}
\usepackage{mathtools}
\usepackage{thmtools,thm-restate}
\usepackage{nicefrac}
\usepackage{centernot}
\usepackage{multirow}
\usepackage{color}
\usepackage[usenames,dvipsnames,svgnames,table]{xcolor}
\usepackage{hyperref}
\hypersetup{%
  colorlinks=true,
  urlcolor=Mahogany, linkcolor=WildStrawberry, citecolor=RoyalBlue,
  pdftitle={Disentangled Representations for Causal Cognition},
  pdfsubject={cs.AI, cs.LG, q-bio.NC},
  pdfauthor={Filippo Torresan, Manuel Baltieri},
  pdfkeywords={causal cognition, causal reinforcement learning, disentangled representations, disentanglement},
}
\usepackage{url}
\usepackage{cleveref}
\usepackage{quiver}
\usepackage{tikzit}

\tikzstyle{rotate +90}=[fill=none, draw=none, shape=circle, rotate=90]
\tikzstyle{rotate -90}=[fill=none, draw=none, shape=circle, rotate=-90]
\tikzstyle{node factors}=[fill=none, draw=orange, shape=circle, inner sep=1, minimum size=1cm]
\tikzstyle{node codes}=[fill=none, draw=teal, shape=circle, inner sep=1, minimum size=1cm]
\tikzstyle{node conf}=[fill=none, draw=blue, shape=circle, inner sep=1, minimum size=1cm]
\tikzstyle{node obs}=[fill=lightgray, draw=black, shape=circle, inner sep=1, minimum size=1cm]
\tikzstyle{function clear}=[fill=white, draw=dgreen, shape=circle, minimum size=6 pt, inner sep=2 pt]
\tikzstyle{function black}=[fill=black, draw=black, shape=circle, inner sep=1, minimum size=2mm]
\tikzstyle{function red}=[fill=red, draw=red, shape=circle, inner sep=1, minimum size=2mm]
\tikzstyle{function blue}=[fill=blue, draw=blue, shape=circle, inner sep=1, minimum size=2mm]
\tikzstyle{function magenta}=[fill=magenta, draw=magenta, shape=circle, inner sep=1, minimum size=2mm]
\tikzstyle{function teal}=[fill=teal, draw=teal, shape=circle, inner sep=1, minimum size=2mm]
\tikzstyle{function brown}=[fill=brown, draw=brown, shape=circle, inner sep=1, minimum size=2mm]
\tikzstyle{function orange}=[fill=orange, draw=orange, shape=circle, inner sep=1, minimum size=2mm]
\tikzstyle{new style 0}=[fill={rgb,255: red,76; green,169; blue,170}, draw={rgb,255: red,76; green,169; blue,170}, shape=circle, inner sep=1]
\tikzstyle{box-.5-.5}=[fill=none, draw=black, shape=rectangle, minimum width=5mm, minimum height=5mm, thick]
\tikzstyle{box-1-.5}=[fill=none, draw=black, shape=rectangle, minimum width=1cm, minimum height=5mm, thick]
\tikzstyle{fbox-1-.5-blue}=[fill={blue!10}, draw=black, shape=rectangle, minimum width=1cm, minimum height=5mm, thick]
\tikzstyle{box-1-.5-blue}=[fill=none, draw=blue, shape=rectangle, minimum width=1cm, minimum height=5mm, thick]
\tikzstyle{fbox-1-.5-red}=[fill={red!10}, draw=black, shape=rectangle, minimum width=1cm, minimum height=5mm, thick]
\tikzstyle{box-1-.5-red}=[fill=none, draw=red, shape=rectangle, minimum width=1cm, minimum height=5mm, thick]
\tikzstyle{box-1-1}=[fill=none, draw=black, shape=rectangle, minimum width=1cm, minimum height=1cm, thick]
\tikzstyle{box-1-1-red}=[fill=none, draw=red, shape=rectangle, minimum width=1cm, minimum height=1cm, thick]
\tikzstyle{box-1-2}=[fill=none, draw=black, shape=rectangle, minimum width=1cm, minimum height=2cm, thick]
\tikzstyle{box-1.5-.5}=[fill=none, draw=black, shape=rectangle, minimum width=1.5cm, minimum height=5mm, thick]
\tikzstyle{box-1.5-1}=[fill=none, draw=black, shape=rectangle, minimum width=1.5cm, minimum height=1cm, thick]
\tikzstyle{box-2-.5}=[fill=none, draw=black, shape=rectangle, minimum width=2cm, minimum height=.5cm, thick]
\tikzstyle{box-2-1}=[fill=none, draw=black, shape=rectangle, minimum width=2cm, minimum height=1cm, thick]
\tikzstyle{box-3-.5}=[fill=none, draw=black, shape=rectangle, minimum width=3cm, minimum height=.5cm, thick]
\tikzstyle{box-3-2}=[fill=none, draw=black, shape=rectangle, minimum width=3cm, minimum height=2cm, thick]
\tikzstyle{box-3-4}=[fill=none, draw=black, shape=rectangle, minimum width=3cm, minimum height=4cm, rounded corners, thick]
\tikzstyle{box-4-.5}=[fill=none, draw=black, shape=rectangle, minimum width=4cm, minimum height=.5cm, thick]
\tikzstyle{box-4-3}=[fill=none, draw=black, shape=rectangle, minimum width=4cm, minimum height=3cm, thick]
\tikzstyle{dashed-box-4-3}=[fill=none, draw=black, shape=rectangle, minimum width=4cm, minimum height=3cm, thick, rounded corners, dashed]
\tikzstyle{box-4-5}=[fill=none, draw=black, shape=rectangle, minimum width=4cm, minimum height=5cm, thick]
\tikzstyle{box-5-2}=[fill=none, draw=black, shape=rectangle, minimum width=5cm, minimum height=2cm, thick]
\tikzstyle{box-5-3}=[fill=none, draw=black, shape=rectangle, minimum width=5cm, minimum height=3cm, thick]
\tikzstyle{box-5-4}=[fill=none, draw=black, shape=rectangle, minimum width=5cm, minimum height=4cm, thick]
\tikzstyle{box-6-3}=[fill=none, draw=black, shape=rectangle, minimum width=6cm, minimum height=3cm, thick]
\tikzstyle{box-6-4}=[fill=none, draw=black, shape=rectangle, minimum width=6cm, minimum height=4cm, thick]
\tikzstyle{box-6-5}=[fill=none, draw=black, shape=rectangle, minimum width=6cm, minimum height=5cm, thick]
\tikzstyle{box-7-5}=[fill=none, draw=black, shape=rectangle, minimum width=7cm, minimum height=5cm, thick]
\tikzstyle{box-8-4}=[fill=none, draw=black, shape=rectangle, minimum width=8cm, minimum height=4cm, thick]
\tikzstyle{box-10-6}=[fill=none, draw=black, shape=rectangle, minimum width=10cm, minimum height=6cm, thick]
\tikzstyle{box-13-6}=[fill=none, draw=black, shape=rectangle, minimum width=13cm, minimum height=6cm, thick]
\tikzstyle{ellipse-23-14-red}=[fill=none, draw=red, shape=ellipse, minimum width=23cm, minimum height=14cm, thick]
\tikzstyle{ellipse-17-9-orange}=[fill=none, draw=Orange, shape=ellipse, minimum width=17cm, minimum height=9cm, thick]
\tikzstyle{ellipse-9-5-green}=[fill=none, draw=ForestGreen, shape=ellipse, minimum width=9cm, minimum height=5cm, thick]
\tikzstyle{ellipse-9-17}=[fill=none, draw=black, shape=ellipse, minimum width=9cm, minimum height=17cm, thick, dashed]
\tikzstyle{ellipse-5-9}=[fill=none, draw=black, shape=ellipse, minimum width=5cm, minimum height=9cm, thick, dashed]
\tikzstyle{fbox-.5-.5-red}=[fill={red}, draw=black, shape=rectangle, minimum width=.5cm, minimum height=.5cm, thick]
\tikzstyle{fbox-.5-.5-orange}=[fill={Orange}, draw=black, shape=rectangle, minimum width=.5cm, minimum height=.5cm, thick]
\tikzstyle{fbox-.5-.5-green}=[fill={ForestGreen}, draw=black, shape=rectangle, minimum width=.5cm, minimum height=.5cm, thick]
\tikzstyle{big-circle-red}=[fill=none, draw=red, shape=circle, inner sep=1, minimum size=10cm, thick]
\tikzstyle{big-circle-orange}=[fill=none, draw=Orange, shape=circle, inner sep=1, minimum size=10cm, thick]
\tikzstyle{big-circle-green}=[fill=none, draw=ForestGreen, shape=circle, inner sep=1, minimum size=10cm, thick]

\tikzstyle{object red}=[-, draw={rgb,255: red,191; green,0; blue,64}, thick]
\tikzstyle{object blue}=[-, draw=blue, thick]
\tikzstyle{object orange}=[-, draw={rgb,255: red,255; green,128; blue,0}, thick]
\tikzstyle{big dashes}=[-, dash pattern=on 8mm off 2mm, thick, dashed]
\tikzstyle{arrow}=[->]
\tikzstyle{thick-line}=[-, thick]
\tikzstyle{thick-blue-line}=[-, thick, draw=blue]
\tikzstyle{thick-red-line}=[-, thick, draw=red]
\tikzstyle{thick-orange-line}=[-, thick, draw=orange]
\tikzstyle{thick-teal-line}=[-, thick, draw=teal]
\tikzstyle{thick-double-line}=[-, thick, double]
\tikzstyle{thick-arrow-black}=[->, thick]
\tikzstyle{thick-arrow-orange}=[->, thick, draw=orange]
\tikzstyle{thick-arrow-teal}=[->, thick, draw=teal]
\tikzstyle{thick-arrow-blue}=[->, thick, draw=blue]
\tikzstyle{thick-arrow-violet}=[->, thick, draw=violet]
\tikzstyle{thick-arrow-green}=[->, thick, draw=green]
\tikzstyle{ultrathick-arrow}=[-, line width=6]
\tikzstyle{thick-blue-arrow}=[->, thick, draw=blue]
\tikzstyle{thick-blue-dashed-arrow}=[->, thick, draw=blue, dashed]
\tikzstyle{thick-red-arrow}=[->, thick, draw=red]
\tikzstyle{tube-blue}=[-, fill=none, draw={rgb,255: red,103; green,161; blue,255}, line width=6]
\tikzstyle{tube-green}=[-, draw={rgb,255: red,110; green,207; blue,108}, line width=6, fill=none]
\tikzstyle{ultrathick-arrow}=[-{Latex}, line width=6]
\usepackage[backref, uniquelist=false, uniquename=false, maxnames=1, style=numeric-comp, language=american, backend=biber]{biblatex}%
\def\eqref#1{equation~\ref{#1}}

\def\1{\bm{1}}

\DeclareMathAlphabet{\mathsfit}{\encodingdefault}{\sfdefault}{m}{sl}
\SetMathAlphabet{\mathsfit}{bold}{\encodingdefault}{\sfdefault}{bx}{n}

\newcommand{\R}{\mathbb{R}}

\theoremstyle{definition} 
\newtheorem{definition}{Definition}[section]
\theoremstyle{plain}

\theoremstyle{plain}

\theoremstyle{plain}

\newcommand{\CueOne}{\mathbf{C}_1}
\newcommand{\CueTwo}{\mathbf{C}_2}
\newcommand{\Outcome}{\mathbf{+}}

\newcommand{\N}{\mathbb{N}}
\newcommand{\id}{\mathsf{id}}

\newcommand{\factors}{\mathbf{s}}
\newcommand{\Factors}{S}
\newcommand{\obs}{\mathbf{x}}
\newcommand{\Obs}{X}
\newcommand{\codes}{\mathbf{z}}
\newcommand{\Codes}{Z}

\newcommand{\nfactors}{n}
\newcommand{\nobs}{d}
\newcommand{\ncodes}{l}

\newcommand{\noise}{N}
\newcommand{\scm}{\mathcal{C}}
\newcommand{\assignment}{h}
\newcommand{\emission}{g}
\newcommand{\PA}{\mathbf{PA}}

\newcommand{\Confounders}{C}
\newcommand{\nconfounders}{m}

\newcommand{\Data}{\mathcal{D}}
\newcommand{\encoderweights}{\boldsymbol{\phi}}
\newcommand{\decoderweights}{\boldsymbol{\theta}}
\newcommand{\ELBO}{\mathcal{L}}

\newcommand{\prob}[1]{P({#1})}
\newcommand{\states}{\factors}
\newcommand{\States}{\Factors}
\newcommand{\actions}{\mathbf{a}}
\newcommand{\Actions}{A}
\newcommand{\reward}{r}
\newcommand{\Reward}{R}

\newcommand{\Transitions}{T}
\newcommand{\Observations}{M}
\newcommand{\stateaction}{\tau}

\newcommand{\policy}{\pi}
\newcommand{\policyparameters}{{\boldsymbol{\omega}}}
\newcommand{\Advantage}{\text{Adv}}
\newcommand{\return}{G}
\newcommand{\expreturn}{J}
\newcommand{\valuefunction}{V}
\newcommand{\beliefs}{B}

\newcommand{\behavioural}{{\boldsymbol{\beta}}}
\newcommand{\importanceweights}{{\boldsymbol{w}_{T}^{i}}}
\newcommand{\discount}{\gamma}
\newcommand{\estexpreturn}{\hat{\expreturn}}
\newcommand{\Qfunction}{Q}
\newcommand{\ntrajectories}{N}
\newcommand{\ntime}{T}
\newcommand{\itertrajectories}{n}
\newcommand{\itertime}{t}
\newcommand{\iterpartial}{i}

\newcommand{\generating}{\emission}
\newcommand{\encoding}{f}
\newcommand{\decoding}{k}
\newcommand{\modularity}{m}
\newcommand{\informativeness}{i}
 
\newcommand{\bisim}{B}

\title{Disentangled Representations for Causal Cognition}

\author{\href{https://orcid.org/0000-0001-8790-2565}{\includegraphics[scale=0.06]{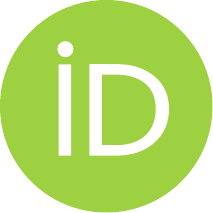}\hspace{1mm}Filippo Torresan} \\
	School of Engineering and Informatics \\
	University of Sussex \\
	Falmer, Brighton, BN1 9RH \\
	\texttt{f.torresan@sussex.ac.uk} \\
	\And
	\href{https://orcid.org/0000-0002-6086-4711}{\includegraphics[scale=0.06]{orcid.pdf}\hspace{1mm}Manuel Baltieri}\thanks{Corresponding author.} \\
	Araya Inc., Tokyo, Japan \\
	and \\
        School of Engineering and Informatics \\
	University of Sussex \\
	Falmer, Brighton, BN1 9RH \\
	\texttt{manuel\_baltieri@araya.org} \\
}

\renewcommand{\shorttitle}{Disentangled Representations for Causal Cognition}

\begin{document}
\maketitle

\begin{abstract}
  Complex adaptive agents consistently achieve their goals by solving problems that seem to require an understanding of causal information, information pertaining to the causal relationships that exist among elements of combined agent-environment systems. Causal cognition studies and describes the main characteristics of causal learning and reasoning in human and non-human animals, offering a conceptual framework to discuss cognitive performances based on the level of apparent causal understanding of a task. Despite the use of formal intervention-based models of causality, including causal Bayesian networks, psychological and behavioural research on causal cognition does not yet offer a computational account that operationalises how agents acquire a causal understanding of the world. Machine and reinforcement learning research on causality, especially involving disentanglement as a candidate process to build causal representations, represent on the one hand a concrete attempt at designing causal artificial agents that can shed light on the inner workings of natural causal cognition. In this work, we connect these two areas of research to build a unifying framework for causal cognition that will offer a computational perspective on studies of animal cognition, and provide insights in the development of new algorithms for causal reinforcement learning in AI.
\end{abstract}

\keywords{causal cognition \and causal reinforcement learning \and disentangled representations \and disentanglement}

\section{Introduction}
\label{sec:intro}

Causal cognition, the ability to acquire and exploit causal information about oneself and the world, is a core aspect of adaptive and intelligent behaviour, in non-human and human animals \parencite{Gopnik2007, McCormack2011, Sloman2015, Goddu2024}.
At the same time, in recent years it has become apparent that artificial systems displaying various forms of seemingly intelligence behaviour still fall short of performing at the level of the majority of non-human animals that showcase various kinds of causal cognitive abilities \parencite{Lake2017, Pearl2018c, Crosby2020a, Shevlin2019}.
It has thus been suggested that an understanding of the mechanisms of causal cognition will play a crucial role in cognitive science and artificial intelligence for the next decades \parencite{Scholkopf2015, Lake2017, Pearl2018c, Shevlin2019, Levine2021, LeCun2022, Goyal2022, Gupta2024}.

The study of causal cognition both in human and non-human animals has a long history, with roots in behavioural studies trying to establish the extent to which an organism's behaviour reflects proper causal understanding of the world instead of simpler forms of associative learning \parencite{Sperber1995, Gopnik2007, Gopnik2004, Tenenbaum2002, Griffiths2005, Sloman2005a, Penn2007, Sloman2015}.
Some of the most influential studies in this area have combined theoretical and modelling work based on the formalism of causal Bayesian networks to account for the cognitive performance of various subjects in tasks designed for testing causal cognition abilities \parencite{Gopnik2000, Gopnik2001, Gopnik2004, Tenenbaum2002, Griffiths2009, Griffiths2005}.

Unlike research on learning from reinforcement in behavioural psychology, which played a pivotal role in the emergence of a mature and rich formal characterisation of modern reinforcement learning \cite{Sutton2018}, the wealth of experimental findings in causal learning has not yet been translated into a set of computational principles for a coherent, pragmatic framework showing how agents \emph{acquire} a causal perspective of the world by acting in it.
In other words, most approaches set aside the core question about causal cognition: how are causal models acquired, or \emph{learnt}, by agents in the first place?
Mainstream Bayesian frameworks overlook this question because they tend to describe processes of causal inference \emph{with a model} \parencite{Bruineberg2022}, i.e. assuming that the cognitive capacities of agents in a certain context can be assessed using a model encoding causal variables and relationships postulated by the scientist.
This descriptive approach however fails to provide a clear (and testable) account of how a causal viewpoint can emerge from an agent’s first-person experience, i.e. \emph{within a model} \parencite{Bruineberg2022}.

On the other hand, the agent-perspective focusing on computational models that implement processes of causal learning and reasoning is emerging as a dominant area of research in the field of artificial intelligence (AI), following a surge of interest in causality by the machine learning and deep learning communities \parencite{Scholkopf2022, Scholkopf2021, Peters2017}.
In these areas, several works have proposed new unsupervised methods to learn causal structure from data (causal structure learning or causal discovery) \parencite{Peters2016, Annadani2021, Faria2022, Lorch2022, Lowe2022, Ke2020}, while others have designed new reinforcement learning algorithms based on some of the principles and ideas developed in causality research \parencite{Sontakke2021a, Seitzer2021, Rezende2020, Huang2022a, Zholus2022, Li2020, Lei2022, Goyal2020, Javed2020}.
Others have also recognised the importance of causality for robotics \parencite{Brawer2021, Hellstrom2021} and new benchmarks and datasets have been introduced to study more rigorously causal inference in AI \parencite{Ahmed2020, Weichwald2022, Liu2023, Beyret2019, Crosby2019, Crosby2020a}.

In this work, we introduce a unifying treatment of research in causal cognition in non-human animals and causal machine learning, with the goal of laying the foundations for a research program where 1) computational work in causal machine learning can help us to gain clarity on the principles driving the acquisition of causal knowledge in adaptive agents, and where 2) causal cognition can in turn inspire the development of new algorithmic implementations in causal machine learning.
In particular, we will provide an explicit blueprint for a theoretical and computational framework centred around the notions of disentanglement and causal representation learning \parencite{Scholkopf2021, Nalmpantis2023, Talon2024}, similar to the process of functional specialisation in neural cells \parencite{Whittington2022, Whittington2023}, that can form a more rigorous basis for conceptual characterisations of causal cognition in terms of explicitness, sources and integration of causal information \parencite{Starzak2021, Woodward2021, Woodward2012, Woodward2011, Woodward2007, Goddu2024}.
Our proposal will bridge the gap between studies of causal cognition, on one side, and mathematical and computational models of adaptive behaviour, on the other, allowing hypotheses about the nature and emergence of different constitutive blocks of causal learning and reasoning to be tested using the power of modern causal machine learning models.

In \cref{sec:what-causal-cog} we will start with an overview of the literature on causal cognition with a heavy focus on non-human animals, briefly going through some of its most relevant paradigm shifts leading to the modern proposals addressing causal understanding. \Cref{sec:towards-frame-ccog} will go through a conceptual framework proposed to unify different accounts of causal understanding and developed in order to characterise causal understanding in terms of causal information on a more finely grained scale that includes three dimensions: explicitness, sources and integration of causal information. Using some of the formal work presented in \cref{sec:math-frame}, collecting the useful notions of disentanglement, structural causal models and (partially observable) Markov decision processes, in \cref{sec:computational} we will see how it is possible to operationalise computationally causal understanding in modern work on deep (reinforcement) learning. \Cref{sec:bridging} will provide a comparative analysis of work on natural (animal cognition) and artificial agents (machine/reinforcement learning) showing common areas of interest between causal cognition and causal machine learning and highlighting the main differences between the two. Finally, in \cref{sec:discussion} we discuss how these two lines of research could benefit from each other's insights moving forward, speculating on proposals that combine them in new ways.

\section{Causal cognition}
\label{sec:what-causal-cog}

\subsection{Causal cues and the debate on associative vs.\ cognitive explanations}
\label{subsec:causal-cues-debate}

Early work on causal cognition in human and non-human animals focused on how subjects learn about the strength of cue-reward relationships, where some of the given cues could be attributed the ``causal power'' of eliciting rewarding outcomes \parencite{Sperber1995, Kelley1973, Cheng1997}.
A significant part of this early work can be contextualised within an old debate on whether causal learning is just a form of associative or contingency learning, the dominant theoretical framework to study animal learning \parencite{Rescorla1972}, or a form of learning that requires more cognition-laden processes.

The crux of this debate was not whether there are causal relationships or structures in the world, which is more a philosophical type of issue.
Granted that there are, associative accounts have usually tried to show that the successful performance of some agents in seemingly causal learning tasks can be modelled, and ultimately explained, by means of purely associative learning mechanisms.
Roughly, these would track the relevant causal associations between certain variables without the need of invoking more sophisticated cognitive processes or structures involving a notion of causality \parencite{Shanks1987, Shanks2007, Dickinson2001a, Dickinson2001}.
In contrast, other works have highlighted how certain behavioural responses, especially from humans, are indicative of \textbf{causal models} that the agent in question would exploit to reason about causal relationships of various sorts \parencite{Waldmann1990, Waldmann1992, Waldmann1995, Blaisdell2006}.

\begin{figure}
    \centering
    \includegraphics[width=\textwidth]{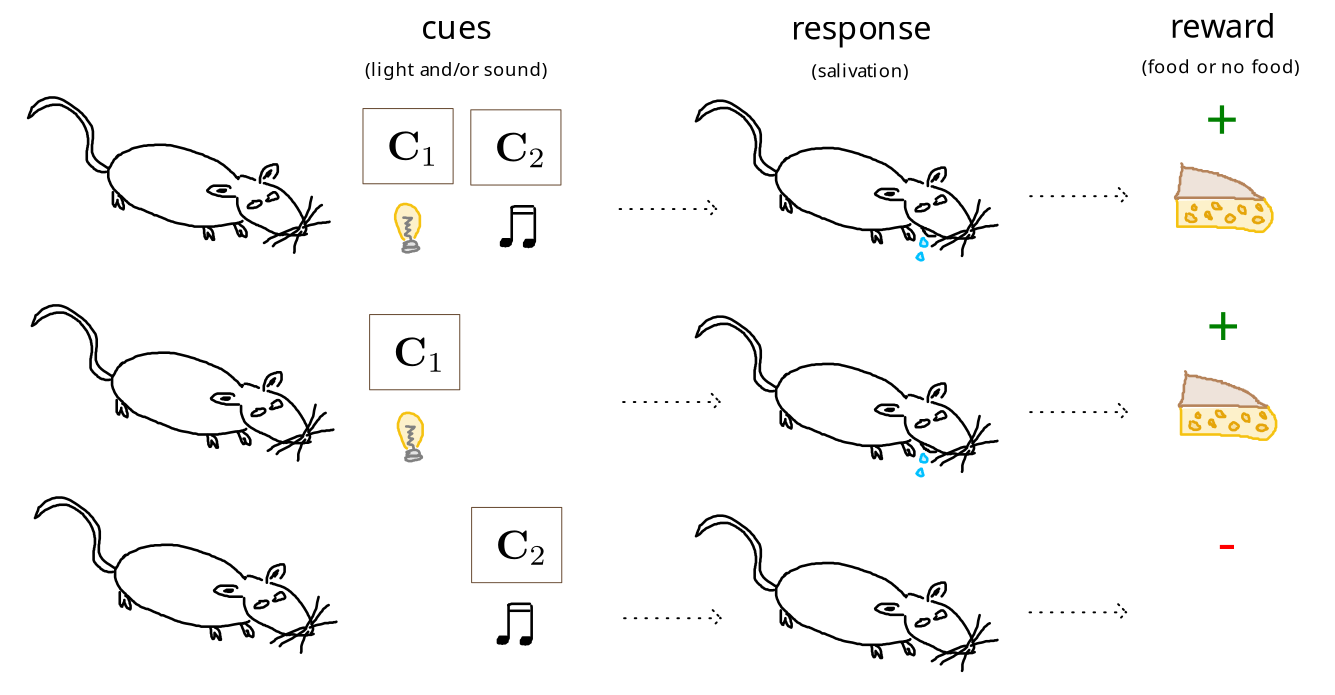}
    \caption{\textbf{An illustration of the phenomenon of backward blocking in rats.}. Subjects are conditioned to elicit a response (salivation) to a stimulus (presence of food) by means of a compound cue (light + acoustic tone) as well as a single cue (light). When tested with the other cue (acoustic tone), the rats do not react as strongly as if they understood that in the compound-cue trials the only cause of the reward was the first cue (light).}
    \label{fig:backward-blocking}
\end{figure}

A paradigmatic example of how this debate has typically unfolded can be found within the analysis of \emph{backward blocking} in conditioning experiments with rats.
In these experiments, test subjects are exposed to cues, say \(\CueOne\), \(\CueTwo\), and the compound cue given by their combination, i.e. \(\CueOne \CueTwo\), that may or may not lead to a rewarding outcome, indicated by \(\Outcome\) (see \cref{fig:backward-blocking}).
After some trials, the subjects, often rats, learn about the relationships between those cues and the reward, and react accordingly when similar cues are shown.
In the backward blocking scenario, after rats have been trained with cue-outcome sequences like \(\CueOne \CueTwo \Outcome\), \(\CueOne \Outcome\) (in that specific order), one finds that they don't react to the presentation of \(\CueTwo\) alone in subsequent trials.
The response to \(\CueTwo\) has been ``blocked'' by the animal upon witnessing \(\CueOne \Outcome\), i.e. the causal role of \(\CueTwo\) is reconsidered after understanding that it is not involved in producing the reward.

This sort of retrospective evaluation (of what happened in a earlier trial) is a problem for associative accounts because they usually assume that a cue-outcome association can increase or decrease in strength only when the cue is present (together or without the reward).
However, in backward blocking scenarios the change in behavioural response to \(\CueTwo\) occurs following the presentation of \(\CueOne \Outcome\) and despite the fact that \(\CueTwo\) has always (or most of the time) appeared at the same time as the reward \(\Outcome\).
An advocate for causal models would see retrospective evaluation as an example of their influence on cue-outcome learning.
Given the evidence represented by \(\CueOne \Outcome\), the decreased response towards \(\CueTwo\) could be explained in terms of a re-evaluation of the role played by that cue when it appeared as part of the sequence \(\CueOne \CueTwo \Outcome\).
Such evidence would in turn suggest that \(\CueTwo\) was not included as part of a causal relationship with the rewarding outcome.

At the same time, over the years several revisions of traditional associative accounts to model retrospective evaluation have been proposed to account for backward blocking and more complex scenarios involving higher-order relations between cues, i.e. relations between two cues that never occur together but that appear in combination with another cue, see for instance \textcite{Dickinson2001a}.
However, these revisions usually depart from traditional associative principles in significant ways, e.g. requiring more sophisticated information-processing capabilities, see for instance the discussion in \textcite{Penn2007} and references therein. This thus leaves us with architectures based on, or inspired by associative principles \parencite{Dickinson2012}, begging however the question of whether these models are still associative, or not.

More recently, the whole dichotomy between associative and cognitive explanations has been put into discussion, as different works argue that empirical and theoretical grounds for such a distinction are too weak \parencite{Penn2007, Buckner2011, Heyes2012, Hanus2016}.
Indeed, one could point out that the whole debate is a consequence of a narrow view of what counts as cognitive, e.g. exclusively representations and processes that have an ``internal semantic or propositional structure'' and that shape thought and behaviour via ``structured inference'' \parencite{Shanks2007}, hence a legacy of cognitivism.
It is this narrow and demanding view of what defines a cognitive representation or process that generate an unhelpful contrast with simpler and more basic ones in the explanation of mental faculties.

Focusing on the friction between these two views make us blind of the possibility that cognition ought to be considered on a spectrum. 
To witness, more recent perspectives have often extended the realm of what counts as cognitive \parencite{Lyon2020, Baluska2016, Barandiaran2006} to more basic, low-level biological processes, or have described cognition, from perception to action and higher-order functions, as primarily a matter of Bayesian inference on different spatiotemporal scales \parencite{Clark2013b, Hohwy2013, Chater2010}.
Pursuing this line of reasoning suggests then that the contrast between associative and cognitive accounts could be simply reframed as that between lower-level and higher-level cognitive processes placed on a common spectrum. 
In this view, the fundamental questions become where processes regarded as part of causal cognition can be found on the cognitive spectrum, whether they can be characterised in terms of simpler building blocks, and how they might interact with each other to realise some form of causal understanding.

\subsection{Causal understanding as a building block for causal cognition}
\label{subsec:causal-understanding}

Moving past the associative vs.\ cognitive debate using a more comprehensive definition of cognition at different levels has however brought forward a perhaps more fundamental disputes about the presence, or not, of forms of \textbf{causal understanding} in agents, and what such an understanding amounts to.
This is especially evident in the behavioural research on causal cognition in non-human animals, where the goal is to design behavioural tasks specifically intended to try and measure some manifestation of causal understanding \parencite{Sperber1995}.
In other words, tasks that would ascertain whether a subject is capable of the feats of causal cognition, where the assumption is that a solution of the task requires certain causal, cognitive strategies.

An example of this research is represented by studies on capuchin monkeys using the trap-tube task \parencite{Visalberghi1989, Visalberghi1993, Visalberghi1994, Visalberghi1996}, where causal cognition is characterised as the comprehension of key cause-effect relationships within the task.
The trap-tube task consists in pushing a food reward out from a transparent tube (anchored to the floor) using some kind of tool (e.g. a stick), by inserting it into one of the tube's two openings, see \cref{fig:trap-tube}.
In general, capuchin monkeys struggle to learn how to solve the task, either because they would pick the wrong kind of tool (a stick that could not be inserted into the tube because of its shape) or because they would pick the wrong side to put the stick, making the reward fall into a trap positioned underneath the tube.

The persistent error patterns of the (few) subjects that could solve the task after extensive trial and error are thought to be evidence of a distinction between 1) successful performance based on a ``stroke of luck'' after extensive active experimentation, and 2) successful performance based on an understanding of relevant causal variables inherent within the task requirements \parencite{Visalberghi1989, Visalberghi1994}.
It is in fact well known that capuchin monkeys have a propensity to produce a wide variety of actions and complex combinations thereof, even involving tools, to the point that they could be described as expert tool-users.
Because of this, it is unclear whether they have an appreciation of the causal relationships between their behaviour and the resulting outcome.
In other words, they might learn that using certain tools is an effective way to achieve certain results, but they may not appreciate the reasons for why their actions are successful \parencite{Visalberghi1994}.

In contrast, experimental evidence in chimpanzees suggests that they may have an understanding of the causal relationships between certain actions and their associated outcomes \parencite{Limongelli1995, Mulcahy2006, Seed2009}.
The key finding here is that some subjects, tested with different configurations of the trap-tube task, were able to select the right side of insertion (almost) immediately, allegedly displaying an ability to plan their actions according to the different causal relationships present in the task configuration.
Consequently, this evidence suggests that the successful subjects were not using heuristics such as a distance-based rule, which would for example determine the correct action based on the distance of the reward from the tube's openings without an understanding of the causal structure of the problem.
Instead, subjects appeared to take into account the causal features of the task configuration and choose beforehand what action to perform.
This would thus amount to a representational strategy that delineates the key requirements of the task in advance and results in the correct behaviour without the need of extensive trial-and-error learning.
More specifically, one could argue that those successful chimpanzees exhibited some kind of causal understanding of the consequences of pushing the stick inside the tube (but see \textcites{Martin-Ordas2008, Seed2009} for opposing views).
Unlike for instance the capuchin monkeys of other experiments \parencite{Visalberghi1989, Visalberghi1994}, where a constant monitoring of the effects of one's ongoing action (to check one is on the the right track) and attempting a variety of actions' combinations (in the hope to find the right one) was instead unnecessary.

In a subsequent review paper, \textcite{Visalberghi1998} argue that an organism can be said to understand causal relationships, or to posses a causal understanding of (parts of) the world, if and only if they are able to see or posit some mediating forces (or variables) between two associated events.
This kind of explanatory attitude is informally described as the key component of causal understanding, one that helps an organism to envisage and navigate the \textbf{web of causal possibilities} of the ``how'' and ``why'' events at different points in time may be connected.
This kind of understanding has an impact on the behavioural strategies an agent might pick to reach certain goals, as novel ways of manipulating the environment are disclosed, targeting those specific mediating forces \parencite{Visalberghi1998}. In this view, an organism equipped with causal knowledge is therefore capable of dealing with unexpected signals from the environment in ways that take into account different possibilities in a farseeing way \parencite{Visalberghi1998}.
Following this conceptual definition, it is thus worth focusing on the relation between causal understanding and behavioural strategies, and more specifically on whether agents can plan ahead how to obtain action-driven outcomes arising from a mastery of the causal structure of the world as opposed to a more basic perception-based understanding of only the causal structure governing one's observations (cf. the distinction between model-based and model-free reinforcement in \cref{subsub:weakDisentanglement}).

\subsection{Causal interventions and tool use}
\label{subsec:causal-interventions}

A strong candidate for the presence of plans based on action-driven outcomes is
the ability to produce a \textbf{causal intervention}, an action that involves a causal control on a particular effect \parencite{Visalberghi1998}.
On the one hand, this seems to provide strong evidence for causal cognition since producing an intervention requires some form of causal understanding. 
In particular, it requires an agent to understand that its actions, in the form of movements of its own body, could be used as external probes for the causal texture of the world (\emph{cf}. second rung of the \emph{causality ladder} in \textcite{Pearl2018c}).

On the other hand, the attribution of causal interventions to cognitive agents appears still controversial because there are only limited reports that hint at intervention-like abilities in, for instance, rats \parencite{Blaisdell2006, Leising2008} and primates \parencite{Volter2016, Goddu2024}.
At the same time it is unclear whether \textbf{tool-use}, the ability to skilfully manipulate objects, common in species like corvids \parencite{Taylor2014, Jacobs2015}, should count as a form of intervention or not.
In general, it is not entirely obvious what the markers of causal interventions are and how to design experiments that could determine their presence or absence.

Work on rats, for example, suggests that these animals can learn a common-cause model, where a light being turned on is perceived as the cause of two effects, a noisy sound and the release of some food.
After exposure to patterns of causal relationships for a certain number of (training) trials, the rats enter a test condition characterised by a lever that produces a noise when pressed.
Interestingly, it has been reported that after (accidental) lever presses, rats exhibit a less resolute search for food (measured by the number of nose-poking in the cage's hopper) than when the noise is presented alone.
A possible explanation for this behavioural response would regard these rats as capable of recognising their action (the level press) as an intervention, an independent self-generated perturbation on one variable of the learnt causal model.
In fact, an effect (noisy sound) cannot be an indication that a cause is present (light) when that effect is produced by an intervention (lever press).
Therefore, by conceiving of their action as an intervention on one variable of the learnt causal model, the rats do not expect that the other effect (food release) will occur, which then induces a less vigorous search for food \parencite{Blaisdell2006}.
While these findings are consistent with the claim that rats can differentiate between predictions based on observations and predictions based on interventions, they do not exhaustively prove that rats can produce interventions to activate a certain causal path, in this case the one leading from the light to the food dispensation (as discussed by \textcite{Blaisdell2006} themselves).

\begin{figure}
    \centering
    \begin{subfigure}[t]{0.45\textwidth}
        \centering
        \includegraphics[width=\textwidth]{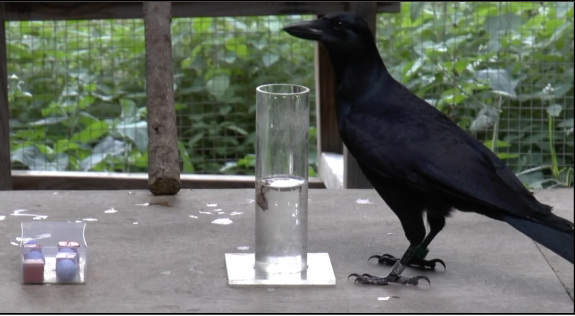}
        \caption{\textbf{Floating-reward task}. An animal subject learns to drop stones into a water-filled cylinder to raise the water level and reach a piece of food. Image credit: \cite{Miller2016}, adapted under the terms of the \href{https://creativecommons.org/licenses/by/4.0/}{Creative Commons Attribution License}.}
        \label{fig:floating-reward-crp}
    \end{subfigure}
    \hspace{2em}
    \begin{subfigure}[t]{0.45\textwidth}
        \centering
        \includegraphics[width=\textwidth]{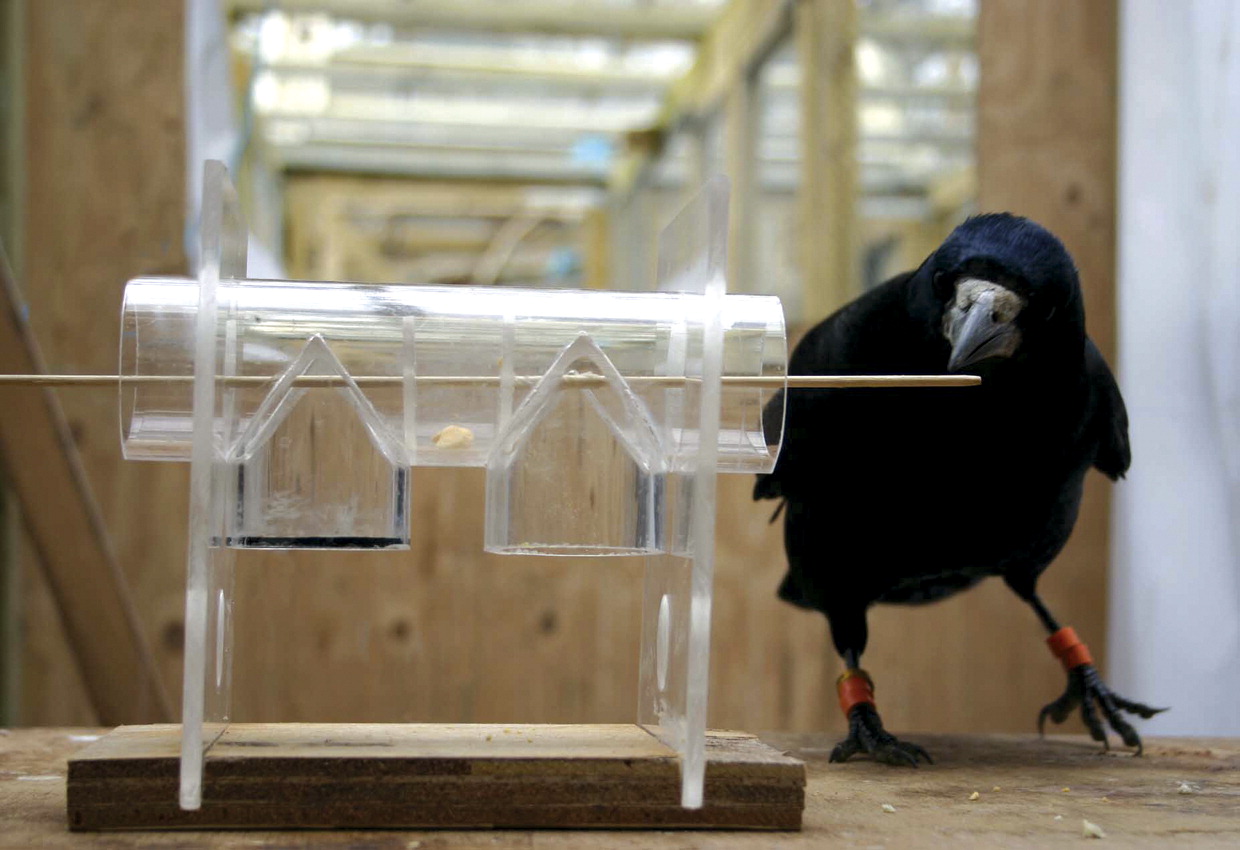}
        \caption{\textbf{A variation of the trap-tube task}. An animal subject learns to retrieve a reward from a transparent tube by pushing it out by means of a stick. In this variation of the task, the crow has to slide the reward over the open hole (the one on the right). Image credit: \parencite{Chappell2006} under \href{https://www.elsevier.com/about/policies-and-standards/open-access-licenses/elsevier-user}{Elsevier user license}.}  
        \label{fig:trap-tube}
    \end{subfigure}
    \caption{\textbf{Causal cognition tasks}.}
    \label{fig:animal-tasks}
\end{figure}

Work on corvids on the other hand, see for instance \textcite{Taylor2009a}, testing New Caledonian crows with a few variations of the trap-tube task (see \cref{fig:trap-tube}), suggests that they possess critical causal understanding abilities, e.g. an appreciation of causal relationships involving object-hole interaction, on which their exceptional tool-using skills might be built.
Similarly, \textcites{Jelbert2014, Logan2014a} present experiments on a task (inspired by Aesop's fables) in which crows have to learn to drop some objects (e.g. stones) into the right water-filled tube so that the water displacement brings a floating reward (e.g. a piece of meat) closer to the tube opening (see \cref{fig:floating-reward-crp}).
The results here point at the fact that the birds managed to solve the task, seemingly by attending to the relevant causal information, e.g. the fact that larger and not hollow objects will produce a bigger water displacement.
For a variation of the task however, where the setup instead consisted of three water-filled tubes arranged in a row on a wooden board, with some space between each other, results were less clear.
In this task, the baited tube is the one in the centre and, crucially, it is connected with one of the others by means of a U-shape tube hidden from view (located underneath the board).
Dropping objects into one of the lateral tubes will have as an effect a water-level rise in the baited tube.
Since the central tube is too narrow to drop anything in it, to bring the reward on the surface it is crucial to recognise this counter-intuitive effect and exploit it, i.e. to infer the presence of and reason about \textbf{hidden causal mechanisms}.
Here, all tested birds performed at chance level, meaning that they dropped objects randomly on either of the two lateral tubes \parencite{Jelbert2014}, see also \textcite{Logan2014a} for similar conclusions on a slightly modified setup.

All this evidence points at the fact that causal understanding could be a key notion for a more systematic study of causal cognition.
At this stage, however, debates on its presence, role and features make it unsuitable for more practical investigations (see, e.g. \textcites{Hennefield2018, Hennefield2019} for a critical meta-analysis of the works on corvids described above).
After reviewing some of the main themes driving current research on causal cognition, moving away from the debate on associative vs.\ cognitive and embracing the challenge of determining what constitutes causal understanding and how to infer it from behavioural experiments (e.g. using tasks involving causal interventions or tool-use), we thus next look at how a large body of research in this areas has been recently organised in a new conceptual framework attempting to capture the more fundamental dimensions of research works on causal understanding.

\section{A Conceptual framework for causal cognition}
\label{sec:towards-frame-ccog}

After decades of research in the field, by now it is evident that different works on causal cognition have often appealed to different conceptualisations of the subject matter, to the point that a consensus has yet to be formed on what even constitutes causal understanding, see \textcites{Penn2007, Sloman2015} for some reviews, and the contributions to \textcites{Gopnik2007, McCormack2011} for other perspectives.
As briefly illustrated by the previous section, different lines of work place causal understanding at different levels of a hypothetical cognitive spectrum, and have portrayed very diverse views on how to characterise it in terms of key cognitive functions and behavioural outputs.
Most researchers might agree on the idea that a causal agent has the kind of behavioural flexibility that is unattainable by agents lacking causal understanding.
Yet, the variety of positions trying to describe what underpins it appears to only contribute to the confusion.

Some works point to a representational strategy for agents to picture in advance what the key causal features for solving a task are. 
This would then characterise a distinction between performing and understanding, i.e. whether the solution of a task is achieved via some sort of shortcut, or in a robust and reliable way \parencite{Visalberghi1989, Visalberghi1993, Visalberghi1994, Limongelli1995}.
Others are more demanding, and see causal understanding as the result of some form of \textbf{causal reasoning}, yet another ambiguous expression that has been described in different ways.
For instance, causal reasoning could involve structural or symbolic (causal) knowledge abstracted from perceptual cues \parencite{Seed2011, Povinelli2011, McCormack2011}, or in other words, the ability to search for cause-effect relations that could reveal how and why two events are connected, or why some actions lead to certain outcomes (i.e. diagnostic causal reasoning), requiring thus the presence of some causal beliefs \parencite{Visalberghi1998, Dickinson2000}.
Others would further maintain that without an ability to perform causal interventions, perhaps even involving unobservable entities (see hidden causal mechanisms in \cref{subsec:causal-interventions}), it is unlikely that an agent is able to grasp causality as opposed to just behave \emph{as if} it does.
Going back to tool-use then, the question of whether adaptive tool use may reveal the presence of some of the abilities just described or whether it may be a confounder instead \parencite{Seed2009, Taylor2014, Jacobs2015} remains unanswered.

In a recent attempt to put causal cognition research on a more precise and coherent footing, we find different proposals discussing experimental findings framed with respect to a few recurring themes, drawing attention to key aspects of causal cognition \parencite{Woodward2021, Woodward2012, Woodward2011, Woodward2007, Starzak2021} (see also \textcite{Goddu2024} for a recent review).
\textcite{Starzak2021} in particular dissect the main disagreements over causal understanding in non-human animals, proposing a more precise way to study causal cognition using a three-dimensional conceptual framework inspired by and overall consistent with other conceptual treatments \parencite{Woodward2021, Woodward2012, Woodward2011, Woodward2007}\footnote{In the remainder of this work, the main argument will be built around the particular characterisation found in \textcite{Starzak2021} while references to the other, previous conceptualisations will be made when a closer analysis makes it necessary.}.
The starting point is to regard causal cognition as the processing of causal information, understood as information about the nature of certain causal relationships, instead of referring to the more ambiguous and ill-defined notion of causal understanding.
One of the advantages of this move is to put on one side normative issues, e.g. what really counts as causal understanding, and instead focus on empirically tractable parts of the matter \parencite{Starzak2021}.
To a first approximation, the main idea is to score the performance of subjects on causal tasks (see \cref{sec:what-causal-cog}) along three dimensions that have the potential to cover the full spectrum of causal cognition.
With these as a way to ground the discussion of different empirical results, \cite{Starzak2021} then suggest ways to draw a more fine-grained comparison of the extent to which non-human animals and humans process causal information.
More specifically, following \textcite{Starzak2021} causal information processing can be characterised along three dimensions:

\begin{enumerate}
    \item the level of \textbf{explicitness} of causal information,
	\item the \textbf{sources} of causal information, and
	\item the level of \textbf{integration} of causal information.
\end{enumerate}

\subsection{The explicitness of causal information}
\label{subsec:explicit-ci}

The explicitness dimension, refining some intuitions presented in \textcite{Woodward2011}, aims to capture a spectrum of causal information where on the one end, \textbf{implicit models} are essentially blind to causal relationships.
These models represent cases where actions and outcomes/rewards are entangled or ``fused'' \parencite{Woodward2011}, i.e. models based on an associative correlations where the causal structure (see the web of causal possibilities in \cref{subsec:causal-understanding}) is essentially hidden and inaccessible to the agent.
In this class of models, agents cannot necessarily come up with a complex plan on how to adjust their actions that is sensitive to the web of causal possibilities in order to achieve a certain goal, since they lack or have a limited understanding of their own actions and other variables in the environment as causally relevant for bringing about certain outcomes or rewards \parencite{Woodward2011}.
They can however take actions in a less structured way, for example using knowledge acquired from repeated trial-and-error in an associative manner, leading to a continuum of explicitness that is apparent in several experimental studies as seen in \cref{sec:what-causal-cog}.
For instance, associatively pairing the action of pressing a button (cause) with the presence (very often, but not always) of some food (end goal, an effect) can be considered as an example of implicit model.

On the other side of the spectrum we find \textbf{explicit models}, models where the causal structure is completely unpacked and relations between actions and outcomes/rewards are disentangled and available for an agent to take advantage of.
Looking at the previous example, we can imagine a different scenario where an agent realises that a button press will activate a food dispenser mechanism (some intermediate variable) and that the food will become available if and only if there is no obstructing object in the mechanism.
In this case, the action of clearing the dispenser from the obstructing piece is an action that can be said to require a more explicit understanding of the causal structure of the world, at least compared to the first situation, an operation that is directed at altering one of the intermediate causal variables, the object obstructing the food dispenser, so as to obtain the food.

A qualitative description of explicitness thus amounts to establishing what an agent can do with the acquired causal information, for example by investigating an agent's degree of flexibility in using what is has learned in a causal task (e.g. clearing the dispenser mechanism of the obstructing object).
To a first approximation, the key idea is that the more explicit a model is, the more causal information is available to an agent, because the means to reach a certain goal have been recognised as distinct from each other and from the goal itself (the mediating variables of a certain causal influence have been identified, \emph{cf}. \textcite{Pearl2009a}), thereby leading to a higher degree of flexibility in behavioural responses.

To see how different degrees of explicitness appear in the animal cognition literature, we can take a closer look at the research on the trap-tube task described in \cref{subsec:causal-understanding}.
Facing the trap-tube, an agent that can only form implicit models where actions and outcomes are entangled, i.e. where actions \(\actions\) leading to states \(\states\) are a-causally associated to outcomes \(\obs\), has a very limited ability to discern the possibly relevant intermediate variables that could be exploited to reach the goal. 
These include for instance the position of the trap, necessary for an understanding of whether its opening (the hole) is on the tube's lower surface or not, i.e. whether it affects the desired outcome (recall that in the latter case the tube has been rotated so the trap is ineffective).

As the literature on these experiments suggests that most capuchin monkeys that were tested do not seem to appreciate the relevance of the trap, and consequently perform poorly when the tube is inverted, we could say that these agents rely only on implicit models of the form “\emph{insert the stick, out comes the reward}” \parencite{Visalberghi1994}.
While one could say that this associative rule encodes some causal information, it is evident that this only happens in a very implicit and vague manner, leading to maladaptive behaviour in most other contexts, especially without retraining.

In contrast, causal representations capture the relevant difference-making relationships present in the task at hand, e.g.\ the role that the trap plays in the trap-tube.
Such relationships express fine-grained information that shows the different causal links between instrumental/intermediate variables and outcome/reward variables, enabling more flexibility in planning or action selection.

More in general, explicitness can be evaluated by assessing an agent performance in contexts where some kind of knowledge transfer is required.
One scenario might involve adjusting a learned strategy to solve a similar task.
For example, an agent might have learned that a rake can be used to fetch a coconut that fell into a pond without getting wet.
But the rake could also be used to detach a coconut from a palm tree without having to climb to its top, and make it fall on the ground for easy retrieval and tasting.
One could also imagine a more drastic context change requiring completely different means for performing successfully in the same task, e.g. if the rake is not available, the subject could look for a similar or different object that may play the same functional role.

Another scenario might instead require using previous knowledge or learned strategies to solve new tasks, e.g. exploiting the same means for a different end.
For example, if an agent has learned that the rake can be used to bring desirable items closer, the same agent should be able to use the rake in other contexts, e.g. to retrieve other types of food.
Finally, one might think of scenarios where knowledge transfer also demands paying attention to different functional properties of the same means (or others) because such properties are now relevant to the solution of a different problem problem.
In this case, according to \textcite{Starzak2021}, the cognitive capacities in question would amount to some form of insight learning.
For example, after having retrieved the coconut, our agent could use the rake to discourage greedy conspecifics or other animals from stealing the just-earned meal.

\subsection{Sources for learning causal information}
\label{subsec:sources-causal-info}

The sources dimension of \textcite{Starzak2021} is also inspired by \textcite{Woodward2011}, where causal cognition is proposed to be best explained in terms of some key, distinct abilities, namely, egocentric and non-egocentric sources of causal information.

\textbf{Egocentric causal information} captures the idea of an agent that can acquire an understanding of the causal structure of the world from its own behaviour, focusing on the performance of their own actions and how they can can reliably and robustly result in certain outcomes.
For example, a subject could learn that performing action \(\actions_{1}\) makes a difference for obtaining outcome \(\obs_{1}\) but not for \(\obs_{2}\).
This is the realm of instrumental conditioning (or learning) investigated extensively in animal research \parencite{Woodward2011}.

Non-egocentric causal information can, on the other hand, be obtained from two main sources: the behaviour of other agents and the unfolding of natural events.
\textbf{Social causal information} implies that an agent can learn about important action-outcome contingencies by paying attention to other agents' behaviour.
For instance, observations of a conspecific performing action \(\actions_{1}\) and reliably obtaining outcome \(\obs_{1}\), but not \(\obs_{2}\), provide important causal information for a subject that is aiming at outcome \(\obs_{1}\) (or \(\obs_{2}\)).

\textbf{Natural causal information} similarly suggests that events in the natural world can disclose ecologically important causal relationships.
For instance, observing a piece of fruit falling from a tree-branch shaken by the wind could reveal to an attentive observer important causal information on how to get some food, as long as it is capable of performing a causal analysis of the situation \parencite{Tomasello1997}.
Compared to the acquisition of the previous types of causal information, natural information imposes, arguably, a higher cognitive load on the subject because the event in question does not tell the subject what action may be (causally) relevant and for what reason(s).
In other words, there is an additional cognitive effort that the subject needs to make to appreciate that, given certain observations, action \(\actions_{1}\) can produce outcome \(\obs_{1}\).
\cite{Starzak2021} remark that empirical evidence so far suggests that egocentric causal learning is the most widespread in the animal kingdom whereas social causal learning and observational causal learning (as they call the two more sophisticated forms of causal information acquisition) are fully present in adult human beings only.
Following \textcite{Woodward2011}, they agree with the idea that in principle these forms of learning could dissociate but they also add that it is not clear the extent to which these abilities are independent from one another, or whether they form a hierarchy of cognitive processes.

\subsection{The integration of causal information}
\label{subsec:integration-ci}

Integration appears more ambiguously in \textcite{Starzak2021} but can be understood, in general, as consisting of operations of update, combination, extension etc.\ of one source of information with another one to form a coherent structure.
More concretely, in our framework integration will be later framed in terms of meaningful combinations of different sources of information (egocentric, causal and natural) that can describe if and when agents are capable of translating observations from different perspectives into their own (egocentric) perspective, forming new sources of \textbf{egocentric + social causal information}, \textbf{egocentric + natural causal information} or \textbf{ complete causal information} (egocentric + social + natural), or whether social and natural information can be integrated without a direct effect on the agent own egocentric perspective to form \textbf{social + natural causal information}.

While laying the foundations for formal characterisation of causal cognition across the animal kingdom, the inherent ambiguity of not only integration, but of explicitness and sources too, mixed with the more general focus on high-level discussion over a clear operationalisation, pushes the idea that the conceptual framework proposed by \textcite{Starzak2021} is in several ways still heavily relying on interpretative work to be done by the reader.
In the next section we fix some core ideas in a mathematical language that will form the basis of our proposal to formalise \textcite{Starzak2021}'s work in a pragmatic way in \cref{sec:computational}.

\section{A mathematical framework for causal cognition}
\label{sec:math-frame}

\subsection{Disentanglement in machine learning}
\label{subsec:disentanglementML}

A key proposal in modern approaches to deep (reinforcement) learning is that of \textbf{disentanglement} \parencite{Bengio2013}, roughly stating that in order to acquire an understanding of the causes behind some given observations, it is necessary to interpret those causes as distinct (high-level) factors, and recognise the different causal power they exert when giving rise to observations \parencite{Scholkopf2021}.
For example, if we see a red ball made of rubber bouncing on the ground, what makes it bounce? 
While different factors including colour, shape, and material, are intertwined and together produce observations captured by our eyes, some of these factors have no causal influence on the bouncing behaviour, i.e. colour.
According to the disentanglement hypothesis, the ability to discern different factors is thus a crucial step in a theory of causal understanding.

Similarly to other influential proposals, disentanglement has been used to characterise a general intuition based, however, on different implementations and interpretations.
Following \textcite{Zhang2023}'s review, we thus look at some of the common structure behind different definitions of disentanglement.
To do so, we focus in particular only on sets and functions between them.
This is technically equivalent to stating we are working in the category of sets and functions, $\mathbf{Set}$ \parencite{MacLane2013}, and while in various ways limiting, this allows us to focus on the central parts of a our proposal to connect disentanglement to explicitness down the line using a relatively simple mathematical toolkit (sets and functions, without introducing more advanced tools from category theory).

\begin{figure}[ht]
    \centering
    \scalebox{.6}{\input{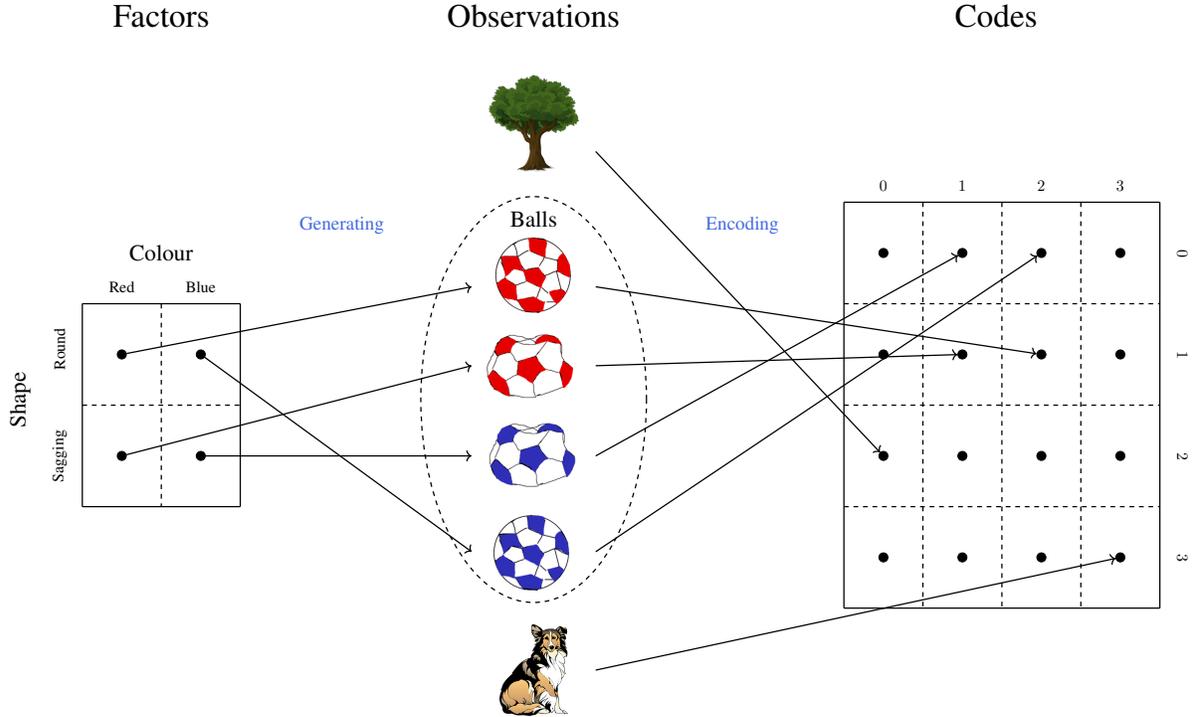}}
    \caption{\textbf{Disentanglement for a bouncing red ball.} Following the example describe in the text, we sketch here (some possible) factors, observations and codes for a system disentangling factors from observations into codes. For such a system to understand what make a red ball bounce, we consider shape and colour as factors generating observations about different balls (round and punctured balls, and of different colours, red and blue), inside the dashed-line oval shape at the centre of the figure. Other observations can be part of the standard repertoire of observables for our system (say, dogs or trees), but their factors are not explicitly drawn as we only focus on (some aspects) of a system capable of disentangling factors that generated observations of different kinds of balls. A disentangled representation is one that ``faithfully'' maps factors to codes via the given observations, according to the assumptions provided in the main text, see \cref{def:disentanglement}.}
    \label{fig:disentanglement-sketch}
\end{figure}

Following \parencite{Zhang2023} we start by defining $\Factors, \Obs, \Codes$ as state-spaces, represented as sets of (generative) \textbf{factors}, \textbf{observations} and \textbf{codes} respectively. The sets of factors and codes are further assumed to be (Cartesian) products of $\nfactors \in \N$ factors and $\ncodes \in \N$ codes:
\begin{align}
    \Factors & = \Factors_1 \times \Factors_2 \times \dots \times \Factors_\nfactors \nonumber \\
    \Codes & = \Codes_1 \times \Codes_2 \times \dots \times \Codes_\ncodes
\end{align}

At a high level, the main idea driving this framework is that representing distinct factors, i.e. having disentangled codes that faithfully map to disentangled factors, is the starting point for acquiring a causal understanding of the world: an agent with knowledge of what factors generated its observations is an agent that understand what data-generating mechanisms brought observed data to its sensory peripheries, see \cref{fig:disentanglement-sketch} for a way to frame the example of the red ball in this initial setup, to be unpacked next. The setup for disentangled representation can then be captured, in a compact form but with more specific constraints to be imposed below, by the following commutative diagram:
\begin{equation}
    \begin{tikzcd}
    	\Factors && \Obs && \Codes && \begin{array}{l} \Factors: \text{factors} \\ \Obs: \text{observations} \\ \Codes: \text{codes} \end{array}
      	\arrow["\generating", from=1-1, to=1-3]
      	\arrow["{\text{generating}}"', draw=none, from=1-1, to=1-3]
      	\arrow["\modularity", curve={height=-30pt}, from=1-1, to=1-5]
      	\arrow["{\text{modularity}}"', curve={height=-30pt}, draw=none, from=1-1, to=1-5]
      	\arrow["\encoding", from=1-3, to=1-5]
      	\arrow["{\text{encoding}}"', draw=none, from=1-3, to=1-5]
      	\arrow["\informativeness"', curve={height=-30pt}, from=1-5, to=1-1]
      	\arrow["{\text{informativeness}}", curve={height=-30pt}, draw=none, from=1-5, to=1-1]
    \end{tikzcd}
\end{equation}
We introduce next the formal definition of disentanglement that we will be referring to throughout this work, following \textcite{Zhang2023}.
\begin{definition}[Disentangled representations]
  \label{def:disentanglement}
  A disentangled representation is a product of codes $\Codes_i$ for $i \in \N$ defined through the following:
  \begin{itemize}
    \item a \textbf{generative process} or data generating process, as a function 
    \begin{align}
      \generating: \Factors \to \Obs
    \end{align}
    that gives rise to observations $\Obs$ from the product of $\nfactors$ factors $\Factors = \Factors_1 \times \Factors_2 \times \dots \times \Factors_\nfactors$; this function is assumed to be injective since we ask that if two observations are identical, they must be mapped to by two identical products of factors, but also allow for the possibility that some observations may not be mapped to from any factor (i.e. $\generating$ is not surjective, which can be seen as stating that the state-space of observations can be of much higher dimensionality compared to the state-space of factors; intuitively, the space of observations could include all the possible scenes a subject has ever been exposed to, but the factors $\Factors$ producing data, e.g. about a ball bouncing around, only map to a small part of all possible observations (all possible variations of the red bouncing ball, while for example saying nothing about a tree whose leaves are moved by the wind, see \cref{fig:disentanglement-sketch}),
    
    \item an \textbf{encoding}, as a function 
    \begin{align}
      \encoding: \Obs \to \Codes
    \end{align}
    that maps observations $\Obs$ to the product of $\ncodes$ codes $\Codes_1 \times \Codes_2 \times \dots \times \Codes_\ncodes$ while also requiring $\encoding$ to be injective on $\generating$'s image, $\generating(\Factors)$; for a more intuitive interpretation we consider the following scenarios:
    \begin{itemize}
      \item $\ncodes \ge \nfactors$, the case considered here (and in \textcite{Suter2019}, see next section), meaning that we assume there are potentially more codes than factors and thus all factors can be encoded by different codes (independently, after we impose more structure below),
      \item $\ncodes = \nfactors$, the case considered by \textcite{Zhang2023} for a slightly simpler treatment, as this doesn't anyway imply that factors $\Factors$ and codes $\Codes$ are the same (and in practice they often aren't, since we rarely know what factors $\Factors$ generated observations in a given dataset),
      \item $\ncodes < \nfactors$, relevant for practical implementations (see for instance variational autoencoders in \cref{subsec:explicit-in-crl-agents}), where we may not have (or want, for reasons including for instance lossy compression) enough codes to model all factors due to constraints or wrong modelling choices, in this case we need to consider approximations to disentanglement, either mathematically \parencite{Zhang2024} or via some (deep) reinforcement learning implementation, which will be discussed in \cref{subsec:explicit-in-crl-agents},
    \end{itemize}
    
    \item a \textbf{modularity} map,
    \begin{align}
      \modularity: \Factors \to \Codes
    \end{align}
    defined as the composition $\modularity = \encoding \circ \generating$\footnote{Think of $\encoding \circ \generating$ as $\encoding(\generating(\factors))$ for some element $\factors \in \Factors$.}, i.e. using $\nfactors$ factors $\Factors$ to generate observations $\Obs$ and then encoding these observations, $\generating(\Factors)$, in $\nfactors (\le \ncodes)$ codes $\Codes$, the constraint can also be visually expressed in the following (string diagrammatic) form to be read bottom to top
    \begin{align}
    	\centering
        \scalebox{.8}{\begin{tikzpicture}
	\begin{pgfonlayer}{nodelayer}
		\node [style=box-3-.5] (0) at (-5, 0.25) {$\modularity$};
		\node [style=none] (1) at (-7, -0.25) {};
		\node [style=none] (2) at (-7, -1.5) {};
		\node [style=none] (3) at (-5.75, -0.25) {};
		\node [style=none] (4) at (-5.75, -1.5) {};
		\node [style=none] (5) at (-3.75, -0.25) {};
		\node [style=none] (6) at (-3.75, -1.5) {};
		\node [style=none] (7) at (-4.75, -1) {$\dots$};
		\node [style=none] (8) at (-7.5, -1) {$\Factors_1$};
		\node [style=none] (9) at (-6.25, -1) {$\Factors_2$};
		\node [style=none] (10) at (-3, -1) {$\Factors_\nfactors$};
		\node [style=none] (11) at (-7, 2) {};
		\node [style=none] (12) at (-7, 0.75) {};
		\node [style=none] (13) at (-5.75, 2) {};
		\node [style=none] (14) at (-5.75, 0.75) {};
		\node [style=none] (15) at (-3.75, 2) {};
		\node [style=none] (16) at (-3.75, 0.75) {};
		\node [style=none] (17) at (-4.75, 1.5) {$\dots$};
		\node [style=none] (18) at (-7.5, 1.5) {$\Codes_1$};
		\node [style=none] (19) at (-6.25, 1.5) {$\Codes_2$};
		\node [style=none] (20) at (-3, 1.5) {$\Codes_\nfactors$};
		\node [style=none] (21) at (-1.25, 0.25) {=};
		\node [style=box-.5-.5] (22) at (0.5, 0.25) {$\modularity_{1,1}$};
		\node [style=box-.5-.5] (23) at (2.75, 0.25) {$\modularity_{2,2}$};
		\node [style=box-.5-.5] (24) at (6.25, 0.25) {$\modularity_{n,n}$};
		\node [style=none] (25) at (0.5, 2) {};
		\node [style=none] (26) at (2.75, 2) {};
		\node [style=none] (27) at (6.25, 2) {};
		\node [style=none] (28) at (0.5, -1.5) {};
		\node [style=none] (29) at (2.75, -1.5) {};
		\node [style=none] (30) at (6.25, -1.5) {};
		\node [style=none] (31) at (4.5, -1) {$\dots$};
		\node [style=none] (32) at (4.5, 1.5) {$\dots$};
		\node [style=none] (33) at (4.5, 0.25) {$\dots$};
		\node [style=none] (34) at (0, 1.5) {$\Codes_1$};
		\node [style=none] (35) at (2.25, 1.5) {$\Codes_2$};
		\node [style=none] (36) at (7, 1.5) {$\Codes_\nfactors$};
		\node [style=none] (37) at (0, -1) {$\Factors_1$};
		\node [style=none] (38) at (2.25, -1) {$\Factors_2$};
		\node [style=none] (39) at (7, -1) {$\Factors_\nfactors$};
	\end{pgfonlayer}
	\begin{pgfonlayer}{edgelayer}
		\draw [style=thick-line] (2.center) to (1.center);
		\draw [style=thick-line] (4.center) to (3.center);
		\draw [style=thick-line] (6.center) to (5.center);
		\draw [style=thick-line] (12.center) to (11.center);
		\draw [style=thick-line] (14.center) to (13.center);
		\draw [style=thick-line] (16.center) to (15.center);
		\draw [style=thick-line] (22) to (28.center);
		\draw [style=thick-line] (25.center) to (22);
		\draw [style=thick-line] (26.center) to (23);
		\draw [style=thick-line] (23) to (29.center);
		\draw [style=thick-line] (27.center) to (24);
		\draw [style=thick-line] (24) to (30.center);
	\end{pgfonlayer}
\end{tikzpicture}}
    	\label{eqn:modularity}
    \end{align}
    stating that in the case of $\ncodes \ge \nfactors$\footnote{If $\ncodes < \nfactors$, we can simply replace $\nfactors$ with $\ncodes$.}, $\modularity = \modularity_{1,1} \times \modularity_{2,2} \times \dots \times \modularity_{\nfactors,\nfactors}$, meaning that the $\nfactors$-th code only encodes the $\nfactors$-th factor, note that $\modularity$ is injective as the composition of an injective function, $\generating$, and a function injective under its image $\generating(\Factors)$, $\encoding$,
    
    \item an \textbf{informativeness} requirement, modelled as the existence of as a function 
    \begin{equation}
      \informativeness : \Codes \to \Factors
    \end{equation} 
    that maps codes to factors in such a way that $\informativeness \circ \modularity = \id_\Factors$\footnote{Technically, we say that $\modularity$ is a split monomorphism \parencite{MacLane2013}.} for the identity map on factors $\id_\Factors$, see the following\footnote{Notice that since we treat modularity and informativeness as separate criteria, we don't include the modular/factorised version of $\modularity$ in \cref{eqn:explicitness}.}
    \begin{equation}
    	\centering
        \scalebox{.8}{\begin{tikzpicture}
	\begin{pgfonlayer}{nodelayer}
		\node [style=box-3-.5] (0) at (-9, -1) {$\modularity$};
		\node [style=none] (1) at (-11, -1.5) {};
		\node [style=none] (2) at (-11, -2.75) {};
		\node [style=none] (3) at (-9.75, -1.5) {};
		\node [style=none] (4) at (-9.75, -2.75) {};
		\node [style=none] (5) at (-7.75, -1.5) {};
		\node [style=none] (6) at (-7.75, -2.75) {};
		\node [style=none] (7) at (-8.75, -2.25) {$\dots$};
		\node [style=none] (8) at (-11.5, -2.25) {$\Factors_1$};
		\node [style=none] (9) at (-10.25, -2.25) {$\Factors_2$};
		\node [style=none] (10) at (-7, -2.25) {$\Factors_\nfactors$};
		\node [style=none] (11) at (-11, 0.75) {};
		\node [style=none] (12) at (-11, -0.5) {};
		\node [style=none] (13) at (-9.75, 0.75) {};
		\node [style=none] (14) at (-9.75, -0.5) {};
		\node [style=none] (15) at (-7.75, 0.75) {};
		\node [style=none] (16) at (-7.75, -0.5) {};
		\node [style=none] (17) at (-8.75, 0) {$\dots$};
		\node [style=none] (18) at (-11.5, 0) {$\Codes_1$};
		\node [style=none] (19) at (-10.25, 0) {$\Codes_2$};
		\node [style=none] (20) at (-7, 0) {$\Codes_\nfactors$};
		\node [style=none] (21) at (-4.5, 0) {=};
		\node [style=box-3-.5] (40) at (-9, 1.25) {$i$};
		\node [style=none] (41) at (-11, 3) {};
		\node [style=none] (42) at (-11, 1.75) {};
		\node [style=none] (43) at (-9.75, 3) {};
		\node [style=none] (44) at (-9.75, 1.75) {};
		\node [style=none] (45) at (-7.75, 3) {};
		\node [style=none] (46) at (-7.75, 1.75) {};
		\node [style=none] (65) at (-8.75, 2.5) {$\dots$};
		\node [style=none] (66) at (-11.5, 2.5) {$\Factors_1$};
		\node [style=none] (67) at (-10.25, 2.5) {$\Factors_2$};
		\node [style=none] (68) at (-7, 2.5) {$\Factors_\nfactors$};
		\node [style=none] (71) at (-2, -2.75) {};
		\node [style=none] (73) at (-0.75, -2.75) {};
		\node [style=none] (75) at (1.25, -2.75) {};
		\node [style=none] (76) at (0.25, -2.25) {$\dots$};
		\node [style=none] (77) at (-2.5, -2.25) {$\Factors_1$};
		\node [style=none] (78) at (-1.25, -2.25) {$\Factors_2$};
		\node [style=none] (79) at (2, -2.25) {$\Factors_\nfactors$};
		\node [style=none] (91) at (-2, 3) {};
		\node [style=none] (93) at (-0.75, 3) {};
		\node [style=none] (95) at (1.25, 3) {};
		\node [style=none] (97) at (0.25, 2.5) {$\dots$};
		\node [style=none] (98) at (-2.5, 2.5) {$\Factors_1$};
		\node [style=none] (99) at (-1.25, 2.5) {$\Factors_2$};
		\node [style=none] (100) at (2, 2.5) {$\Factors_\nfactors$};
		\node [style=box-3-.5] (101) at (0, 0) {$\id_S$};
		\node [style=none] (102) at (-2, 0.5) {};
		\node [style=none] (103) at (-2, -0.5) {};
		\node [style=none] (104) at (-0.75, 0.5) {};
		\node [style=none] (105) at (-0.75, -0.5) {};
		\node [style=none] (106) at (1.25, 0.5) {};
		\node [style=none] (107) at (1.25, -0.5) {};
		\node [style=none] (111) at (7.5, -2.75) {};
		\node [style=none] (112) at (8.75, -2.75) {};
		\node [style=none] (113) at (10.75, -2.75) {};
		\node [style=none] (114) at (9.75, -2.25) {$\dots$};
		\node [style=none] (115) at (7, -2.25) {$\Factors_1$};
		\node [style=none] (116) at (8.25, -2.25) {$\Factors_2$};
		\node [style=none] (117) at (11.5, -2.25) {$\Factors_\nfactors$};
		\node [style=none] (118) at (7.5, 3) {};
		\node [style=none] (119) at (8.75, 3) {};
		\node [style=none] (120) at (10.75, 3) {};
		\node [style=none] (121) at (9.75, 2.5) {$\dots$};
		\node [style=none] (122) at (7, 2.5) {$\Factors_1$};
		\node [style=none] (123) at (8.25, 2.5) {$\Factors_2$};
		\node [style=none] (124) at (11.5, 2.5) {$\Factors_\nfactors$};
		\node [style=none] (125) at (4.5, 0) {=};
	\end{pgfonlayer}
	\begin{pgfonlayer}{edgelayer}
		\draw [style=thick-line] (2.center) to (1.center);
		\draw [style=thick-line] (4.center) to (3.center);
		\draw [style=thick-line] (6.center) to (5.center);
		\draw [style=thick-line] (12.center) to (11.center);
		\draw [style=thick-line] (14.center) to (13.center);
		\draw [style=thick-line] (16.center) to (15.center);
		\draw [style=thick-line] (42.center) to (41.center);
		\draw [style=thick-line] (44.center) to (43.center);
		\draw [style=thick-line] (46.center) to (45.center);
		\draw [style=thick-line] (118.center) to (111.center);
		\draw [style=thick-line] (119.center) to (112.center);
		\draw [style=thick-line] (120.center) to (113.center);
		\draw [style=thick-line] (102.center) to (91.center);
		\draw [style=thick-line] (103.center) to (71.center);
		\draw [style=thick-line] (105.center) to (73.center);
		\draw [style=thick-line] (107.center) to (75.center);
		\draw [style=thick-line] (106.center) to (95.center);
		\draw [style=thick-line] (104.center) to (93.center);
	\end{pgfonlayer}
\end{tikzpicture}}
    	\label{eqn:explicitness}
    \end{equation}
    i.e. $\informativeness$ is a left inverse for $\modularity$, or in other words (post)composing $\modularity$ with its left inverse $\informativeness$ means that factors can be recovered (we have identities on the right hand side of \cref{eqn:explicitness}), it doesn't however tell us that they will be disentangled ($\informativeness$ is not itself modular/factorised),

    \item a \textbf{disentanglement} requisite on informativeness, this ensures that the codes are \emph{independent} faithful representations of different factors and can be stated using the following diagram
    \begin{equation}
    	\centering
        \scalebox{.8}{\begin{tikzpicture}
	\begin{pgfonlayer}{nodelayer}
		\node [style=box-3-.5] (0) at (-4.25, -0.75) {$\modularity$};
		\node [style=none] (1) at (-6.25, -1.25) {};
		\node [style=none] (2) at (-6.25, -2.5) {};
		\node [style=none] (3) at (-5, -1.25) {};
		\node [style=none] (4) at (-5, -2.5) {};
		\node [style=none] (5) at (-3, -1.25) {};
		\node [style=none] (6) at (-3, -2.5) {};
		\node [style=none] (7) at (-4, -2) {$\dots$};
		\node [style=none] (8) at (-6.75, -2) {$\Factors_1$};
		\node [style=none] (9) at (-5.5, -2) {$\Factors_2$};
		\node [style=none] (10) at (-2.25, -2) {$\Factors_\nfactors$};
		\node [style=none] (11) at (-6.25, 1) {};
		\node [style=none] (12) at (-6.25, -0.25) {};
		\node [style=none] (13) at (-5, 1) {};
		\node [style=none] (14) at (-5, -0.25) {};
		\node [style=none] (15) at (-3, 1) {};
		\node [style=none] (16) at (-3, -0.25) {};
		\node [style=none] (17) at (-4, 0.25) {$\dots$};
		\node [style=none] (18) at (-6.75, 0.25) {$\Codes_1$};
		\node [style=none] (19) at (-5.5, 0.25) {$\Codes_2$};
		\node [style=none] (20) at (-2.25, 0.25) {$\Codes_\nfactors$};
		\node [style=none] (21) at (0.25, 0.25) {=};
		\node [style=box-.5-.5] (22) at (2.5, 1.5) {$\informativeness_{1,1}$};
		\node [style=box-.5-.5] (23) at (4.75, 1.5) {$\informativeness_{2,2}$};
		\node [style=box-.5-.5] (24) at (8.5, 1.5) {$\informativeness_{n,n}$};
		\node [style=none] (25) at (2.5, 3.25) {};
		\node [style=none] (26) at (4.75, 3.25) {};
		\node [style=none] (27) at (8.5, 3.25) {};
		\node [style=none] (28) at (2.5, -0.25) {};
		\node [style=none] (29) at (4.75, -0.25) {};
		\node [style=none] (30) at (8.5, -0.25) {};
		\node [style=none] (31) at (6.5, 2.75) {$\dots$};
		\node [style=none] (33) at (6.5, 1.5) {$\dots$};
		\node [style=none] (37) at (2, 2.75) {$\Factors_1$};
		\node [style=none] (38) at (4.25, 2.75) {$\Factors_2$};
		\node [style=none] (39) at (9.25, 2.75) {$\Factors_\nfactors$};
		\node [style=box-3-.5] (40) at (-4.25, 1.5) {$\informativeness$};
		\node [style=none] (41) at (-6.25, 3.25) {};
		\node [style=none] (42) at (-6.25, 2) {};
		\node [style=none] (43) at (-5, 3.25) {};
		\node [style=none] (44) at (-5, 2) {};
		\node [style=none] (45) at (-3, 3.25) {};
		\node [style=none] (46) at (-3, 2) {};
		\node [style=box-4-.5] (51) at (5.5, -0.75) {$\modularity$};
		\node [style=none] (52) at (3.5, -1.25) {};
		\node [style=none] (53) at (3.5, -2.5) {};
		\node [style=none] (54) at (4.75, -1.25) {};
		\node [style=none] (55) at (4.75, -2.5) {};
		\node [style=none] (56) at (6.75, -1.25) {};
		\node [style=none] (57) at (6.75, -2.5) {};
		\node [style=none] (58) at (5.75, -2) {$\dots$};
		\node [style=none] (59) at (3, -2) {$\Factors_1$};
		\node [style=none] (60) at (4.25, -2) {$\Factors_2$};
		\node [style=none] (61) at (7.5, -2) {$\Factors_\nfactors$};
		\node [style=none] (62) at (2, 0.25) {$\Codes_1$};
		\node [style=none] (63) at (4.25, 0.25) {$\Codes_2$};
		\node [style=none] (64) at (9.25, 0.25) {$\Codes_\nfactors$};
		\node [style=none] (65) at (-4, 2.75) {$\dots$};
		\node [style=none] (66) at (-6.75, 2.75) {$\Factors_1$};
		\node [style=none] (67) at (-5.5, 2.75) {$\Factors_2$};
		\node [style=none] (68) at (-2.25, 2.75) {$\Factors_\nfactors$};
		\node [style=none] (69) at (6.5, 0.25) {$\dots$};
	\end{pgfonlayer}
	\begin{pgfonlayer}{edgelayer}
		\draw [style=thick-line] (2.center) to (1.center);
		\draw [style=thick-line] (4.center) to (3.center);
		\draw [style=thick-line] (6.center) to (5.center);
		\draw [style=thick-line] (12.center) to (11.center);
		\draw [style=thick-line] (14.center) to (13.center);
		\draw [style=thick-line] (16.center) to (15.center);
		\draw [style=thick-line] (22) to (28.center);
		\draw [style=thick-line] (25.center) to (22);
		\draw [style=thick-line] (26.center) to (23);
		\draw [style=thick-line] (23) to (29.center);
		\draw [style=thick-line] (27.center) to (24);
		\draw [style=thick-line] (24) to (30.center);
		\draw [style=thick-line] (42.center) to (41.center);
		\draw [style=thick-line] (44.center) to (43.center);
		\draw [style=thick-line] (46.center) to (45.center);
		\draw [style=thick-line] (53.center) to (52.center);
		\draw [style=thick-line] (55.center) to (54.center);
		\draw [style=thick-line] (57.center) to (56.center);
	\end{pgfonlayer}
\end{tikzpicture}}
    	\label{eqn:disentanglement}
    \end{equation}
    where $\informativeness = \informativeness_{1,1} \times \informativeness_{2,2} \times \dots \times \informativeness_{n,n}$; \textcolor{black}{this condition is particularly relevant because there may be cases where, for instance, without a proper causal understanding of the factors generating a bouncing ball, one might mistakenly assume material and colour, together, are a relevant factor: if all the balls an agent ever saw bouncing were red and made of rubber, colour and material could be assumed to be \emph{jointly} necessary factors to understand how the ball bounces\footnote{Further constraints, of general importance for a more precise definition of disentangled representations that can deal with undesirable special cases (e.g. redundancy of information in the codes) can be found in \textcite{Zhang2023} but will not be further discussed here.}.}
  \end{itemize}
Combining these conditions, we obtain finally the following diagram for \textbf{disentangled representations} $\Codes_1 \times \dots \times \Codes_\nfactors$ satisfying the following equation
\begin{equation}
    \centering
    \scalebox{.8}{\begin{tikzpicture}
	\begin{pgfonlayer}{nodelayer}
		\node [style=box-.5-.5] (22) at (-8.75, -1.25) {$\modularity_{1,1}$};
		\node [style=box-.5-.5] (23) at (-6.5, -1.25) {$\modularity_{2,2}$};
		\node [style=box-.5-.5] (24) at (-3, -1.25) {$\modularity_{n,n}$};
		\node [style=none] (28) at (-8.75, -3) {};
		\node [style=none] (29) at (-6.5, -3) {};
		\node [style=none] (30) at (-3, -3) {};
		\node [style=none] (31) at (-4.75, -2.5) {$\dots$};
		\node [style=none] (32) at (-4.75, 0) {$\dots$};
		\node [style=none] (33) at (-4.75, -1.25) {$\dots$};
		\node [style=none] (34) at (-9.25, 0) {$\Codes_1$};
		\node [style=none] (35) at (-7, 0) {$\Codes_2$};
		\node [style=none] (36) at (-2.25, 0) {$\Codes_n$};
		\node [style=none] (37) at (-9.25, -2.5) {$\Factors_1$};
		\node [style=none] (38) at (-7, -2.5) {$\Factors_2$};
		\node [style=none] (39) at (-2.25, -2.5) {$\Factors_n$};
		\node [style=box-.5-.5] (62) at (-8.75, 1.25) {$\informativeness_{1,1}$};
		\node [style=box-.5-.5] (63) at (-6.5, 1.25) {$\informativeness_{2,2}$};
		\node [style=box-.5-.5] (64) at (-3, 1.25) {$\informativeness_{n,n}$};
		\node [style=none] (65) at (-8.75, 3) {};
		\node [style=none] (66) at (-6.5, 3) {};
		\node [style=none] (67) at (-3, 3) {};
		\node [style=none] (71) at (-4.75, 2.5) {$\dots$};
		\node [style=none] (72) at (-4.75, 1.25) {$\dots$};
		\node [style=none] (73) at (-9.25, 2.5) {$\Factors_1$};
		\node [style=none] (74) at (-7, 2.5) {$\Factors_2$};
		\node [style=none] (75) at (-2.25, 2.5) {$\Factors_n$};
		\node [style=none] (76) at (2.5, -2.75) {};
		\node [style=none] (77) at (3.75, -2.75) {};
		\node [style=none] (78) at (5.75, -2.75) {};
		\node [style=none] (79) at (4.75, -2.25) {$\dots$};
		\node [style=none] (80) at (2, -2.25) {$\Factors_1$};
		\node [style=none] (81) at (3.25, -2.25) {$\Factors_2$};
		\node [style=none] (82) at (6.5, -2.25) {$\Factors_\nfactors$};
		\node [style=none] (83) at (2.5, 3) {};
		\node [style=none] (84) at (3.75, 3) {};
		\node [style=none] (85) at (5.75, 3) {};
		\node [style=none] (86) at (4.75, 2.5) {$\dots$};
		\node [style=none] (87) at (2, 2.5) {$\Factors_1$};
		\node [style=none] (88) at (3.25, 2.5) {$\Factors_2$};
		\node [style=none] (89) at (6.5, 2.5) {$\Factors_\nfactors$};
		\node [style=none] (90) at (-0.5, 0) {=};
	\end{pgfonlayer}
	\begin{pgfonlayer}{edgelayer}
		\draw [style=thick-line] (22) to (28.center);
		\draw [style=thick-line] (23) to (29.center);
		\draw [style=thick-line] (24) to (30.center);
		\draw [style=thick-line] (65.center) to (62);
		\draw [style=thick-line] (66.center) to (63);
		\draw [style=thick-line] (67.center) to (64);
		\draw [style=thick-line] (22) to (62);
		\draw [style=thick-line] (23) to (63);
		\draw [style=thick-line] (24) to (64);
		\draw [style=thick-line] (83.center) to (76.center);
		\draw [style=thick-line] (84.center) to (77.center);
		\draw [style=thick-line] (85.center) to (78.center);
	\end{pgfonlayer}
\end{tikzpicture}}
    \label{eqn:disentangledrepresentations}
\end{equation}
\end{definition}

In this view, a disentangled representation is thus one that captures or reflects in a meaningful way the expressivity of data-generating factors underlying some given observations. 
While useful as a general guideline for a more precise notion of disentanglement, it is however still unclear at this stage how to relate this notion of disentanglement to different dimensions of causal understanding \parencite{Starzak2021} in a more formal way.
So far, we haven't in fact discussed how this idea can be connected to formal accounts of causality.
To form the basis of this connection we thus introduce next some key concepts from causal representation learning.
Due to the probabilistic setup underlying the following definition(s), one could argue that the notion of disentanglement we reviewed here (using sets and functions) can't be applied directly to the models in the following sections.
However, relatively straightforward generalisations of the notion of disentanglement to categories other than $\mathbf{Set}$, including (Markov) categories handling probabilistic reasoning in a ``function-like'' manner, have already been put forward in \textcite{Zhang2023}.
With this in mind, we will from here onwards use ``disentanglement'' in a slightly looser way, taking it capture different flavours of disentangled representations with the same class of structural desiderata but applied to different, e.g. probabilistic, setups.

\subsection{Causal representation learning}
\label{subsec:from-disent-to-crl}

As highlighted by \textcite{Zhang2023}, disentanglement has been described in various different ways across the literature.
In one of the most influential accounts, disentanglement can be viewed as a component of causal models recovering the causal factorisation of a process generating a collection of observations of interest \parencite{Scholkopf2021, Wang2022}.
In this view, disentanglement is thus a crucial part of the answer to the question of how causal model are acquired, providing a way to operationalise a process deemed necessary for an agent to \emph{learn} a causal characterisation of the world it acts in \parencite{Scholkopf2021, Wang2022}.
As we will see, taking this perspective allows us to relate explicitness, one of the dimensions of causal cognition (see~\cref{subsec:explicit-ci}), to disentanglement, in light of the nascent field of causal machine learning. 

Causal machine learning is a collection of methods and applications based on the notion that exploiting causal information in data can lead to a more robust, accurate, and efficient kind of (data or system) modelling, thereby viewing causality as a fundamental notion to move past some of the limitations of machine learning methods based on statistical learning (from now on we will refer to these methods as ``traditional machine learning'') \parencite{Scholkopf2022, Kaddour2022, Tibshirani2009}.

Within this line of research, the subfield of causal representation learning can be regarded as a way to recover disentangled representations from data.
Traditionally, representation learning has been conceived as the task of learning a \textbf{generative model} in the form of a low-dimensional feature vector (codes) of high-dimensional data (observations) produced by a \textbf{generative process} whose features (factors) remain hidden.
The idea driving this approach is that if those codes capture key, informative, aspects of a dataset, they would aid in solving downstream tasks (i.e. predicting a label) \parencite{Bengio2013}.
However, these models have often sidestepped questions regarding the origins of particular datasets, overlooking structural knowledge of the data-generating process that could have produced them.
This in turn affects what a generative model can account for, often limiting its scope to only statistical correlations with little to no causal power.

Causal representation learning extends these ideas by bringing into representation learning (and, more generally, deep learning) some of the principles, methodologies, and objectives of classic causal inference research \parencite{Pearl2009a, Peters2017, Hernan2020}, with the goal to learn a low-dimensional vector of \emph{causal} codes from high-dimensional observations generated by \emph{causal} factors\footnote{Note that in the causal representation learning literature, both causal codes and causal factors are usually addressed as simply ``causal variables''. Here we will instead maintain an explicit distinction.} \parencite{Scholkopf2021, Kaddour2022, Berrevoets2023}.
In this paradigm, the data-generating process can be formalised as a \textbf{structural causal model} capturing the causal relationships between factors underlying the data distribution.
Importantly, recalling the distinction between generative process and generative model, learning a generative model means to represent something about the generative process described as a structural causal model.
In the best case scenario, a generative model recovering the full gamut of causal information assumed to exist in the generative process can itself be described as a structural causal model of the same form as the generative process.
This particular scenario assumes however \emph{causal sufficiency}, i.e. that there are no hidden common causes (also referred to as hidden \textbf{confounders}) on factors in the generative process, meaning that every common cause of any two or more variables is already accounted for and included in the model (see \textcite[Ch. 10]{Spirtes2000} and \textcite[Ch. 9]{Peters2017}).
More often, we will instead only look at cases where hidden confounders are present, thereby violating causal sufficiency, which will then allow us to distinguish between weak and strong disentanglement in \cref{subsec:explicit-in-crl-agents}\footnote{Note that a causal factor or code does not have to be hidden to be a confounder. Any common cause of two or more variables has a confounding effect on certain causal relationships (see the definition of confounding in \textcite[113]{Peters2017}).}.
To better understand the role confounders can play, we define next a structural causal model as the following.

\begin{definition}[Structural causal model of the generative process]
  \label{def:scm}
  Given the following:
  \begin{itemize}
    \item a collection $\Factors = (\Factors_{1}, \dots, \Factors_{\nfactors})$ of $\nfactors \in \N$ causal variables, or causal factors,
    \item a collection $\Obs = (\Obs_{1}, \dots, \Obs_{\nobs})$ of $\nobs \, (\ge \nfactors) \in \N$ observables,
    \item a collection  $\Confounders = (\Confounders_{1}, \dots, \Confounders_{\nobs})$ of $\nconfounders \in \N$ confounders,
    \item a collection $\noise^{\Factors} = (\noise^{\Factors}_{1}, \dots, \noise^{\Factors}_{\nfactors})$ of $\nfactors$ noise variables on causal variables,
    \item a collection $\noise^{\Obs} = (\noise^{\Factors}_{1}, \dots, \noise^{\Factors}_{\nobs})$ of $\nobs$ noise variables on observables,
  \end{itemize}
  and assuming that all the noise variables are jointly independent, a structural causal model $\scm$ \parencite{Pearl2009a} of the data-generating process is:
  \begin{itemize}  
    \item a collection $(\assignment_{1}, \dots, \assignment_{\nfactors})$ of $\nfactors$ structural assignments, each assigning a value to a corresponding causal variables $\Factors_{j}$ for $j \in \{1, \dots, \nfactors\}$ based on 
    \begin{itemize}
        \item its parents (direct causes) in the set 
            \begin{align}
                \PA_{j} \subset \{\Factors_{\setminus j}, \Confounders\} \quad (\text{where } \Factors_{\setminus j} \coloneqq \Factors \setminus \Factors_{j})
                \label{eqn:parents}
            \end{align}
            and,
        \item the noise variable \(\noise^{\Factors}_{j}\),
    \end{itemize}
    \item and an emission map (or mixing function) $\emission$ generating observables $\Obs$, cf.\ the generating process in \cref{def:disentanglement},
  \end{itemize} 
  that is:
  \begin{align}
    \label{eq:str-ass}
    \Factors_{j} & = \assignment_{j}\, (\PA_{j}, \noise^{\Factors}_{j}), \qquad j \in \{1, \dots, \nfactors\} \nonumber \\
    \Obs & = \emission \, (\Factors, \noise^{\Obs}). 
  \end{align}
\end{definition}

Importantly, one can show that a structural causal model $\scm$ entails a corresponding directed acyclic graph, \(\mathcal{G}\), and a unique probability distribution\footnote{For the use of probabilistic language from now onwards, the following conventions are adopted:
\begin{enumerate}
  \item \(P\) stands for a probability distribution or function, and \(P_{X}\) is the probability distribution for a random vector \(X\) (if one-dimensional, $X$ should be understood as a random variable),
  \item \(P(X=x)\) is shortened by simply writing \(P(x)\), i.e. using only the value the random variable takes,
  \item \(p(x)\) is either the probability mass function or probability density function evaluated at \(x\) for the probability distribution \(P_{X}\),
  \item the subscript of \(P_{X}\) will be usually omitted if the random variable is clear from context.
\end{enumerate}} 
$P_{\Factors, \Obs}$ defined over factors $\Factors$ and \(\Obs\) \parencite{Peters2017}.
Concisely, a structural causal model induces a causal Bayesian network, where the conditional distributions relating causes to effects, e.g. \(P(\Obs|\Factors), P(\Factors_{i}|\Factors_{j})\), etc., are often referred to as \textbf{causal mechanisms}, capturing ways in which causes produce effects (the underlying structural assignments of the structural causal model describe those mechanisms in greater detail, providing a functional specification).
Conversely, we can also think that any empirical probability distribution has an associated structural causal model that induces it.
In this case, however, it can be shown that such a structural causal model is not unique.
It is nonetheless possible to define an equivalence class of structural causal models consistent with that same distribution \parencite{Peters2017}.
Starting from a joint probability distribution $P_{\Factors, \Obs}$ rather than a specific model $\scm$, there is a corresponding equivalence relation, or partition, of directed acyclic graphs with respect to which the $P_{\Factors, \Obs}$ can be factorised.
A particular graph can be chosen, then, to reflect the particular causal mechanisms of the true structural causal models.

While crucial for unpacking a notion of disentangled representations with connections to the literature on causal representation learning, the tools we introduced so far remain fundamentally rooted in a body of work in machine learning that does not take into account how agents make use of these tools for decision making over time.
To introduce a notion of agent as a system with goals capable of solving a particular class of problems we thus look at the framework of reinforcement learning \cite{Sutton2018} and review some of its basic components that will be later involved in our account of disentanglement for agents.
From the perspective of \textcite{Zhang2023}, one can reasonably expect that the definition of disentanglement given in \cref{subsec:disentanglementML} can also be generalised to classes of models with dynamics (i.e. time-dependent models)\footnote{For example, by using monoids instead of groups in \textcite{Zhang2023}.}, however at this stage the possibly non-trivial interplay between the categorical (in the sense of being based on category theory) disentanglement of \textcite{Zhang2023} and a comparable categorical reinforcement learning setup \parencite{Hedges2024} remains admittedly unclear\footnote{For instance, we can consider bisimulations of POMDPs, described in \cref{subsubsec:strong-disentanglement-examples} as a possible implementation to achieve a high degree of disentanglement, and whether the definition of equivalence classes of states it embodies includes rewards or not. If it doesn't, one can obtain equivalence classes of states for all actions with transition dynamics leading to the same equivalence classes of states, i.e. a task-irrelevant partition of states consistent with their dynamics that does not depend on the reward signal perceived by a particular agent. If it includes rewards on the other hand, one in general obtains a finer partition that considers states \emph{having the same reward} from which transitions for all actions lead to the same equivalence classes of states, i.e. a partition of states sensitive to instantaneous rewards and thus to task-relevant aspects of a problem for a particular agent.}.

\subsection{(Causal) Reinforcement learning}
\label{sec:causalRL}

\textbf{Reinforcement learning (RL)} provides a natural avenue for ways to combine work from machine learning and decision making in agents \parencite{Sutton2018}.
Similarities between classical RL and causality have been put forward in previous works \parencite{Gershman2010, Gershman2015, Gershman2017}, however a clear-cut notion of causality appears to be missing \parencite{Kaddour2022}.
Here we provide some background for standard RL implementations, which will be then placed in context and used once we overview recent work in causal RL in \cref{sec:bridging}.

A typical reinforcement learning setup involves the definition of a problem in terms of (a model of) an environment represented by a (discrete-time) \textbf{Markov decision process (MDP)}.
\begin{definition}[Markov decision process (MDP)]
  A Markov decision process is a tuple $(\States, \Actions, \Transitions, \discount, \reward)$, where:
  \begin{itemize}
    \item $\States$ is the state space\footnote{Our choice of using $\States$ to represent states of the environment conforms with the distinction between factors $\Factors$ of a generative process and codes $\Codes$ representing states of an agent's generative model.},
    \item $\Actions$ is the action space,
    \item $\Transitions : \States \times \Actions \to \prob{\States}$ is the transitions dynamics, such that for a given state $\states$ and $\actions$, $\Transitions(\states, \actions)$ gives a probability distribution of states $\prob{\States}$ an agent can transition to from state $\states$ while taking action $\actions$, often written as $\prob{\states_{t+1} | \states_{t}, \actions_t}$,
    \item $\gamma \in [0,1)$ is called the discount factor,
    \item $\reward: \States \times \Actions \to \R$ is the reward function, giving a reward every time a transition is taken.
  \end{itemize} 
\end{definition}
Alternatively, it is also common to define a problem as a \textbf{partially observable Markov decision process (POMDP)}, where information of the environment is only indirectly available through some observations.
\begin{definition}[Partially observable Markov decision process (POMDP)]
  A partially observable Markov decision process is a tuple $(\States, \Actions, \Obs, \Transitions, \Observations, \discount, \reward)$, where $\States, \Actions, \Transitions, \discount, \reward$ follow the definition of an MDP and
  \begin{itemize}
    \item $\Obs$ is the observation space,
    \item $\Observations : \States \to \prob{\Obs}$ is the observation or measurement map.
  \end{itemize} 
\end{definition}

The goal of agents in an RL setup is to select sequences of actions that maximise expected cumulative discounted reward, also known as expected return, based on past and current experiences acquired through meaningful interactions with the environment.
Action policies representing sequences of actions are defined by the following
\begin{definition}[Action policy]
  Given an MDP $(\States, \Actions, \Transitions, \discount, \reward)$, a policy $\policy$ is defined as either
  \begin{itemize}
    \item a deterministic function $\policy: \States \to \Actions$, or
    \item a stochastic map $\policy: \States \to \prob{\Actions}$ assigning a distribution of actions to each state in $\States$.
  \end{itemize}
  For a POMDP $(\States, \Actions, \Obs, \Transitions, \Observations, \discount, \reward)$, a policy $\policy$ is usually defined instead as either
  \begin{itemize}
    \item a deterministic function $\policy: \beliefs \to \Actions$, or
    \item a stochastic map $\policy: \beliefs \to \prob{\Actions}$,
  \end{itemize}
  where \emph{beliefs} $\beliefs$, encoding probabilities on states given histories of observation-action sequences \(\stateaction \coloneqq [\Obs_{0}, \Actions_{0}, \dots, \Obs_{T}, \Actions_{T}]\), $\prob{\States | \stateaction}$, play the role of sufficient statistics of these histories. In the case of MDPs, these beliefs are trivialised by the full observability of the process and the Markov property, such that states observed $\States$ at a specific time step are sufficient statistics of histories of state-action sequences \(\stateaction \coloneqq [\states_{0}, \actions_{0}, \dots, \states_{T}, \actions_{T}]\).
  As we will see, in practice policies are often parameterised by the weights \(\policyparameters\) of a neural network that outputs the action probabilities after processing \(\states\), and in this case, they will be defined using the following notation: $\policy_{\policyparameters}$.
\end{definition}

For probabilistic state transitions and action policies, the expected cumulative discounted reward is defined as follows.
\begin{definition}[Expected cumulative discounted reward]
    Let $\States_\itertime \subseteq \States, \Actions_\itertime \subseteq \Actions$ be state and action subspaces of their respective spaces, indexed by time $\itertime \in \ntime$.
    We define time-indexed rewards $\Reward_\itertime \subseteq \mathbb{R}$ for $\itertime \in \ntime$ as
    \begin{align}
        \Reward_\itertime \coloneqq \reward(\States_\itertime, \Actions_\itertime),
    \end{align}
    which, in turn, can be used to define the cumulative discounted reward, or return $\return_{\itertime}$, indicated by
    \begin{align}
         \return_{\itertime} \coloneqq \sum_{\iterpartial= \itertime}^{\ntime} \gamma^{\iterpartial - \itertime} \Reward_{t}
         \label{eqn:return}
    \end{align}
    with \(i = 1\) for rewards over a full trajectory, and \(i > 1\) for rewards over a partial one, respectively (the former appears in the next equation, while the latter will be used in \cref{eqn:rl-actor-critic}).
    
    The \emph{expected} cumulative discounted reward is then defined as
    \begin{align}
        \expreturn(\policyparameters) = \mathbb{E}_{\stateaction \sim p_{\policy_{\policyparameters}}(\stateaction)} \left[ \return_1 \right]
        \label{eqn:exp-return}
    \end{align}
    with respect to a probability distribution over trajectories, \(p_{\policy_{\policyparameters}}(\stateaction)\).
\end{definition}

After introducing all the necessary, mathematical background for disentanglement and causality as treated in this work, in the next section we will look at how these notions have been described, implemented and studied in deep (reinforcement) learning.
This step will in particular help to bridge the gap between the formal, but somewhat simplified definition presented in this section (with agents modelled with, or without, disentangled representations and in a causal or simply a-causal way) and measures on degrees of disentanglement that can tell us \emph{how much} disentanglement and causal understanding can be found in different systems, i.e. how far subjects are from the ideal disentangled and causal representations introduced in this section.

\section{A computational framework for causal cognition}
\label{sec:computational}

\subsection{Explicitness as degrees of disentanglement}
\label{subsec:explicit-in-crl-agents}

\begin{figure}[ht]
    \centering
    \begin{tikzpicture}[auto, node distance = 3 cm, on grid, semithick,
        obs/.style = {circle, draw, black, text = black, minimum width = 0.5cm, fill=black!20},
        factor/.style = {circle, draw, black, fill=RedOrange, text = black, minimum width = 0.5cm},
        reconob/.style = {circle, draw, black, fill=Dandelion, text = black, minimum width = 0.5cm},
        box/.style args = {#1/#2}{draw, fill=#1, minimum size=#2},
        network/.style={trapezium, trapezium stretches=true, fill=#1!20, draw=#1!75, text=black}]
    
        {
          \node[rounded corners=0.5cm] (rect1) at (0,-2) [minimum width=1cm,minimum height=5cm, fill=black!10] {};
          \node[rounded corners=0.5cm] (rect2) at (5,-2) [minimum width=1cm,minimum height=2cm, fill=black!10] {};
          \node[rounded corners=0.5cm] (rect3) at (10,-2) [minimum width=1cm,minimum height=5cm, fill=black!10] {};
        }
         
        \node[obs] (Xj) at (0, -2) [draw=none, fill=black!10] {\(\Obs\)};
        
        \node[factor] (Zj) at (5, -2) [draw=none, fill=black!10] {\(\Codes\)};
        
        \node[reconob] (Xrj) at (10, -2) [draw=none, fill=black!10] {\(\Obs^{'}\)};
        
        
        \node [network=yellow, minimum width=3cm, minimum height=3cm, shape border rotate=270] at (2.5, -2) {\(q_{\encoderweights}(\codes|\obs)\)};
        \node [network=green, minimum width=3cm, minimum height=3cm, shape border rotate=90] at (7.5, -2) {\(p_{\decoderweights}(\obs|\codes)\)};
        
        
        \node (in) [above =3cm of Xj] {Observations};
        \node (lat) [above =1.5cm of Zj] {Codes};
        \node (out) [above =3cm of Xrj] {Reconstructions};
        \node (encoder) [left  =2.5cm of lat, yshift=5mm] {Encoder}; 
        \node (decoder) [right =2.5cm of lat, yshift=5mm] {Decoder};
    \end{tikzpicture}
    \caption{\textbf{Variational autoencoder}. An intuitive representation of a variational autoencoder, combining an encoder $q_{\encoderweights}(\codes|\obs)$ taking observations $\obs \in \Obs$ as inputs and producing $\codes \in \Codes$ as outputs, these outputs are then used by a decoder $p_{\decoderweights}(\obs|\codes)$ providing reconstructions of observations $\obs \in \Obs'$ with the goal of making them ``as close as possible'' to the original observations.}
    \label{fig:vae}
\end{figure}
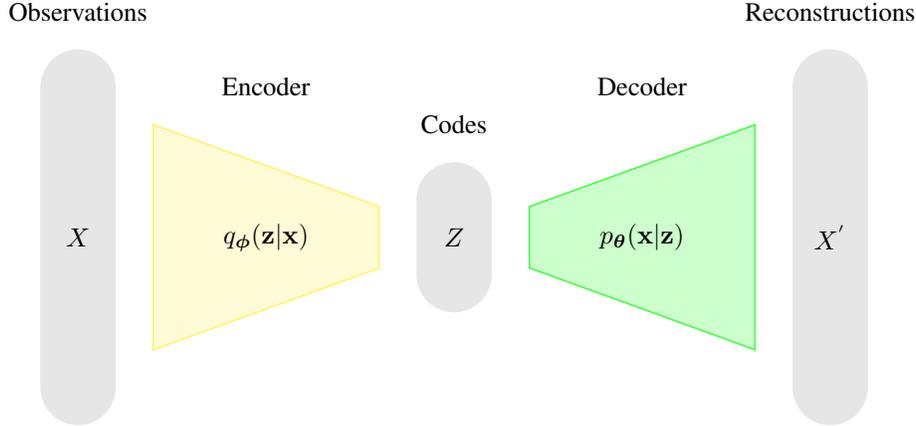

One of the main practical instantiations of disentanglement originates with models built on the architecture of \textbf{variational autoencoders (VAEs)} \parencite{Kingma2022, Kingma2019}, where a disentangled representation is defined as one where single latent units (codes) of a VAE are independently responsive to single factors generating observations \parencite{Higgins2017}.

Following the notation in \cref{subsec:disentanglementML}, the goal of a VAE is to learn, given a dataset \(\Data\) of observations \(\Obs\), a probabilistic generative model that can approximate factors \(\Factors\) using hidden variables (codes) \(\Codes\).
To do so, in a standard VAE architecture we find two neural networks, aptly named encoder and decoder, see \cref{fig:vae}.
An \textbf{encoder} with weights \(\encoderweights\) parameterises a distribution $Q^{\encoderweights}_{\Codes \mid \Obs}$, while a \textbf{decoder} network with weights \(\decoderweights\) parameterises a distribution $P^{\decoderweights}_{\Obs \mid \Codes}$.
The encoder plays the role of the function $\encoding$ in \cref{def:disentanglement}, mapping observations to codes, while the decoder is a clever construction that corresponds to a map
\begin{align}
  \decoding : \Codes \to \Obs'
\end{align}
where $\Obs'$ can be seen as \textbf{reconstructions} of $\Obs$, i.e. observations that should be ``as close as possible'' to the original ones according to some measure, in this case given by the VAE optimisation function provided below.
Notice that, since we only required $\encoding$ to be injective for $\generating(\Factors)$, $\decoding$ need not be a function and thus is not well-defined for the simple setup of sets and functions we adopted in \cref{subsec:disentanglementML}.
It is however a map that can easily be defined for more general setups \parencite{Zhang2023}\footnote{One could also just impose further restrictions on $\encoding$, for example making it surjective so that $\decoding$ becomes its right inverse. Alternatively, one could see $\decoding$ simply as a \emph{partial multi-valued} function and work in the category of sets and relations, or partial multi-valued functions, however this may not be the best choice for an introductory treatment of formal notions of disentanglement as the presence of monoidal (and not Cartesian) products complicates the definition of disentanglement \parencite{Zhang2023}.}.
Combining these, a VAE is trained to learn weights $(\encoderweights, \decoderweights)$ so as to maximise the \emph{evidence-lower bound} (ELBO):
\begin{align}
    \ELBO(\encoderweights, \decoderweights) = \mathbb{E}_{\codes \sim q_{\encoderweights}(\codes|\obs)} \Big[\log p_{\decoderweights}(\obs|\codes)\Big] - D_{KL}[q_{\encoderweights}(\codes|\obs) || p(\codes)] \\
    \label{eq:elbo}
\end{align}
via stochastic gradient descent using (batches of) observations sampled from a dataset \parencite{Doersch2021, Kingma2019}:
\begin{align}
  (\encoderweights, \decoderweights) \leftarrow (\encoderweights, \decoderweights) + \frac{1}{|\Data|} \sum_{\obs \in \Data} \nabla \ELBO(\encoderweights, \decoderweights).
\end{align}

Looking at the ELBO more closely, one can notice how the VAE is tasked with competing objectives.
Trying to maximise the first term, usually called the reconstruction loss, amounts to tweaking the weights of both the encoder and the decoder, $\encoderweights$ and $\decoderweights$ respectively, in such a way that the latent variables (codes $\Codes$) inferred by the former are more likely given a certain input (observations $\Obs$) and that the decoder can then use those variables to reconstruct it (reconstructed observations $\Obs'$).
In other words, weight values that lead to bad inferences and/or that do not afford a good reconstruction of the original inputs will be penalised.
The second term is negative (because the KL divergence, measuring closeness between distributions, is always greater than or equal to zero) so its maximisation tries to bring the divergence to zero, i.e. bringing the posterior distribution close to the prior.
This is asking the VAE to map the inputs to latent variables that are as close as possible to the given prior distribution.
Overall, the VAE should learn hidden variables, or codes, \(\codes\) that lead to good reconstructions \emph{and} whose probability can be brought close to the one indicated by the prior.

To improve the disentanglement performance of VAEs, one usually adds a hyperparameter \(\beta > 1\) to the KL divergence term of the original objective, leading to what is usually addressed as \(\beta\)-VAE \parencite{Higgins2017}.
Despite the empirical confirmation of disentanglement, however, the rationale for the $\beta$ tweak is not entirely clear. 
The intuition is that by putting more emphasis on learning statistically independent variables in the latent representation, one could also get a disentangled representation \parencite{Burgess2018, Higgins2017}.
Other proposals, based on the VAE, to improve disentanglement have seen the introduction of different architectures or regularisation terms to the optimisation objective \parencite{Kim2018b, Chen2019, Rubenstein2018a, Ridgeway2018, Eastwood2018}.
However, fundamental limitations remain because (1) without some supervision or crucial inductive bias disentangled representation learning cannot be achieved in practice \parencite{Locatello2019, Locatello2020, Trauble2021, Shu2020, Khemakhem2020} and (2) there is no general agreement on how to quantitatively measure disentanglement in the units of the latent representation \parencite{Sepliarskaia2021, Do2021, Wang2022}.

Some insights can nonetheless be obtained by looking at the optimisation objective of the \(\beta\)-VAE using the \textbf{information bottleneck principle}, a method to define a trade-off between compression and descriptiveness of a model given some data \parencite{Tishby2000, Tishby2015} with connections to fundamental laws of thermodynamics \parencite{Still2020, Daimer2023}.
More formally, \textcite{Alemi2017} shows that one can derive a version of the \(\beta\)-VAE objective from an unsupervised variation of the information bottleneck principle.
In this version, instead of having a mutual information term involving a latent representation of codes \(\Codes\) for labels in a dataset \(\Data\) to be maximised, we have one capturing how much information the latent representation encodes about a single given data point from the dataset, to be minimised because the generative model is supposed to learn a class of codes compatible with all the data points, not just one.
Using this principle, one can also understand the KL-divergence term \(D_{KL}[q_{\encoderweights}(\codes|\obs)||p(\codes)]\) modulated by $\beta$ \parencite{Higgins2017} as imposing an information bottleneck limiting the amount of information that can be encoded from the data into the latent representation, or in information theoretic terms: as limiting the channel capacity of the encoder \parencite{Burgess2018}.

As explicitly argued in \cref{subsec:from-disent-to-crl} treatments of causal representation learning can be regarded as attempts at learning a \emph{causal generative model} that identifies causal factorisations of a \emph{causal generative process}.
Depending on the goals and features of a specific implementation however, we will see that different architectures recover different kinds of latent representations.
To highlight what we believe to be the biggest difference, we will define two macro categories, of weak and strong disentanglement, based on whether a model can recover only factors and their relations to observables or factors and causal mechanisms involving observables and other factors of a generative process, respectively.

\begin{figure}[ht!]
	\centering
	\scalebox{.68}{\begin{tikzpicture}
	\begin{pgfonlayer}{nodelayer}
		\node [style=none] (11) at (0, 13) {\Large{Generative process}};
		\node [style=none] (20) at (14, 0) {};
		\node [style=none] (21) at (-14, 0) {};
		\node [style=none] (22) at (0, 12) {};
		\node [style=none] (23) at (0, -11) {};
		\node [style=node conf] (28) at (4.75, 10) {$\Confounders$};
		\node [style=none] (71) at (-7.25, -9.25) {\large{\textcolor{RoyalBlue}{(structurally exact)}}};
		\node [style=none] (72) at (7.25, -9.25) {\large{\textcolor{RoyalBlue}{(structurally approximate)}}};
		\node [style=node factors] (73) at (-12, 6.5) {$\Factors_1$};
		\node [style=node factors] (74) at (-8.75, 6.5) {$\Factors_2$};
		\node [style=node factors] (75) at (-5.5, 6.5) {$\Factors_3$};
		\node [style=node factors] (76) at (-2.5, 6.5) {$\Factors_4$};
		\node [style=node obs] (77) at (-7, 2.5) {$\Obs$};
		\node [style=node codes] (78) at (-12, -6.5) {$\Codes_1$};
		\node [style=node codes] (79) at (-8.75, -6.5) {$\Codes_2$};
		\node [style=node codes] (80) at (-5.5, -6.5) {$\Codes_3$};
		\node [style=node codes] (81) at (-2.5, -6.5) {$\Codes_4$};
		\node [style=node obs] (82) at (-7, -2.5) {$\Obs'$};
		\node [style=node factors] (83) at (2.5, 6.5) {$\Factors_1$};
		\node [style=node factors] (84) at (5.75, 6.5) {$\Factors_2$};
		\node [style=node factors] (85) at (9, 6.5) {$\Factors_3$};
		\node [style=node factors] (86) at (12, 6.5) {$\Factors_4$};
		\node [style=node obs] (87) at (7.5, 2.5) {$\Obs$};
		\node [style=node codes] (88) at (2.5, -6.5) {$\Codes_{1,2}$};
		\node [style=node codes] (90) at (9, -6.5) {$\Codes_3$};
		\node [style=node codes] (91) at (12, -6.5) {$\Codes_4$};
		\node [style=node obs] (92) at (7.5, -2.5) {$\Obs'$};
		\node [style=node factors] (93) at (-18.5, -18) {};
		\node [style=node codes] (94) at (-18.5, -21) {};
		\node [style=node conf] (95) at (-10.5, -18) {};
		\node [style=node obs] (96) at (-10.5, -21) {};
		\node [style=none] (97) at (-3, -20.75) {};
		\node [style=none] (98) at (0, -20.75) {};
		\node [style=none] (99) at (-16, -18) {Factors};
		\node [style=none] (100) at (-16, -21) {Codes};
		\node [style=none] (101) at (-7, -18) {Confounders};
		\node [style=none] (102) at (-7, -21) {Observations};
		\node [style=none] (103) at (-0.25, -16) {Causal mechanisms:};
		\node [style=none] (104) at (4.25, -20.25) {from factors};
		\node [style=none] (105) at (4.25, -21) {to other factors};
		\node [style=none] (106) at (-18.25, -14.75) {\large{\textcolor{red}{Legend}}};
		\node [style=none] (107) at (-4, -17) {};
		\node [style=none] (108) at (-4, -22.25) {};
		\node [style=none] (109) at (-3, -18.5) {};
		\node [style=none] (110) at (0, -18.5) {};
		\node [style=none] (111) at (4.25, -18) {from confouders};
		\node [style=none] (112) at (4.25, -18.75) {to factors};
		\node [style=none] (113) at (9, -18.5) {};
		\node [style=none] (114) at (12, -18.5) {};
		\node [style=none] (115) at (16.25, -18) {from factors/codes};
		\node [style=none] (116) at (16.25, -18.75) {to observations};
		\node [style=none] (117) at (9, -20.75) {};
		\node [style=none] (118) at (12, -20.75) {};
		\node [style=none] (119) at (16.25, -20.25) {from codes};
		\node [style=none] (120) at (16.25, -21) {to other codes};
		\node [style=none] (121) at (0, -12) {\Large{Generative model}};
		\node [style=none] (122) at (0, 16) {{\Huge Weak disentanglement}};
	\end{pgfonlayer}
	\begin{pgfonlayer}{edgelayer}
		\draw [style=big dashes] (21.center) to (20.center);
		\draw [style=big dashes] (23.center) to (22.center);
		\draw [style=thick-arrow-black] (74) to (77);
		\draw [style=thick-arrow-black] (75) to (77);
		\draw [style=thick-arrow-black] (76) to (77);
		\draw [style=thick-arrow-black] (73) to (77);
		\draw [style=thick-arrow-black] (79) to (82);
		\draw [style=thick-arrow-black] (80) to (82);
		\draw [style=thick-arrow-black] (81) to (82);
		\draw [style=thick-arrow-black] (78) to (82);
		\draw [style=thick-arrow-black] (84) to (87);
		\draw [style=thick-arrow-black] (85) to (87);
		\draw [style=thick-arrow-black] (86) to (87);
		\draw [style=thick-arrow-black] (83) to (87);
		\draw [style=thick-arrow-blue] (28) to (83);
		\draw [style=thick-arrow-black] (90) to (92);
		\draw [style=thick-arrow-black] (91) to (92);
		\draw [style=thick-arrow-black] (88) to (92);
		\draw [style=thick-arrow-blue] (28) to (84);
		\draw [style=thick-arrow-orange] (97.center) to (98.center);
		\draw [style=big dashes] (108.center) to (107.center);
		\draw [style=thick-arrow-blue] (109.center) to (110.center);
		\draw [style=thick-arrow-black] (113.center) to (114.center);
		\draw [style=thick-arrow-teal] (117.center) to (118.center);
	\end{pgfonlayer}
\end{tikzpicture}}
        \caption{\textbf{Weak disentanglement}. Weak disentanglement as a variant of causal representation learning concerned only with learning causal codes $\Codes$ from high-dimensional inputs (observations \(\Obs\)) generated by causal factors $\Factors$, and the identification of causal mechanisms that map causal factors to observations (no causal mechanisms between causal codes). A generative model can be considered structurally approximate (with respect to the assumed generative process) if it fails to recover completely disentangled codes (some codes remain entangled, e.g. the code $\Codes_{1,2}$ suggestively standing in for two factors $\Factors_1, \Factors_2$) and/or if it leaves out causal mechanisms relating codes to observations. Note that in an actual implementation, like the VAE, entanglement might manifest as correlations among two or more codes in the latent representation, all capturing the same factors at the same time. In other words, the node $\Codes_{1,2}$ may in practice represent a groups of nodes with possibly bidirectional influences amongst each other.}
        \label{fig:weakDisentanglement}
\end{figure}
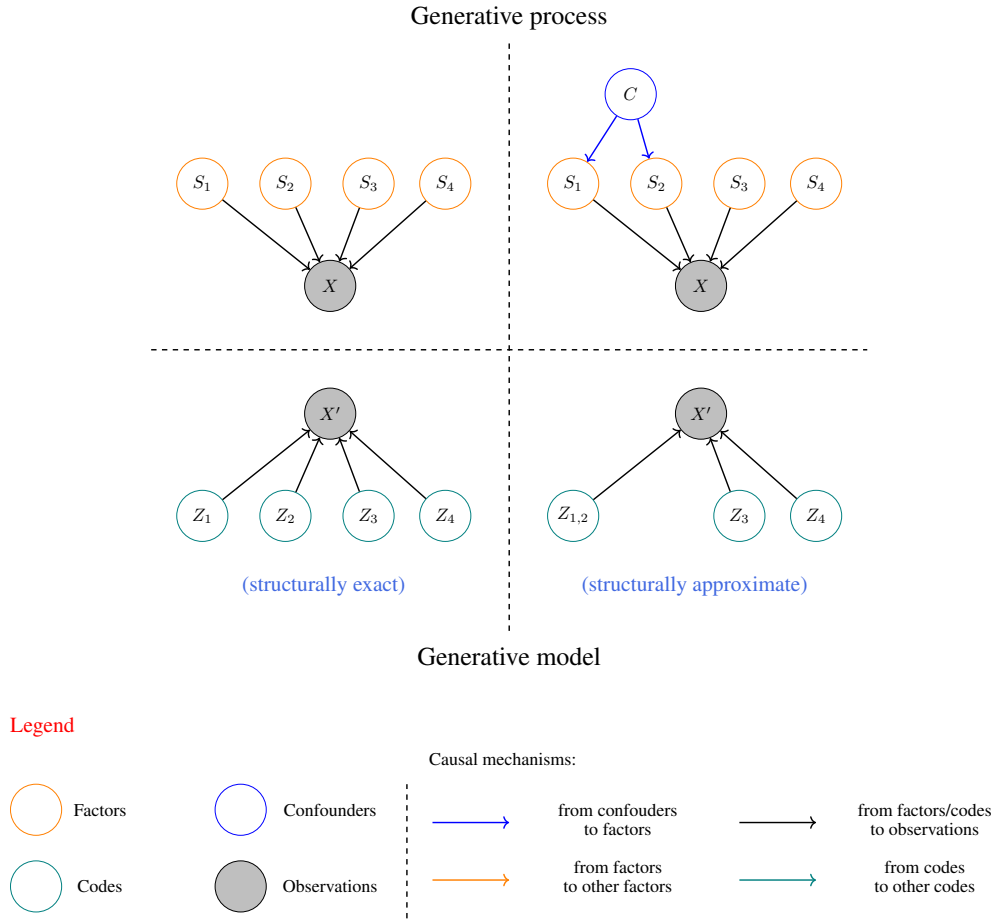

\begin{definition}[\textbf{Weak disentanglement}]
    Weak disentanglement captures the idea of learning a mapping between observations $\Obs$ and causal codes $\Codes$, without requiring that the causal relationships among the factors (potentially involving confounders $\Confounders$) are also recovered (see \cref{def:scm}).
    This approach appears for example in \textcite{Suter2019}, where confounders in the generative process cannot be encoded by the generative model, and there are no causal mechanisms between causal codes or between confounders and causal codes, thus rendering all assignments $\assignment_j$ trivial (i.e. identities) (cf. \cref{eqn:parents}), see \cref{fig:weakDisentanglement}:
    \begin{align}
        \PA_j \subset \{\}, \qquad j \in \{1, \dots, \nfactors\}.
    \end{align}
\end{definition}

In weak disentanglement approaches, causal factors are conceived of as elementary ingredients that independently influence the observations through the mapping \(\generating\)\footnote{If confounders are not present (or not considered), see left panel in \cref{fig:weakDisentanglement}, then the problem of weak disentanglement resembles that of independent component analysis (ICA) \parencite{Hyvarinen2001}, for connections between ICA and causal representation learning, see \textcite{Hyvarinen2013, Gresele2021, Wendong2023}.}.
The goal of an agent is then to learn a causal generative model that reflects this scenario, e.g. by relying on a measure of robustness with respect to interventions \parencite{Suter2019} or by enforcing an independence (orthogonality) constraint on the Jacobian of the appropriate version of the \(\generating\) map, whose elements quantify the influence of each causal factor on observations from the dataset $\Data$ \parencite{Gresele2021}.
Further, if observations provide information only about a subset of the causal factors (partial observability), thereby introducing potential confounders, a sparsity constraint imposed on the latent representation can help to recover the ground-truth causal factors \parencite{Xu2024}.
Despite the absence of any reference to causality, practical examples of disentangled representation learning illustrated by means of the \(\beta\)-VAE earlier achieve something similar insofar as one prepares a dataset with known causal factors and usually shows that their method can recover all those factors in the latent representation (in this case the correspondence is expected to be one-to-one, in contrast with \textcite{Suter2019} where it is assumed to be one-to-many).

\begin{definition}[\textbf{Strong disentanglement}]
    Strong disentanglement aims to recover not only the causal factors in the latent representation, but also confounders and the causal relationships present among them all (i.e. the causal mechanisms).
    In other words, in strong disentanglement we consider the more general scenario described in \cref{subsec:from-disent-to-crl} where the causal relations derived from the sets of parents $\PA_j$ of each variable are not assumed to be only confounders (see 
    \begin{align}
      \PA_{j} \subset \{\Factors_{\setminus j}, \Confounders\} \quad (= \cref{eqn:parents}).
    \end{align}
\end{definition}

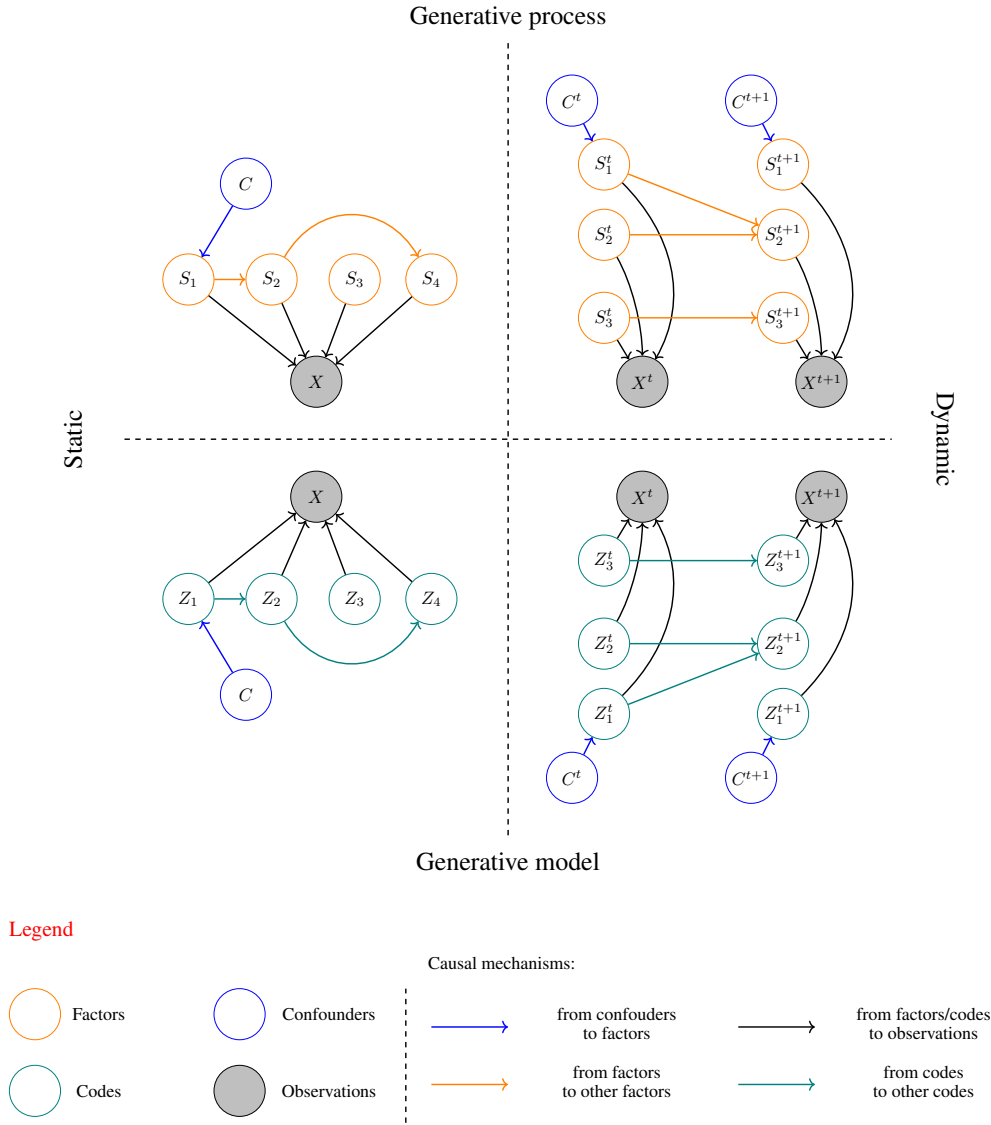
\begin{figure}[ht!]
	\centering
	\scalebox{.68}{\begin{tikzpicture}
	\begin{pgfonlayer}{nodelayer}
		\node [style=none] (6) at (0, 16.5) {\Large{Generative process}};
		\node [style=none] (8) at (15, 0) {};
		\node [style=none] (9) at (-15, 0) {};
		\node [style=none] (10) at (0, 15.5) {};
		\node [style=none] (11) at (0, -15.5) {};
		\node [style=node factors] (12) at (-12.5, 6.25) {$\Factors_1$};
		\node [style=node factors] (13) at (-9.25, 6.25) {$\Factors_2$};
		\node [style=node factors] (14) at (-6, 6.25) {$\Factors_3$};
		\node [style=node factors] (15) at (-3, 6.25) {$\Factors_4$};
		\node [style=node conf] (16) at (-10.25, 10) {$\Confounders$};
		\node [style=node obs] (17) at (-7.5, 2.25) {$\Obs$};
		\node [style={rotate +90}] (33) at (-17, 0) {\Large{Static}};
		\node [style=rotate -90] (34) at (17, 0) {\Large{Dynamic}};
		\node [style=node factors] (35) at (-18.5, -22.5) {};
		\node [style=node codes] (36) at (-18.5, -25.5) {};
		\node [style=node conf] (37) at (-10.5, -22.5) {};
		\node [style=node obs] (38) at (-10.5, -25.5) {};
		\node [style=none] (41) at (-3, -25.25) {};
		\node [style=none] (42) at (0, -25.25) {};
		\node [style=none] (45) at (-16, -22.5) {Factors};
		\node [style=none] (46) at (-16, -25.5) {Codes};
		\node [style=none] (47) at (-7, -22.5) {Confounders};
		\node [style=none] (48) at (-7, -25.5) {Observations};
		\node [style=none] (49) at (-0.25, -20.5) {Causal mechanisms:};
		\node [style=none] (52) at (4.25, -24.75) {from factors};
		\node [style=none] (53) at (4.25, -25.5) {to other factors};
		\node [style=none] (56) at (-18.25, -19.25) {\large{\textcolor{red}{Legend}}};
		\node [style=none] (57) at (-4, -21.5) {};
		\node [style=none] (58) at (-4, -26.75) {};
		\node [style=node codes] (59) at (-12.5, -6.25) {$\Codes_1$};
		\node [style=node codes] (60) at (-9.25, -6.25) {$\Codes_2$};
		\node [style=node codes] (61) at (-6, -6.25) {$\Codes_3$};
		\node [style=node codes] (62) at (-3, -6.25) {$\Codes_4$};
		\node [style=node conf] (63) at (-10.25, -10) {$\Confounders$};
		\node [style=node obs] (64) at (-7.5, -2.25) {$\Obs$};
		\node [style=node factors] (67) at (3.75, 10.75) {$\Factors_1^t$};
		\node [style=node factors] (68) at (3.75, 8) {$\Factors_2^t$};
		\node [style=node factors] (69) at (3.75, 4.75) {$\Factors_3^t$};
		\node [style=node conf] (70) at (2.5, 13.25) {$\Confounders^t$};
		\node [style=node obs] (71) at (5.25, 2.25) {$\Obs^t$};
		\node [style=node factors] (73) at (10.75, 10.75) {$\Factors_1^{t+1}$};
		\node [style=node factors] (74) at (10.75, 8) {$\Factors_2^{t+1}$};
		\node [style=node factors] (75) at (10.75, 4.75) {$\Factors_3^{t+1}$};
		\node [style=node conf] (76) at (9.5, 13.25) {$\Confounders^{t+1}$};
		\node [style=node obs] (77) at (12.25, 2.25) {$\Obs^{t+1}$};
		\node [style=node codes] (81) at (3.75, -10.75) {$\Codes_1^t$};
		\node [style=node codes] (82) at (3.75, -8) {$\Codes_2^t$};
		\node [style=node codes] (83) at (3.75, -4.75) {$\Codes_3^t$};
		\node [style=node conf] (84) at (2.5, -13.25) {$\Confounders^t$};
		\node [style=node obs] (85) at (5.25, -2.25) {$\Obs^t$};
		\node [style=node codes] (87) at (10.75, -10.75) {$\Codes_1^{t+1}$};
		\node [style=node codes] (88) at (10.75, -8) {$\Codes_2^{t+1}$};
		\node [style=node codes] (89) at (10.75, -4.75) {$\Codes_3^{t+1}$};
		\node [style=node conf] (90) at (9.5, -13.25) {$\Confounders^{t+1}$};
		\node [style=node obs] (91) at (12.25, -2.25) {$\Obs^{t+1}$};
		\node [style=none] (93) at (-3, -23) {};
		\node [style=none] (94) at (0, -23) {};
		\node [style=none] (97) at (4.25, -22.5) {from confouders};
		\node [style=none] (98) at (4.25, -23.25) {to factors};
		\node [style=none] (101) at (9, -23) {};
		\node [style=none] (102) at (12, -23) {};
		\node [style=none] (103) at (16.25, -22.5) {from factors/codes};
		\node [style=none] (104) at (16.25, -23.25) {to observations};
		\node [style=none] (108) at (9, -25.25) {};
		\node [style=none] (109) at (12, -25.25) {};
		\node [style=none] (110) at (16.25, -24.75) {from codes};
		\node [style=none] (111) at (16.25, -25.5) {to other codes};
		\node [style=none] (116) at (0, -16.5) {\Large{Generative model}};
		\node [style=none] (117) at (0, 19.5) {{\Huge Strong disentanglement}};
	\end{pgfonlayer}
	\begin{pgfonlayer}{edgelayer}
		\draw [style=big dashes] (9.center) to (8.center);
		\draw [style=big dashes] (11.center) to (10.center);
		\draw [style=thick-arrow-blue] (16) to (12);
		\draw [style=thick-arrow-black] (13) to (17);
		\draw [style=thick-arrow-orange] (12) to (13);
		\draw [style=thick-arrow-orange] (41.center) to (42.center);
		\draw [style=big dashes] (58.center) to (57.center);
		\draw [style=thick-arrow-black] (14) to (17);
		\draw [style=thick-arrow-black] (15) to (17);
		\draw [style=thick-arrow-blue] (63) to (59);
		\draw [style=thick-arrow-black] (60) to (64);
		\draw [style=thick-arrow-teal] (59) to (60);
		\draw [style=thick-arrow-black] (61) to (64);
		\draw [style=thick-arrow-black] (62) to (64);
		\draw [style=thick-arrow-black, in=60, out=-45] (67) to (71);
		\draw [style=thick-arrow-black, in=90, out=-60, looseness=0.75] (68) to (71);
		\draw [style=thick-arrow-black] (69) to (71);
		\draw [style=thick-arrow-blue] (70) to (67);
		\draw [style=thick-arrow-black, in=60, out=-45] (73) to (77);
		\draw [style=thick-arrow-black, in=90, out=-60, looseness=0.75] (74) to (77);
		\draw [style=thick-arrow-black] (75) to (77);
		\draw [style=thick-arrow-blue] (76) to (73);
		\draw [style=thick-arrow-orange] (67) to (74);
		\draw [style=thick-arrow-orange] (68) to (74);
		\draw [style=thick-arrow-orange] (69) to (75);
		\draw [style=thick-arrow-black, in=-60, out=45] (81) to (85);
		\draw [style=thick-arrow-black, in=-90, out=60, looseness=0.75] (82) to (85);
		\draw [style=thick-arrow-black] (83) to (85);
		\draw [style=thick-arrow-blue] (84) to (81);
		\draw [style=thick-arrow-black, in=-60, out=45] (87) to (91);
		\draw [style=thick-arrow-black, in=-90, out=60, looseness=0.75] (88) to (91);
		\draw [style=thick-arrow-black] (89) to (91);
		\draw [style=thick-arrow-blue] (90) to (87);
		\draw [style=thick-arrow-teal] (81) to (88);
		\draw [style=thick-arrow-teal] (82) to (88);
		\draw [style=thick-arrow-teal] (83) to (89);
		\draw [style=thick-arrow-blue] (93.center) to (94.center);
		\draw [style=thick-arrow-black] (101.center) to (102.center);
		\draw [style=thick-arrow-teal] (108.center) to (109.center);
		\draw [style=thick-arrow-black] (12) to (17);
		\draw [style=thick-arrow-black] (59) to (64);
		\draw [style=thick-arrow-orange, bend left=60, looseness=1.25] (13) to (15);
		\draw [style=thick-arrow-teal, bend right=60, looseness=1.25] (60) to (62);
	\end{pgfonlayer}
\end{tikzpicture}}
    \caption{\textbf{Strong disentanglement}. Strong disentanglement as a variant of causal representation learning concerned with learning causal codes and causal mechanisms of different kinds, to observations and to other causal codes.
    In a static setting (left panel), strong disentanglement involves discovering the causal mechanism among causal factors and, potentially, the structural assignments of the underlying structural causal model. 
    In a dynamic setting (right panel), strong disentanglement denotes finding causal structure in the transition dynamics (e.g. of an MDP), determining causal mechanisms between causal factors at different time steps. Importantly, we don't include causal mechanisms within the same time slice as we assume that there can be no instantaneous interactions among factors. Note that even when we include dynamics there is still an underlying SCM that can be partitioned into sets of causal factors ``over time'', i.e. causal factors are process extended over time, or sets of variables (in this case conveniently given the same name but a different time index) belonging to different time steps. Intuitively, in the dynamics setting more causal structure is uncovered because one is considering causal mechanisms over more causal factors (assigned to different time slices).}
    \label{fig:strongDisentanglement}
\end{figure}

This is implemented for instance by the CausalVAE architecture \parencite{Yang2021a}, where a modified VAE is augmented with a mask layer, in the form of an adjacency matrix, trained and applied to the latent vector of codes $\Codes$ to implement the structural assignments defining the causal mechanism among the codes (effectively, this step amounts to sampling from a structural causal model), see \cref{fig:strongDisentanglement}.
The vanilla CausalVAE architecture is however limited to static data, e.g. images, and does not thus lend itself to treatments of causality for dynamical processes where causal factors can be causally related in time, typical for example of reinforcement learning treatments of decision making in agents.
To tackle this, building on the CausalVAE and others similar works, \textcite{Yao2022a} proposed an extension handling temporal data, see \cref{fig:strongDisentanglement}.
Their VAE-based architecture is more elaborate, and designed 1) to encode and decode features of objects (the causal factors) from observations using convolutional neural networks, CNNs, 2) to integrate temporal information in the latent representation with gated recurrent units, GRUs, 3) to infer the causal codes while enforcing constraints on their latent causal dynamics (e.g. that the noise terms are mutually independent or follow a particular distribution) to ensure that the causal generative process is identifiable from data (see \cref{sec:causalRL}).

At a high level, differences between the two approaches can be described in terms of causal understanding of properties vs objects:

\begin{itemize}
  \item models with weak disentanglement can essentially account for \emph{properties}, as in the red bouncing ball example where these would colour, shape and material, such properties are by definition independent as we wouldn't want or expect them to posses causal relationships among them (if they did, they would not be different properties), while
  \item models of strong disentanglement aim to describe \emph{objects}, in the red bouncing ball example these could be wind, the ground and a player, and unlike for properties, for objects we would also want a model to capture causal relations among them, in a way that makes them noticeable and relevant for an interacting agent.
\end{itemize}

Overall, in the spectrum of models implementing disentanglement, from weak to strong, a causal representation that only identifies causal factors and their relations to observations can be said to be less explicit than one also recovering the causal mechanisms that exist among them.
The latter is in turn less explicit than a causal representation that identifies the functional form of the causal mechanism, etc.
In other words, a more fine-grained and detailed causal representation encodes more causal information, thereby describing more explicitly the causal structure of a certain domain.
This is in line with the informal depiction of explicitness offered by \cref{subsec:explicit-ci}, where the notion was related to the idea of having more causal information to operate in a certain context, e.g. knowing what means can be manipulated to reach a certain goal, and will be used hereafter as the foundation of our proposal to relate work on causality in animal cognition and deep (reinforcement) learning (see \cref{subsec:explicitness-causal-rep}).

\subsection{Trajectories as sources of causal information in RL}
\label{subsec:trajectories-causal-rl}

In \textbf{online learning} an agent uses the current policy to perform an action in the (training) environment, which responds with a reward signal, at each time step.
Trajectories of state-action \(\stateaction \coloneqq [\states_{0}, \actions_{0}, \dots, \states_{T}, \actions_{T}]\) or observation-action sequences \(\stateaction \coloneqq [\obs_{0}, \actions_{0}, \dots, \obs_{T}, \actions_{T}]\), often called histories, see \cref{sec:causalRL}, can be stored in a \textbf{replay buffer}\footnote{This procedure is also known as \emph{experience replay} and was part and parcel of one of the first breakthroughs of deep reinforcement learning, see \textcite{Mnih2015}.} as sequences of tuples together with their respective rewards at each time step, e.g, \((\states_{t}, \actions_{t}, \reward_{t}, \states_{t+1})\) or \((\obs_{t}, \actions_{t}, \reward_{t}, \obs_{t+1})\), where one sequence corresponds to a trajectory.
This information forms an agent's experience: the state the agent was in, \(\states_{t}\), the action it performed, \(\actions_{t}\), the reward it collected, \(\reward_{t}\), and the next state \(\states_{t+1}\) reached from \(\states_{t}\) by performing \(\actions_{t}\) \parencite{Lin1991, Mnih2015, Mnih2016}.

Concretely, this experience can be used to obtain an estimate of the expected cumulative discounted reward in \cref{eqn:exp-return} based on trajectories sampled from the replay buffer, which is regularly updated and acts as a rudimentary database of memories for the agent.
To see this idea in action, we look at a popular class of approaches represented by \textbf{actor-critic} methods for which the approximate gradient of the RL objective (see \cref{eqn:exp-return}), used to update policy parameters $\policyparameters$, is computed as follows \parencite{Sutton2018}:

\begin{align}
    \label{eqn:rl-actor-critic}
    \nabla_\policyparameters \estexpreturn(\policyparameters) = \frac{1}{\ntrajectories} \sum_{\itertrajectories=1}^{\ntrajectories} \sum_{\itertime=1}^{\ntime} \nabla_{\policyparameters} \log \policy_{\policyparameters}(\actions_{\itertrajectories, \itertime}|\states_{\itertrajectories, \itertime}) \biggl( \sum_{\iterpartial= \itertime}^{\ntime} \gamma^{\iterpartial - \itertime} \reward(\states_{\itertrajectories, \iterpartial}, \actions_{\itertrajectories, \iterpartial}) - \valuefunction^{\policy_{\policyparameters}}(\states_{\itertrajectories, \iterpartial}) \biggr) 
\end{align}

where $\ntrajectories$ is the number of trajectories sampled from the replay buffer; \(\policy_{\policyparameters}\) is the current policy whose parameters \(\policyparameters\) will be updated with the computed gradient; $\sum_{\iterpartial= \itertime}^{\ntime} \gamma^{\iterpartial - \itertime} \reward(\states_{\itertrajectories, \iterpartial}, \actions_{\itertrajectories, \iterpartial})$ is the discounted sum of rewards ($\return_{\itertime}$, cf. \cref{eqn:return}) evaluated for the (partial) sampled trajectory $\itertrajectories$ acquired from time step \(\itertime\) until the terminal state $\ntime$, and $\valuefunction^{\policy_\policyparameters}$ is a (state) value function that acts as a baseline with respect to the discounted return, defined as the expected return $\expreturn(\policyparameters)$ from a chosen state sampled from trajectory $\itertrajectories$ at time $\iterpartial$, $\states_{\itertrajectories, \iterpartial}$, if a policy \(\policy_\policyparameters\) is followed from that point onwards, i.e.

\begin{align}
    \valuefunction^{\policy_{\policyparameters}}(\states_{\itertrajectories, \iterpartial}) \coloneqq \mathbb{E}_{\policy_{\policyparameters}} [\return_{\itertime} | \States_\itertime = \states_{\itertrajectories, \iterpartial}].
\end{align}

In \cref{eqn:rl-actor-critic}, it is common \parencite{Sutton2018, Schulman2018, Gu2017, Degris2012} to approximate the discounted return \(\sum_{\iterpartial= \itertime}^{\ntime} \gamma^{\iterpartial - \itertime} \reward(\states_{\itertrajectories, \iterpartial}, \actions_{\itertrajectories, \iterpartial})\), the sum of rewards obtained from a particular, realised trajectory $\itertrajectories$, with

\begin{align}
    \Qfunction^{\policy_{\policyparameters}}(\states_{\itertrajectories, \iterpartial}, \actions_{\itertrajectories, \iterpartial}) \coloneqq \mathbb{E}_{\policy_{\policyparameters}} [\return_{\itertime} | \States_\itertime = \states_{\itertrajectories, \iterpartial}, \Actions_\itertime = \actions_{\itertrajectories, \iterpartial}].
\end{align}

This is the \(\Qfunction\)-function (or action value function) for the policy under consideration, quantifying the value of performing an action in a certain state, after which the policy is followed until the end of the episode.
With this substitution, one defines the advantage $\Advantage^{\policy_{\policyparameters}}$, specifying how much better it is to take action an action $\actions_{\itertrajectories, \iterpartial}$ as opposed to an average action\footnote{This can be seen by writing down explicitly the relationship between $\valuefunction$ and $\Qfunction$, in our case
\begin{align}
    \valuefunction^{\policy_{\policyparameters}}(\states_{\itertrajectories, \iterpartial}) = \mathbb{E}_{\actions \sim \policy_{\policyparameters}} \left[\Qfunction^{\policy_{\policyparameters}}(\states_{\itertrajectories, \iterpartial}, \actions_{\itertrajectories, \iterpartial}) \right]
\end{align}
}

\begin{align}
    \Advantage^{\policy_{\policyparameters}}(\states_{\itertrajectories, \iterpartial}, \actions_{\itertrajectories, \iterpartial}) \coloneqq \Qfunction^{\policy_{\policyparameters}}(\states_{\itertrajectories, \iterpartial}, \actions_{\itertrajectories, \iterpartial}) - \valuefunction^{\policy_{\policyparameters}}(\states_{\itertrajectories, \iterpartial})
    \label{eqn:advantage}
\end{align}

that is approximated using a \emph{critic} neural network trained to estimate only the state value function from the reward (because the \(\Qfunction\)-function can be rewritten as the sum of the reward at the current state and the expected state value function at the next, i.e. $\Qfunction^{\policy_{\policyparameters}}(\states_{\itertrajectories, \iterpartial}, \actions_{\itertrajectories, \iterpartial}) = \reward(\states_{\itertrajectories, \iterpartial}, \actions_{\itertrajectories, \iterpartial}) + \mathbb{E}_{\states_{\itertrajectories, \iterpartial + 1} \sim \prob{\states_{\itertrajectories, \iterpartial + 1} | \states_{\itertrajectories, \iterpartial}, \actions_{\itertrajectories, \iterpartial}}} \left[\valuefunction^{\policy_{\policyparameters}}(\states_{\itertrajectories, \iterpartial + 1})\right]$) \parencite{Sutton2018, Schulman2018, Gu2017, Degris2012}.
The \emph{actor} part is represented instead by the policy $\policy_\policyparameters$, parameterised by a policy neural network that outputs the most suitable action given a certain state. 
A learning step then involves sampling a batch of trajectories from the buffer, using them to evaluate \(\nabla_\policyparameters \estexpreturn(\policyparameters)\), i.e. the gradient of the estimated objective with respect to the policy parameters, and updating these parameters, $\policyparameters$, using the gradient to derive a better policy, conducive to expected cumulative reward maximisation.

Importantly, the learning problem becomes \textbf{off-policy} if the sampled trajectories used to compute the gradients are not collected by the current policy (as implemented by the policy network at the current time step) but by a different one (implemented, for instance, by an old parameters configuration of the policy network).
In practical situations, this is often the case because the replay buffer does not store only the most recent trajectory, collected by the current policy, but also past trajectories.
Therefore, sampling a batch of them from the replay buffer turn the learning problem into an off-policy one.
Similarly, in an imitation learning scenario, policy optimisation is also by default off-policy because the gradients are computed using sampled trajectories that come from an expert demonstrator and not from the policy currently followed by the agent (see \cref{subsubsec:social-causal-info-examples} for a more detailed overview on imitation learning).
In an off-policy setting, estimating the gradient of the objective considered above, $\estexpreturn(\policyparameters)$, is problematic because the parameters of the current policy could be updated based on actions and/or reward information (e.g. value functions) that in reality characterise a different/earlier actor.
In other words, the gradient information in $\estexpreturn(\policyparameters)$ might be inaccurate for updating the current policy.
Corrections can be applied to the gradient depending on the exact RL method considered, going from a variety of importance sampling techniques (for policy gradient methods, see \cref{subsec:combining-sources-rl}) to the use of appropriate value functions, i.e. using the actions of the current actor (in actor critic methods).

\subsection{Combining different sources of causal information in RL}
\label{subsec:combining-sources-rl}

Following the idea of a buffer containing stored trajectories representing experience to update a value function or policy, integration can be interpreted as the ability of an agent to combine different kinds of experience into its own decision-making process, appropriately weighted based on context, task demands, origin, resources, etc.
In principle, these kinds of experience can include single-source causal information, e.g. when an agent uses experience acquired at different points in time or in different environments/tasks, such as in multi-task or meta-RL (see \textcite{Beck2023}).
However, in this context we focus on integration of different sources (see \cref{subsec:sources-causal-info}), with the goal of highlighting synergistic forms of causal understanding that truly take advantage of the amalgamation of different causal perspectives.
This allows us thus to delineate an operational, computational account of the notional idea of integration presented in \textcite{Starzak2021}.

In computational terms, the question of how best to integrate and use information coming from different sources is a foundational aspect of \textbf{offline RL}, where the replay buffer can store trajectory data collected by any policy in a variety of virtual environments, more or less realistic, or from real-world tasks.
For example, recent datasets for offline RL tend to include trajectories from experts (e.g. hand-designed controllers or human demonstrators), from other RL agents trained online in a certain domain, from the same agent operating in the same environment but performing slightly different tasks (multi-task, past experience), from unsupervised (i.e. reward-free) exploratory policies \parencite{Fu2021, Gulcehre2020, Yarats2022, Zhou2023}.

The challenge of designing an offline RL algorithm is precisely that of exploiting the collected data in such a way that the learned policy can be safely applied to a given environment.
This means that the learning algorithm has to acquire and integrate causal information from various (PO)MDPs (those in the training set) in such a way that the most appropriate actions for new downstream tasks/environments can be extrapolated successfully from past experience and generalised into unfamiliar contexts.
A vanilla approach consists of using \textbf{importance sampling}, originally tailored for dealing with off-policy data (see previous section) \parencite{Kahn1955, Peshkin2002, Precup2000, Koller2009, Jie2010, Sutton2018}.

In this context, importance sampling corresponds to the introduction of importance weights, ratios (computed over a trajectory) between the current policy to be optimised, \(\policy_{\policyparameters}\), and a behavioural policy, \(\policy_{\behavioural}\), used to collect the transitions sampled from the replay buffer

\begin{align}
    \importanceweights = \prod_{\itertime=1}^{\ntime} \frac{\policy_{\policyparameters}(\actions_{\iterpartial, \itertime}|\states_{\iterpartial, \itertime})}{\policy_{\behavioural}(\actions_{\iterpartial, \itertime}|\states_{\iterpartial, \itertime})}
\end{align}

These weights are then used in \cref{eqn:rl-actor-critic}, obtaining:

\begin{align}
    \nabla_\policyparameters \estexpreturn(\policyparameters) = \frac{1}{\ntrajectories} \sum_{\itertrajectories=1}^{\ntrajectories} \importanceweights \sum_{\itertime=1}^{\ntime} \nabla_{\policyparameters} \log \policy_{\policyparameters}(\actions_{\itertrajectories, \itertime}|\states_{\itertrajectories, \itertime}) \biggl( \sum_{\iterpartial= \itertime}^{\ntime} \gamma^{\iterpartial - \itertime} \reward(\states_{\itertrajectories, \iterpartial}, \actions_{\itertrajectories, \iterpartial}) - \valuefunction^{\policy_{\policyparameters}}(\states_{\itertrajectories, \iterpartial}) \biggr)
\end{align}

Weights $\importanceweights$ adjust gradient information coming from trajectory collected \emph{off the current policy} (i.e. using the behavioural policy) according to how much the two policies are in agreement.
Crucially, this approach works under the assumptions that the marginal state probabilities with respect to the current and old policy networks' parameters, say \(p_{\policyparameters^\text{\,cur}}(\states)\) and \(p_{\policyparameters^\text{\,old}}(\states)\), are sufficiently close to each other, and that the dynamics of the respective (PO)MDPs are the same.
In the off-policy case, these assumptions hold true because during learning the replay buffer is regularly updated with newly experienced trajectories and cleared from the older ones, and because the (PO)MDP dynamics are usually assumed to be fixed.
In the offline setting, this is however not guaranteed, see \textcite{Levine2020} for a recent review of some of these and other open challenges, and the next section for a more complete overview of relevant works in causal RL on integration, explicitness and sources across causal reinforcement learning and animal cognition.

\section{Bringing together causality in natural and artificial agents}
\label{sec:bridging}

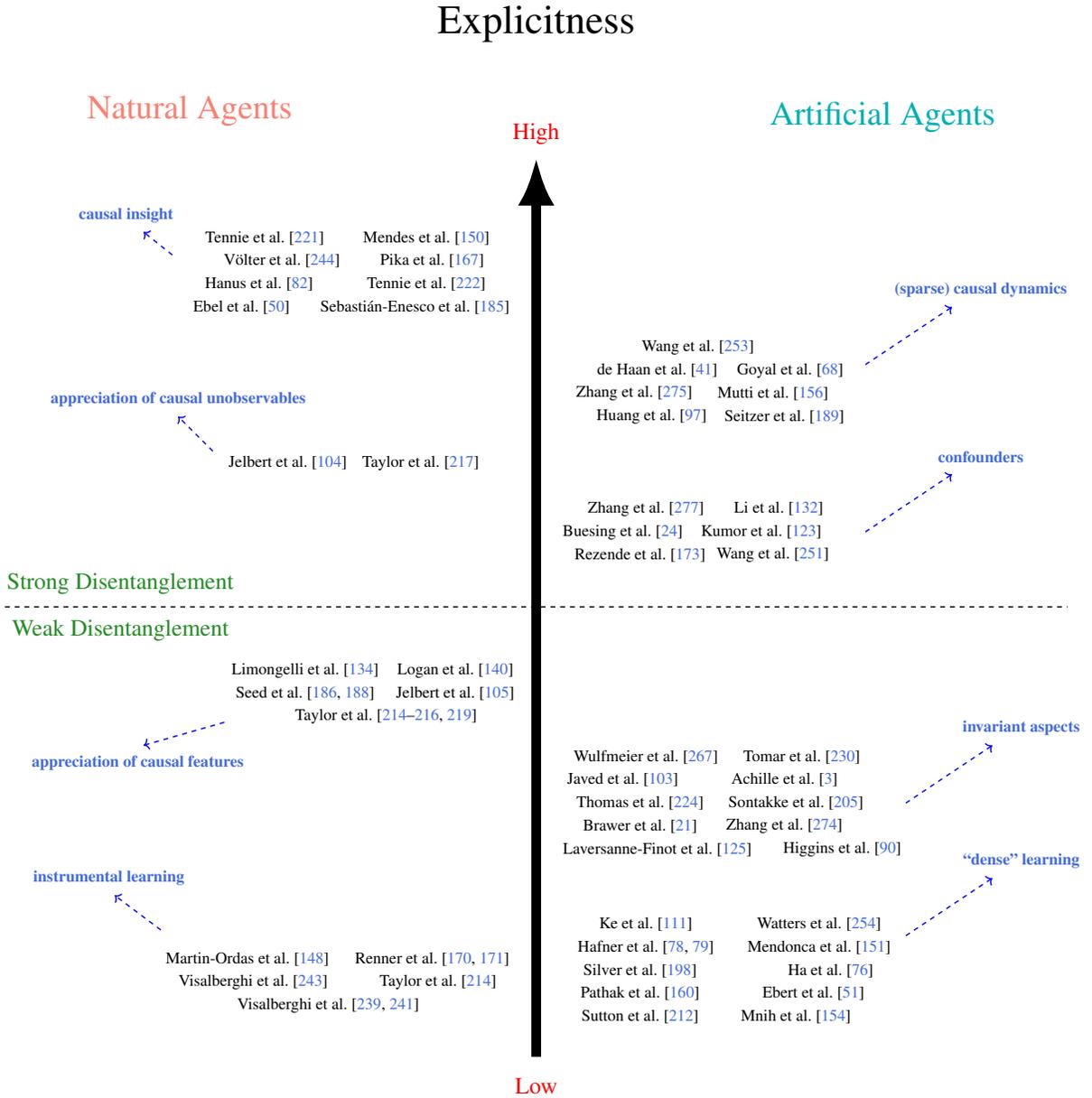
\begin{figure}[ht!]
	\centering
	\scalebox{.68}{\begin{tikzpicture}
	\begin{pgfonlayer}{nodelayer}
		\node [style=none] (0) at (0, -21.5) {};
		\node [style=none] (1) at (0, 17.5) {};
		\node [style=none] (2) at (-5.25, 11) {\textcite{Sebastian-Enesco2022}};
		\node [style=none] (3) at (-12, 12) {\textcite{Hanus2011}};
		\node [style=none] (4) at (-12.75, 11) {\textcite{Ebel2019}};
		\node [style=none] (5) at (-11, 13) {\textcite{Volter2016}};
		\node [style=none] (6) at (-11.75, 14) {\textcite{Tennie2019}};
		\node [style=none] (7) at (-4.5, 13) {\textcite{Pika2020}};
		\node [style=none] (8) at (-4.75, 12) {\textcite{Tennie2010}};
		\node [style=none] (9) at (-4.75, 14) {\textcite{Mendes2007}};
		\node [style=none] (11) at (4.25, 7.25) {\textcite{Zhang2021}};
		\node [style=none] (12) at (5, 6.25) {\textcite{Huang2022a}};
		\node [style=none] (13) at (5.25, 8.25) {\textcite{DeHaan2019}};
		\node [style=none] (14) at (7, 9.25) {\textcite{Wang2022c}};
		\node [style=none] (15) at (11, 8.25) {\textcite{Goyal2020}};
		\node [style=none] (16) at (10.25, 7.25) {\textcite{Mutti2023}};
		\node [style=none] (17) at (10.75, 6.25) {\textcite{Seitzer2021}};
		\node [style=none] (18) at (-6.5, -6.75) {\textcite{Taylor2007, Taylor2009a, Taylor2012a, Taylor2012b}};
		\node [style=none] (20) at (-3.5, -4.75) {\textcite{Logan2014a}};
		\node [style=none] (21) at (-10, -5.75) {\textcite{Seed2006, Seed2009}};
		\node [style=none] (22) at (-10.75, 4.25) {\textcite{Jelbert2019}};
		\node [style=none] (23) at (-3.5, -5.75) {\textcite{Jelbert2014}};
		\node [style=none] (25) at (-5, 4.25) {\textcite{Taylor2012}};
		\node [style=none] (26) at (-10, -4.75) {\textcite{Limongelli1995}};
		\node [style=none] (27) at (3.75, 1.25) {\textcite{Buesing2019}};
		\node [style=none] (28) at (4.5, 0.25) {\textcite{Rezende2020}};
		\node [style=none] (29) at (4.75, 2.25) {\textcite{Zhang2020a}};
		\node [style=none] (31) at (10.5, 2.25) {\textcite{Li2020}};
		\node [style=none] (32) at (9.75, 1.25) {\textcite{Kumor2021}};
		\node [style=none] (33) at (10.25, 0.25) {\textcite{Wang2021a}};
		\node [style=none] (34) at (3.75, -9.5) {\textcite{Javed2020}};
		\node [style=none] (35) at (4.5, -10.5) {\textcite{Thomas2018}};
		\node [style=none] (36) at (4.75, -8.5) {\textcite{Wulfmeier2021}};
		\node [style=none] (38) at (11.5, -8.5) {\textcite{Tomar2021}};
		\node [style=none] (39) at (10.75, -9.5) {\textcite{Achille2018b}};
		\node [style=none] (40) at (11.25, -10.5) {\textcite{Sontakke2021a}};
		\node [style=none] (41) at (5.25, -12.5) {\textcite{Laversanne-Finot2018}};
		\node [style=none] (43) at (4.5, -11.5) {\textcite{Brawer2021}};
		\node [style=none] (44) at (10.75, -11.5) {\textcite{Zhang2020}};
		\node [style=none] (45) at (13.25, -12.5) {\textcite{Higgins2017a}};
		\node [style=none] (46) at (4.75, -16.75) {\textcite{Hafner2019, Hafner2020}};
		\node [style=none] (47) at (4.5, -17.75) {\textcite{Silver2018a}};
		\node [style=none] (48) at (4.75, -15.75) {\textcite{Ke2019a}};
		\node [style=none] (50) at (12.25, -15.75) {\textcite{Watters2019}};
		\node [style=none] (51) at (12.25, -16.75) {\textcite{Mendonca2021}};
		\node [style=none] (52) at (12.75, -17.75) {\textcite{Ha2018}};
		\node [style=none] (53) at (4.5, -19.75) {\textcite{Sutton2018}};
		\node [style=none] (54) at (4.5, -18.75) {\textcite{Pathak2017}};
		\node [style=none] (55) at (12, -18.75) {\textcite{Ebert2018}};
		\node [style=none] (56) at (11.25, -19.75) {\textcite{Mnih2015}};
		\node [style=none] (57) at (-9, -19.25) {\textcite{Visalberghi1993, Visalberghi1994}};
		\node [style=none] (58) at (-12.5, -17.25) {\textcite{Martin-Ordas2008}};
		\node [style=none] (59) at (-12.25, -18.25) {\textcite{Visalberghi1989}};
		\node [style=none] (62) at (-4.5, -17.25) {\textcite{Renner2017, Renner2021}};
		\node [style=none] (63) at (-4.25, -18.25) {\textcite{Taylor2012a}};
		\node [style=none] (64) at (15, 19.25) {{\huge \textcolor{TealBlue}{Artificial Agents}}};
		\node [style=none] (65) at (-15, 19.5) {{\huge \textcolor{Salmon}{Natural Agents}}};
		\node [style=none] (66) at (-17.75, 15) {\textcolor{RoyalBlue}{\textbf{causal insight}}};
		\node [style=none] (67) at (-15.75, 13.25) {};
		\node [style=none] (68) at (-17, 14.25) {};
		\node [style=none] (69) at (-23, -2) {};
		\node [style=none] (70) at (23, -2) {};
		\node [style=none] (71) at (-18, -1) {{\Large \textcolor{ForestGreen}{Strong Disentanglement}}};
		\node [style=none] (72) at (-18, -3) {{\Large \textcolor{ForestGreen}{Weak Disentanglement}}};
		\node [style=none] (73) at (-18.5, -13.75) {\textcolor{RoyalBlue}{\textbf{instrumental learning}}};
		\node [style=none] (74) at (-16.25, -16) {};
		\node [style=none] (75) at (-18.25, -14.5) {};
		\node [style=none] (76) at (-15.5, 7) {\textcolor{RoyalBlue}{\textbf{appreciation of causal unobservables}}};
		\node [style=none] (77) at (-14, 4.75) {};
		\node [style=none] (78) at (-15.5, 6.25) {};
		\node [style=none] (79) at (-17.25, -8.75) {\textcolor{RoyalBlue}{\textbf{appreciation of causal features}}};
		\node [style=none] (80) at (-13.5, -7) {};
		\node [style=none] (81) at (-17, -8) {};
		\node [style=none] (82) at (19.25, 4.5) {\textcolor{RoyalBlue}{\textbf{confounders}}};
		\node [style=none] (83) at (14.25, 1.25) {};
		\node [style=none] (84) at (18, 3.75) {};
		\node [style=none] (85) at (19.25, 11.75) {\textcolor{RoyalBlue}{\textbf{(sparse) causal dynamics}}};
		\node [style=none] (86) at (14.25, 8.5) {};
		\node [style=none] (87) at (18, 11) {};
		\node [style=none] (88) at (21, -13) {\textcolor{RoyalBlue}{\textbf{``dense'' learning}}};
		\node [style=none] (89) at (16, -16.25) {};
		\node [style=none] (90) at (19.75, -13.75) {};
		\node [style=none] (91) at (21, -7.25) {\textcolor{RoyalBlue}{\textbf{invariant aspects}}};
		\node [style=none] (92) at (16, -10.5) {};
		\node [style=none] (93) at (19.75, -8) {};
		\node [style=none] (94) at (0, 23.25) {{\Huge Explicitness}};
		\node [style=none] (95) at (0, 18.5) {{\Large \textcolor{red}{High}}};
		\node [style=none] (96) at (0, -22.75) {{\Large \textcolor{red}{Low}}};
	\end{pgfonlayer}
	\begin{pgfonlayer}{edgelayer}
		\draw [style=ultrathick-arrow] (0.center) to (1.center);
		\draw [style=thick-blue-dashed-arrow] (67.center) to (68.center);
		\draw [style=big dashes] (69.center) to (70.center);
		\draw [style=thick-blue-dashed-arrow] (74.center) to (75.center);
		\draw [style=thick-blue-dashed-arrow] (77.center) to (78.center);
		\draw [style=thick-blue-dashed-arrow] (80.center) to (81.center);
		\draw [style=thick-blue-dashed-arrow] (83.center) to (84.center);
		\draw [style=thick-blue-dashed-arrow] (86.center) to (87.center);
		\draw [style=thick-blue-dashed-arrow] (89.center) to (90.center);
		\draw [style=thick-blue-dashed-arrow] (92.center) to (93.center);
	\end{pgfonlayer}
\end{tikzpicture}}
        \caption{\textbf{The explicitness spectrum.}}
        \label{fig:explicitness}
\end{figure}

Recent work in (deep) reinforcement learning, in the area now called causal reinforcement learning, can help us shed light on ways to translate algorithms from machine learning into a more systematic study of causal learning agents.
Using this line of work, we thus review a series of algorithmic implementations and models from causal RL and place them on a spectrum of increasingly high disentanglement, providing a comparative analysis with empirical and conceptual works in the animal cognition literature, see \cref{fig:explicitness}.
This will in turn suggest a more concrete connection with the explicitness dimension of \textcite{Starzak2021}'s framework, paving then the way for an understanding of causal information from different sources and possible strategies to integrate them sensibly.

\subsection{Explicitness of causal representations}
\label{subsec:explicitness-causal-rep}

\subsubsection{Weak disentanglement}
\label{subsub:weakDisentanglement}

At the lower end of the explicitness spectrum (see \cref{fig:explicitness}), we find agents of traditional (non-causal) deep RL setups that are successful at solving a variety of narrow tasks by engaging in forms of \textbf{dense learning}, meaning that they often appear to learn at least some of the dependencies between their actions and desired outcomes/rewards \parencite{Sutton2018, Mnih2015, Ha2018, Silver2018a}, and in some cases they are augmented with more sophisticated forms of planning, curiosity-based exploration and the ability to achieve a variety of goals in high-dimensional environments \parencite{Ebert2018, Pathak2017, Hafner2019, Ke2019a, Watters2019, Hafner2020, Mendonca2021}.
Nonetheless, the web of dependencies learned by these agents are usually dense because, as dependencies that are associative in nature, they include spurious features and/or relationships.
In other words, dense learning in these agents goes hand in hand with a lack of causal information processing.
These agents are in several ways akin to animal subjects engaging in \textbf{instrumental learning} \parencite{Visalberghi1989, Visalberghi1993, Visalberghi1994, Taylor2012a, Renner2021, Renner2017, Martin-Ordas2008} (see \cref{sec:what-causal-cog}), except for the amount of data samples used during training.

To have a better understanding of algorithms and empirical results higher in our explicitness scale, and their relation to weak and strong disentanglement, it is then useful to look at more specific features of causal representations. 
In classical RL, particularly when the problem is presented as a POMDP, a representation can be understood in two different ways:
\begin{itemize}
    \item in \emph{model-free RL}, these are representations of factors (i.e. codes) given as inputs to a policy, \(\policy(\actions|\states)\), i.e. the state representations (implemented as vectors of state variables) fed to the policy network to produce an action, while
    \item in \emph{model-based RL}, the term representation points at a model of the transition dynamics (involving, in turn, the state representations) \(\prob{\states_{t+1}|\states_{t}, \actions_{t}}\).
\end{itemize}

Based on this distinction, we suggest that there are two different ways to understand what a causal representation involves in causal reinforcement learning: 1) in causal model-free RL, a causal representation describes a particular encoding, often a compression, of the observations into latent states (with a causal flavour) while 2) in causal model-based RL, a causal representation models both the latent states and the causal dynamics of the environment.
Here we suggest that the first kind of causal representation has a lower degree of explicitness than the second one, since it fails to capture the causal dynamics of the environment. 
We thus view it as a possible example of \emph{weak disentanglement}.
The latter instead is defined precisely in terms of its ability to map causal dynamics and is thus an example of \emph{strong disentanglement} (\cref{subsec:explicit-in-crl-agents}).

In causal model-free RL, in particular for a partially observable setting, an agent is said to learn an explicit causal representation if it can map high-dimensional observations to state representations that are disentangled, uncovering codes that represent some parts of the causal structure (the causal factors) of the data-generating process, but not including a complete disentangled representation (see \cref{subsec:explicit-in-crl-agents}).
Thus, an agent can be said to exploit this (partial) causal information if such information facilitates policy learning or has a positive impact on policy execution when the causal representation is fed to the policy network.

More specifically, the benefits of this kind of causal representation would follow from capturing \textbf{invariant aspects} of an environment, represented by the causal codes, making learning and using a policy more robust to changes in secondary, irrelevant variables (i.e. improving on generalisation)\footnote{The connection between invariant aspects of an environment and causality has been established, for instance in \textcite{Pearl2009a, Peters2017, Spirtes2000}. However, there is ongoing disagreement on how traditional machine learning methods should be adjusted to capture invariance in data (with some theoretical guarantees) \parencite{Arjovsky2020, Choe2020, Kamath2021, Rosenfeld2021}. Similarly, it remains unclear how the RL framework could be impacted by such theoretical advancements in the long term. See also works cited in the main text.}.
There are many works on invariant representation learning for RL \parencite{Zhang2020a, Bica2021, Sonar2021, Stojanov2021, Zhang2021a, Lu2022, Lu2022a}, but it is not entirely clear whether all these invariant representations qualify as disentangled representations in the weak sense discussed here.

Conversely, as long as causal codes can be considered invariant aspects of the environment, disentangled representations in a RL agent are necessarily some kind of invariant representations.
Empirically, the extent to which this is key to credit assignment, i.e. the ability of an agent to determine which actions (and/or states) contributed the most to successful performance in a certain domain, remains however to be proved.
At present, there is some evidence indicating that learning a causal representation, one achieving weak disentanglement, provides several benefits in RL, e.g. a better exploration of the state space and more robust learning \parencite{Achille2018b, Laversanne-Finot2018, Higgins2017a, Wulfmeier2021, Thomas2018, Sontakke2021a, Zhang2020, Brawer2021, Tomar2021, Javed2020}.

At a comparable level of explicitness, in the animal cognition literature we find subjects that can successfully solve trap-tube tasks, water-displacement tasks, and/or tasks with similar setups where an attention to objects' physical and functional properties is essential.
Recognising these objects' invariants and their causal role for the purpose of solving the given task is suggestive of an \textbf{appreciation of causal features} constituting factors of the generative process, by disambiguating them, at least in part, from non-causal ones \parencite{Limongelli1995, Taylor2007, Taylor2009a, Jelbert2014, Logan2014a, Seed2009, Seed2006, Taylor2012b, Taylor2012a}.
For instance, there is empirical evidence that crows can drop stones into baited, water-filled tubes according to stones' width and water levels.
Lower water levels and wide tubes hinder in fact water displacement with the available stones, which is necessary to reach the reward \parencite{Logan2014a}, see \cref{subsec:causal-interventions}.

\subsubsection{Strong disentanglement}
\label{subsubsec:strong-disentanglement-examples}

In causal model-based reinforcement learning the agent is specifically trained to learn a \emph{world model}, a model of the dynamics of the environment, then used for planning and decision-making (i.e. selecting the next action).
In this context, causality-inspired approaches involve revealing and exploiting more causal aspects of the (modelled) environmental dynamics, regarded as crucial for having more capable learning agents that, for instance, do not fall prey to spurious correlations like agents with less explicit models might.

To achieve this, we have attempts to handle \textbf{confounders}, hidden common causes that can have an impact on factors and their state transitions (see \cref{subsec:from-disent-to-crl}), by deconfounding the dynamics of a POMDP.
Practically, this entails modelling state-transitions as affected by confounders, whose presence is either assumed from the start \parencite{Li2020, Buesing2019, Zhang2020a, Kumor2021}, or can emerge from initially unaccounted parts of the dynamics/predictive model through a process of decomposition of observations into confounding and relevant state information \parencite{Rezende2020, Wang2021a}.
This leads to more explicit models because the effects of confounding factors are isolated to obtain a more robust understanding of how events in the environment unfold.

More in detail, we can look at \textcite{Li2020} as an example of the first kind (known confounders), based on object-centric learning (using graph-neural networks) combined with a model of the transition dynamics that is assumed to be confounded by time-invariant hidden variables, e.g. the object's masses, friction coefficients, etc.
The goal of the agent here is to solve a POMDP, but this requires learning a generative model that is deconfounded, estimating the confounding variables for each object (using tools from do-calculus \parencite{Pearl2009a, Pearl2018c}), which in turn can be used to generate accurate observation trajectories had the initial conditions been different (e.g. the objects' position).
Effectively, this causal world model enables a kind of future counterfactual planning that starts with the question of what would have happened under alternative initial conditions, i.e. given an intervention that changes the starting states.
On the other hand, for the second group (unknown confounders) we can consider \textcite{Rezende2020} where agents with partial models, i.e. models learnt using past actions and the initial agent's state as opposed to the full trajectory of past observations, are shown to be less robust to policy changes.
These partial models are in fact confounded by past observations, which are not used to train the model but do anyway influence the policy picked by the agent, but can be adjusted for such confounders by using once again techniques from do-calculus.

In the animal cognition literature, understanding the influence of potential confounders can be linked to an \textbf{appreciation of causal unobservables}, such as in crows adjusting their actions depending on changes in experimental variables that are not visible to them \parencite{Taylor2012, Jelbert2019}.
In one study, crows were tested on task consisting of extracting some food from a box, placed on a table and in front of a curtain.
From behind the curtain, a human could operate a wooden stick that through a hole in the curtain could come close to the food box, therefore causing trouble for the crows trying to reach the food.
The presence/absence of the human thus confounds whether it is ``safe'' to go and retrieve the food from the box (because in principle a stick's movement does not create danger, unless it is intentionally used to poke through a hole, for example by a human experiment), therefore it would be useful to be able to reason about what is behind the curtain.
The evidence reported by \textcite{Taylor2012} suggests that crows can attribute the movement of the stick to a hidden agent behind the curtain and act accordingly, e.g. being more cautious when they do not observe anyone leaving the experiment's room (because the stick could move again).
Similarly, in a context where a food dispenser is activated by means of placing objects on it and where an object's weight confounds the food release (only heavy enough objects activate the dispenser), crows can learn to infer the weights of the objects from their movements in a breeze and pick the appropriate ones to get the food from the dispenser \parencite{Jelbert2019}.
In both studies, the animal subjects were able to adjust their behaviour by paying attention to the reward dynamics, i.e. to whether narrow or wide tube were to be preferred (according to the respective water level), or to whether light or heavy objects were activating the dispenser.

Beyond confounding factors, model-based reinforcement learning can be improved by observing that key causal relationships in the environment, relevant for solving a particular problem, do not involve all state variables and transitions among them.
That is, the causal dependencies among variables in the environment that an agent can have an effect on, for the purpose of reaching a certain goal state, form a causal structure that is \emph{sparse}, in the sense that it only captures some deeper facts about a whole class of problems (or environments) for a particular agent.
For example, an agent might be capable of accessing a certain area of a building by means of a detailed world model that accurately predicts what happens when a red button located next to a glass door is pressed, while standing on a floor with hexagonal tiles.
This detailed model keeps track of all possible dependencies among the colours of buttons, the material of adjacent doors, and the geometric shape of the floor tiles on which the agent stands (and it might predict with some confidence that only when a combination of those features is encountered, then access to a certain area will be granted).
The problem is that most of those dependencies are likely just spurious correlations, hiding the most fundamental (causal) fact that a button next to a door in general tends to open that door when pressed.

Thus, instead of learning indiscriminately every conditional dependence relation (causal or not) among state variables at adjacent time steps, an agent should strive to learn a causal transition model that identifies the causal relationships that matters for the class of tasks at hand. 
Precisely, focusing on the \textbf{(sparse) causal dynamics} present in a given environment means to identify the subset of latent state variables, or causal factors $\States$, that are likely to form generalisable causal relationships, exploitable not only for the task at hand but for similar tasks as well \parencite{Huang2022a, Wang2022c, Zhang2021, Seitzer2021, Goyal2020, DeHaan2019, Mutti2023}.
Therefore, in contrast to modelling dense dynamics, leveraging causal sparsity is more computationally efficient and can help an agent to avoid learning spurious correlations over time.

One influential approach hinges upon notions of state abstraction \parencite{Tomar2021} (see also \textcite{Li2006} for more background).
A state abstraction can be regarded as a compact (latent) representation that is invariant to task-irrelevant information (i.e. only information relevant to a specific problem is encoded), and is technically defined as a (probabilistic) bisimulation.
A bisimulation is a structure-preserving equivalence relation of states of a (PO)MDP, $\bisim \subseteq \States \times \States$, describing equivalence classes of states $\States / \bisim$, for all actions in the action space $\Actions$, with transition dynamics leading from states with the same reward $\Reward$ to the same equivalence classes of states, i.e. for $\states_1, \states_2 \in \bisim \text{ and } \forall i, a \in \States / \bisim, \Actions$ the following conditions apply \parencite{Zhang2020, Zhang2021, Zhang2021a, Tomar2021}

\begin{align}
    \prob{i \mid \states_1, a} & = \prob{i \mid \states_2, a} \nonumber \\
    \reward(\states_1, a) & = \reward(\states_2, a)
\end{align}

see also the notion of ``causal states'' in \textcite{Shalizi2001}, roughly probabilistic bisimulations for stationary stochastic processes, without rewards and actions.

This line of work can lead to a higher level of explicitness, via strong(er) disentanglement, as exemplified for instance by \textcite{Wang2022c}, showing empirically that their architecture can learn codes identifying causal factors as well as the causal mechanisms between them (across times steps).
This in particular reveals causal transition dynamics for each state, i.e. determining whether a causal relationship between a state at time $t$ and its successor at time $t+1$ is present or not, and is relevant for an agent solving a particular problem\footnote{To be precise, in \textcite{Wang2022c} the state abstraction is learned on the states of a MDP, so the challenge of deriving a disentangled latent representation (from high-dimensional observations) is bypassed. In other words, weak disentanglement is taken for granted.}.

In \textcite{Starzak2021}, a high-level of explicitness is connected with adaptive behaviour supported by a flexible use of causal information, which enables an animal to re-use acquired knowledge or past behaviours (with the appropriate changes, if necessary) to reach new goals, the same goals but in a slightly different context, and/or solve tasks never encountered before.
Therefore, at the top level of the explicitness spectrum, we find \textbf{causal insight}, a term we use to refer broadly to this type of generalisation abilities, chiefly involving a deeper realisation of what a problems/scenario entails, based on causal knowledge.
These can be often seen in transfer learning and innovative/insightful problem solving, e.g. in the floating-peanut task \parencite{Hanus2011, Tennie2010, Ebel2019, Sebastian-Enesco2022, Mendes2007, Pika2020, Volter2016, Tennie2019}, in which the solution to a task is allegedly reached via an adaptive restructuring of one's experience \parencite{Thorpe1956, Kounios2014}.
The main idea here is that non-human animal appear to be capable of cognitive feats thanks to highly explicit causal models of both state variables and transition dynamics, thought there is some disagreement in the experimental literature on the extent to which this effectively happens \parencite{Taylor2012b, Shupe2024, Lind2009}.

\subsection{Sources of causal information}
\label{subsec:sources-causal-rl}

\begin{figure}[ht!]
	\centering
	\scalebox{.68}{\begin{tikzpicture}
	\begin{pgfonlayer}{nodelayer}
		\node [style=ellipse-23-14-red] (0) at (-0.25, -2) {};
		\node [style=ellipse-17-9-orange] (1) at (3.75, -2) {};
		\node [style=ellipse-9-5-green] (2) at (9.75, -2) {};
		\node [style=none] (4) at (-24.75, -2) {};
		\node [style=none] (5) at (24.25, -2) {};
		\node [style=none] (12) at (-8.25, -3.75) {\textcite{Renner2021}};
		\node [style=none] (13) at (-7, -4.75) {\textcite{Tennie2010}};
		\node [style=none] (14) at (-4, -6.75) {\textcite{Pika2020}};
		\node [style=none] (15) at (-5.5, -5.75) {\textcite{Renner2017}};
		\node [style=none] (16) at (-0.75, -8.75) {\textcite{Call2005}};
		\node [style=none] (17) at (-2.5, -7.75) {\textcite{Horner2005}};
		\node [style=none] (18) at (1, -9.75) {\textcite{Tennie2006}};
		\node [style=none] (19) at (-9.25, -1.25) {\textcite{DeHaan2019}};
		\node [style=none] (20) at (-9, 0.75) {\textcite{Zhang2020a}};
		\node [style=none] (21) at (-9, -0.25) {\textcite{Ross2010}};
		\node [style=none] (22) at (-5.75, 1.75) {\textcite{Kumor2021}};
		\node [style=none] (23) at (-2.5, 0.75) {\textcite{Ho2016}};
		\node [style=none] (24) at (-2.25, -0.25) {\textcite{Lu2022b}};
		\node [style=none] (25) at (-3, -1.25) {\textcite{Li2017a}};
		\node [style=none] (26) at (0.25, 6) {\textcite{Wang2021a}};
		\node [style=none] (27) at (0.25, 5) {\textcite{Kumar2020}};
		\node [style=none] (28) at (0.25, 3) {\textcite{Levine2020}};
		\node [style=none] (29) at (0.25, 4) {\textcite{Ghosh2022a}};
		\node [style=none] (30) at (13.5, 3.25) {\textcite{Ziebart2008}};
		\node [style=none] (31) at (9.25, 5.25) {\textcite{Wulfmeier2016}};
		\node [style=none] (32) at (11.75, 4.25) {\textcite{Abbeel2004}};
		\node [style=none] (33) at (7, 6.25) {\textcite{Fu2018}};
		\node [style=none] (34) at (4.25, -3.25) {\textcite{Tennie2006}};
		\node [style=none] (35) at (5.25, -4.25) {\textcite{Hopper2008}};
		\node [style=none] (36) at (5.75, -5.25) {\textcite{Hopper2010}};
		\node [style=none] (37) at (14.5, -2.75) {\textcite{Jelbert2019}};
		\node [style=none] (38) at (13.5, -3.75) {\textcite{Volter2016}};
		\node [style=none] (39) at (11.5, -5.75) {\textcite{Tennie2019}};
		\node [style=none] (40) at (12.75, -4.75) {\textcite{Taylor2012}};
		\node [style=none] (41) at (-19.25, -1) {\textcite{Sutton2018}};
		\node [style=none] (42) at (-17, 3) {\textcite{Igl2018}};
		\node [style=none] (43) at (-18.5, 1) {\textcite{Schulman2017}};
		\node [style=none] (44) at (-15.5, 5) {\textcite{Haber2018a}};
		\node [style=none] (45) at (-17.75, 2) {\textcite{Mnih2016}};
		\node [style=none] (46) at (-16, 4) {\textcite{Sontakke2021a}};
		\node [style=none] (47) at (-15.25, 6) {\textcite{Javed2020}};
		\node [style=none] (48) at (-18, 0) {\textcite{Ha2018}};
		\node [style=none] (49) at (-1.25, 9.75) {\textcite{Mendonca2021}};
		\node [style=none] (50) at (-7.5, 8.75) {\textcite{Silver2018a}};
		\node [style=none] (51) at (-1.25, 7.75) {\textcite{Buesing2019}};
		\node [style=none] (52) at (-7.5, 9.75) {\textcite{Tomar2021}};
		\node [style=none] (53) at (-7.5, 7.75) {\textcite{Hafner2020}};
		\node [style=none] (54) at (-0.75, 8.75) {\textcite{Andrychowicz2017}};
		\node [style=none] (55) at (-4.25, 10.75) {\textcite{Zhang2021}};
		\node [style=none] (56) at (-18.75, -3.5) {\textcite{Seed2006, Seed2009}};
		\node [style=none] (57) at (-17.5, -4.5) {\textcite{Martin-Ordas2008}};
		\node [style=none] (58) at (-17.75, -6.5) {\textcite{Taylor2009a}};
		\node [style=none] (59) at (-18.5, -5.5) {\textcite{Limongelli1995}};
		\node [style=none] (60) at (-15.25, -8.5) {\textcite{Visalberghi1993, Visalberghi1994}};
		\node [style=none] (61) at (-16.75, -7.5) {\textcite{Visalberghi1989}};
		\node [style=none] (62) at (-8.5, -10.75) {\textcite{Taylor2007}};
		\node [style=none] (63) at (-8.5, -11.75) {\textcite{Taylor2012b}};
		\node [style=none] (64) at (-8.5, -12.75) {\textcite{Taylor2012}};
		\node [style=none] (65) at (5.25, -13.25) {\textcite{Hanus2011}};
		\node [style=none] (66) at (-0.75, -13.25) {\textcite{Jelbert2014}};
		\node [style=none] (67) at (-0.75, -12.25) {\textcite{Mendes2007}};
		\node [style=none] (68) at (2.5, -14.25) {\textcite{Sebastian-Enesco2022}};
		\node [style=none] (69) at (5.75, -12.25) {\textcite{Logan2014a}};
		\node [style=none] (70) at (-20, 16.5) {{\huge \textcolor{TealBlue}{Artificial Agents}}};
		\node [style=none] (71) at (-20.25, -20.5) {{\huge \textcolor{Salmon}{Natural Agents}}};
		\node [style=fbox-.5-.5-red] (72) at (-15.5, 19.5) {};
		\node [style=fbox-.5-.5-orange] (73) at (-3.25, 19.5) {};
		\node [style=fbox-.5-.5-green] (74) at (9, 19.5) {};
		\node [style=none] (79) at (-10.25, 19.5) {\textcolor{red}{egocentric causal information}};
		\node [style=none] (82) at (1.25, 19.5) {\textcolor{Orange}{social causal information}};
		\node [style=none] (83) at (13.75, 19.5) {\textcolor{ForestGreen}{natural causal information}};
		\node [style=none] (84) at (-20.75, 8.75) {\textcolor{RoyalBlue}{\textbf{online learning}}};
		\node [style=none] (85) at (-18.25, 5.5) {};
		\node [style=none] (86) at (-20.75, 8) {};
		\node [style=none] (87) at (-5.25, 13.75) {\textcolor{RoyalBlue}{\textbf{online + offpolicy learning}}};
		\node [style=none] (88) at (-4.75, 11.5) {};
		\node [style=none] (89) at (-5.25, 13) {};
		\node [style=none] (90) at (-14.5, 12) {\textcolor{RoyalBlue}{\textbf{imitation learning}}};
		\node [style=none] (91) at (-9.25, 2.25) {};
		\node [style=none] (92) at (-14.5, 11.25) {};
		\node [style=none] (93) at (10, 13.75) {\textcolor{RoyalBlue}{\textbf{offline RL}}};
		\node [style=none] (94) at (3.25, 6.5) {};
		\node [style=none] (95) at (10, 13) {};
		\node [style=none] (96) at (20.25, 11.25) {\textcolor{RoyalBlue}{\textbf{inverse RL}}};
		\node [style=none] (97) at (12.75, 5.75) {};
		\node [style=none] (98) at (20.25, 10.5) {};
		\node [style=none] (99) at (-21, -13) {\textcolor{RoyalBlue}{\textbf{trap-tube task}}};
		\node [style=none] (102) at (-18.25, -9.5) {};
		\node [style=none] (103) at (-21, -12.25) {};
		\node [style=none] (104) at (-13.5, -16.75) {\textcolor{RoyalBlue}{\textbf{tool-tube task}}};
		\node [style=none] (105) at (-11.5, -13) {};
		\node [style=none] (106) at (-13.5, -16) {};
		\node [style=none] (107) at (3.75, -18.75) {\textcolor{RoyalBlue}{\textbf{floating-reward task}}};
		\node [style=none] (108) at (3.25, -15.25) {};
		\node [style=none] (109) at (3.75, -18) {};
		\node [style=none] (110) at (-5.75, -18.25) {\textcolor{RoyalBlue}{\textbf{imitation vs.  emulation tasks}}};
		\node [style=none] (111) at (-4, -9.25) {};
		\node [style=none] (112) at (-5.75, -17.5) {};
		\node [style=none] (113) at (12.5, -17.25) {\textcolor{RoyalBlue}{\textbf{"ghost"-condition tasks}}};
		\node [style=none] (114) at (6.5, -6) {};
		\node [style=none] (115) at (12.5, -16.5) {};
		\node [style=none] (116) at (19, -13.75) {\textcolor{RoyalBlue}{\textbf{observational causal tasks}}};
		\node [style=none] (117) at (12.5, -6.5) {};
		\node [style=none] (118) at (19, -13) {};
		\node [style=none] (119) at (0, 24.5) {{\Huge Sources}};
	\end{pgfonlayer}
	\begin{pgfonlayer}{edgelayer}
		\draw [style=big dashes] (4.center) to (5.center);
		\draw [style=thick-blue-dashed-arrow] (85.center) to (86.center);
		\draw [style=thick-blue-dashed-arrow] (88.center) to (89.center);
		\draw [style=thick-blue-dashed-arrow] (91.center) to (92.center);
		\draw [style=thick-blue-dashed-arrow] (94.center) to (95.center);
		\draw [style=thick-blue-dashed-arrow] (97.center) to (98.center);
		\draw [style=thick-blue-dashed-arrow] (102.center) to (103.center);
		\draw [style=thick-blue-dashed-arrow] (105.center) to (106.center);
		\draw [style=thick-blue-dashed-arrow] (108.center) to (109.center);
		\draw [style=thick-blue-dashed-arrow] (111.center) to (112.center);
		\draw [style=thick-blue-dashed-arrow] (114.center) to (115.center);
		\draw [style=thick-blue-dashed-arrow] (117.center) to (118.center);
	\end{pgfonlayer}
\end{tikzpicture}}
        \caption{\textbf{Sources of causal information.}}
        \label{fig:sources}
\end{figure}
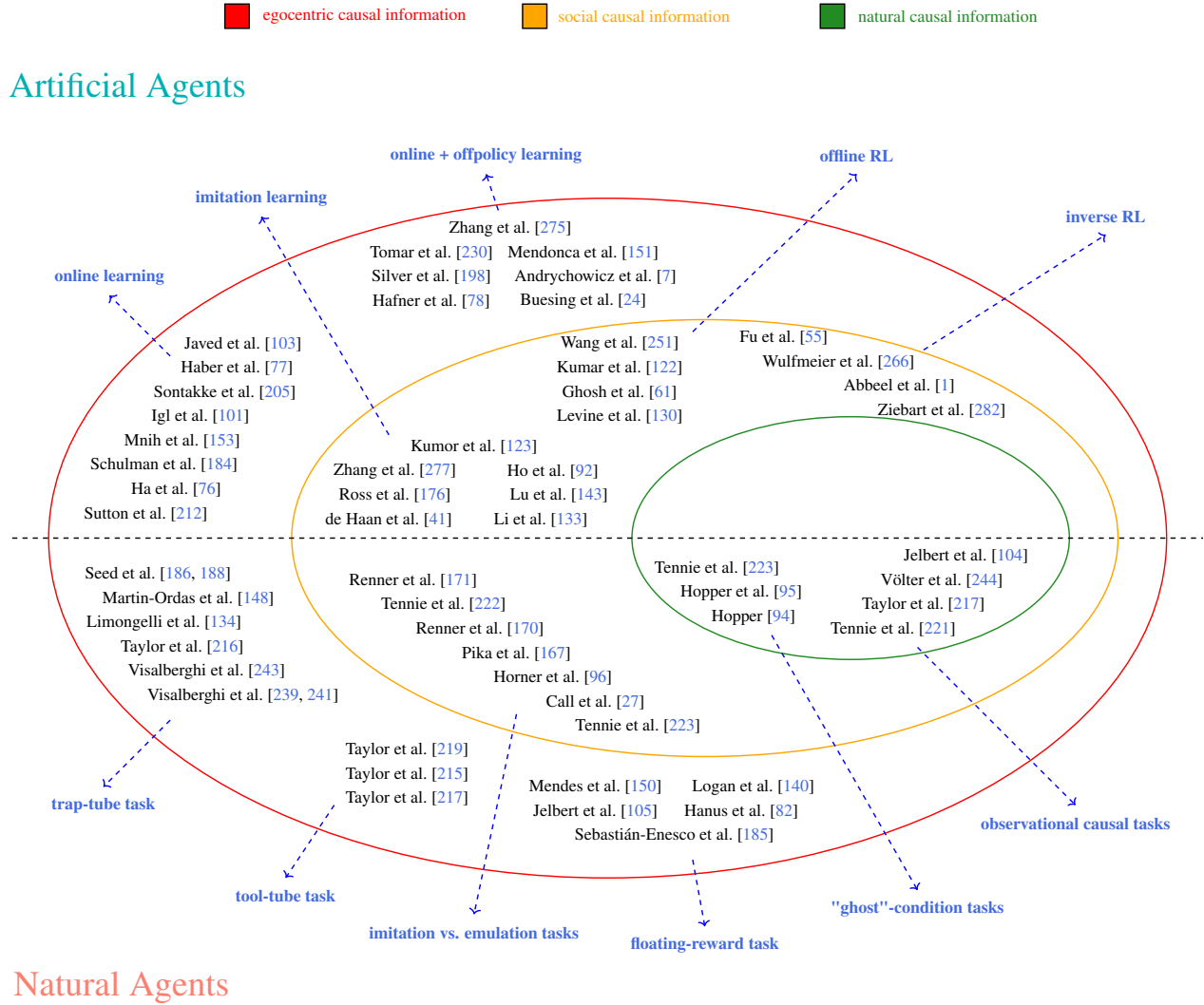

\subsubsection{Egocentric causal information}
Successfully learning from online interactions in an environment implies appreciating, to some extent, the relevance of certain action-outcome contingencies for reward maximisation or reaching a certain goal.
Learning from online interaction amounts to instrumental learning, which chiefly involves egocentric causal information (see \cref{fig:sources}) and has been extensively studied in animals (see \cref{subsec:sources-causal-info}), particularly with a variety of tasks including \textbf{trap-tube tasks} \parencite{Visalberghi1989, Visalberghi1993, Visalberghi1994, Limongelli1995, Taylor2009a, Seed2009, Seed2006, Martin-Ordas2008} (see \cref{subsec:causal-understanding}), various \textbf{tool-use tasks} \parencite{Taylor2007, Taylor2012b, Taylor2012} and \textbf{floating-reward tasks} \parencite{Jelbert2014, Logan2014a, Mendes2007, Hanus2011, Sebastian-Enesco2022} (among several others).

Similarly, most RL agents are designed to be egocentric causal learners, with varying abilities to latch onto causal information provided by their own experience, which is ultimately shaped by the provided reward signal.
These abilities come for instance from strategies to boost \textbf{online learning} via particular training/optimisation techniques (e.g. uncertainty-based or curiosity-driven exploration) \parencite{Sutton2018, Ha2018, Schulman2017, Mnih2016, Igl2018, Haber2018a, Javed2020, Sontakke2021a}, or from methods to maximise the benefits of \textbf{online + off-policy learning} \parencite{Buesing2019, Silver2018a, Tomar2021, Zhang2021, Mendonca2021, Andrychowicz2017, Hafner2020}.

With the exception of some of the works cited in \cref{subsec:explicit-ci}, most of these approaches have not adopted a causal terminology.
Furthermore, the agents in question do not process feedback from the environment as causal information, e.g. by paying attention to key causal relationships with techniques from causal machine learning.
Nonetheless, we refer to them as egocentric causal learners because they have the ability to process (at least partially) the consequences of their actions.

\subsubsection{Social causal information}
\label{subsubsec:social-causal-info-examples}
As already mentioned in \cref{subsec:trajectories-causal-rl}, artificial agents can be designed to learn to solve a task through \textbf{imitation learning}, e.g. by relying on demonstrations of the expected behaviour for the given task.
The imitation learning pipeline can be implemented in various ways, tailored to the specific domain of application (for a recent review of the main techniques, see~\textcite{Hussein2017}).
In RL, the general idea is to allow a learning agent to have access to the experience of an expert, i.e. trajectories of optimal interactions for solving the task at hand, which are conveniently pre-processed in the same representational format of information in the replay buffer, so that they can guide the learning process towards a policy that achieves similar rewards \parencite{Ross2010, Ho2016, Li2017a}.
While broadly successful, imitation learning approaches do not necessarily entail the processing of social causal information in a comparable way to natural agents.
Imitation learning in and by itself does not in fact prevent a learning agent from simply exploiting correlations between state variables and actions present in the dataset of expert's demonstrations to learn an optimal policy for a certain task.
In the presence of distributional shift, which arises every time trajectory information used for training comes from a policy different from the one currently used by the learning agent, 
agents that learn by imitation, but without causal knowledge, are usually prone to causal confusion or misidentification (e.g. of what prompted the expert to act in certain ways) \parencite{DeHaan2019}.
If a correlation ceases to exist, performing the same action in response to a certain state could in fact turn out to be inappropriate in most cases.
This knowledge deficit has been highlighted and studied in depth by a few recent causal RL works, making a first important step towards artificial agents trained via imitation that are better equipped to deal with confounders and spurious correlations \parencite{DeHaan2019, Zhang2020a, Kumor2021, Lu2022b}, making them more ``aware'' of the causal structure of the problem under consideration.

The emergence of \textbf{offline RL} has marked another milestone in approaches to learning from imitation insofar as the emphasis is placed on the ability to learn from a dataset of previously recorded trajectories, potentially coming from other agents performing similar or different tasks \parencite{Levine2020, Wang2021a}. 
This represents a more challenging problem because during training the agent can no longer receive feedback from the environment, using its current policy to collect more trajectories through trial-and-error, as is typically done in imitation learning.
Optimal behaviour must be learned from a dataset that is not updated during training, and that inevitably will not provide a complete picture of the environment/task in which the agent will be deployed.
Techniques to ensure that a policy will perform well enough when deployed include conservative methods to bound the learned value functions (to avoid the risk of assigning high values to wrong states) \parencite{Kumar2020}, algorithms that take into consideration the agent's uncertainty about the identity of the test environment (enabling a kind of policy adaptation at test time) \parencite{Ghosh2022a}, and causal approaches to off-policy policy evaluation (see \textcite{Levine2020} and \textcite{Bannon2020} for comprehensive reviews).

To gain a better understanding of the extent to which current imitation learning approaches in RL are linked to causal cognition, it is instructive to consider a line of research in the animal cognition literature directed at investigating what kind of learning strategy is adopted in a social context by non-human primates, using \textbf{imitation vs.\ emulation tasks}.
The distinction between imitation and emulation revolves around the particular ``copying'' strategy used by the learning agent when observing the behaviour of a conspecific, i.e. either adhering to the demonstrator's actions (imitation) or focusing more on the action's results or outcomes (emulation) \parencite{Tomasello1990, Tomasello1998, Tomasello1997, Whiten2009, Zentall2022}.
The imitating agent will reproduce virtually the very same actions of the demonstrator whereas the emulating agent will try to reproduce the results of those actions, e.g. a rewarding outcome, using the same or different behavioural strategies, depending on context \parencite{Call2005, Tennie2006, Tennie2010, Renner2017, Renner2021, Pika2020, Horner2005}.
For instance, to collect a floating peanut from a water-filled tube (an example of a floating-reward task), one has to increase the water level in the cylindrical container; a higher water level is the key instrumental result (or precondition) required to solve the task.
In a social setting with expert demonstrators, a subject that overlooks that piece of information and learns to solve the task by copying all the particular actions of the expert conspecific (e.g. the ambulatory behaviour to collect the water) will fail at the task if those actions are no longer appropriate, or available, to produce the desired outcome (e.g. the water can be accessed only by climbing) \parencite{Tennie2010}.
Importantly, there is empirical evidence suggesting that adopting an emulative vs.\ imitative learning strategy can depend on the availability of causal information about the effects of certain actions, and their connection with the final, desired outcome.
For instance, in \textcite{Horner2005} chimpanzees witnessed a human demonstrator securing a reward from a puzzle-box using a tool.
When the box was opaque, hiding the relevant tool movements necessary to unlock the reward, at test time the subject reproduced all the actions seen in the demonstrations (learning by imitation).
Conversely, with a clear box the subjects learned to ignore the irrelevant actions, thereby solving the task more efficiently.
Thus, learning by emulation implies attending to goal/instrumental information (e.g. higher water levels in the tube) and being able to act upon it whether relevant behaviour has been demonstrated or not, i.e. attending to causal information pertaining to the causal states and/or variables that form a sort of precondition to reach a final outcome.
As such, this learning strategy affords efficiency and flexibility because the learning subject is free to explore and select the best course of action to reach an end goal.

An alternative RL framework for imitation learning can be found in \textbf{inverse RL}, where the aim is to design an agent capable of inferring the objective of the expert demonstrator, i.e. what reward function is shaping its behaviour, and of looking for an optimal policy based on that \parencite{Abbeel2004, Ziebart2008, Wulfmeier2016, Fu2018}.

\subsubsection{Natural causal information}
\label{subsubsec:natural-causal-info-examples}
Beyond the ability of learning from online interactions and social demonstrations, some (natural) agents also display a propensity for the acquisition of causal information from natural sources.
Natural causal information is precisely information about the existence (or absence) of certain causal relationships or structures that is gleaned from observing the occurrence of natural events (see \cref{subsec:sources-causal-info}).

Since the tree-branch thought experiment of \textcite{Tomasello1997}, it seems that the general consensus on observational (or impersonal) causal learning being an exclusively human ability has not shifted \parencite{Goddu2024}.
However, there is empirical evidence coming from some \textbf{observational causal tasks}, in which key causal relations can only be inferred from observations, suggesting that observational causal cognition in some non-human animals might be more developed than what it has been normally thought.
For instance, corvids have demonstrated an ability to take into consideration the potential effects of hidden causes, e.g. other agents or properties like the weight of an object, from observations alone \parencite{Taylor2012, Jelbert2019} while chimpanzees have been shown to be capable of inferring the presence of causal relationships from patterns of covariation (with a blicket-like experiment) \parencite{Volter2016} and using temporal cues \parencite{Tennie2019}.

Similarly, \textbf{``ghost''-condition tasks}, showing an apparatus in a final desired state and/or how a mechanism works (by pulling invisible strings), used to study emulation learning, also suggest that non-human primates exploit observational causal information to guide subsequent successful behaviour \parencite{Hopper2008, Hopper2010, Tennie2006}.

Thus, arguably, despite not reaching the performance achieved by humans, some non-human animals appear to have the ability to learn about the causal structure of not only systems they interact with, but also of systems they can merely observe.
This places them in a category beyond imitation (causal) learners, as they can make use of experience other than theirs, processing and capturing in causal terms events generated by external sources with a different body or physical configuration.

\subsection{Integration of Causal Information in Natural and Artificial Agents}
\label{subsec:integration-causal-info-rl}

\begin{figure}[ht!]
	\centering
	\scalebox{.68}{\begin{tikzpicture}
	\begin{pgfonlayer}{nodelayer}
		\node [style=big-circle-green] (0) at (-7.25, -2.5) {};
		\node [style=big-circle-orange] (1) at (7.25, -2.5) {};
		\node [style=big-circle-red] (2) at (0, 4) {};
		\node [style=none] (3) at (-5.5, 4) {\textcite{Tennie2019}};
		\node [style=none] (4) at (-5.5, 3) {\textcite{Volter2016}};
		\node [style=none] (5) at (-5.5, 1) {\textcite{Hopper2008}};
		\node [style=none] (6) at (-5.5, 2) {\textcite{Jelbert2019}};
		\node [style=none] (7) at (5.5, -1) {\textcite{Zhang2020a}};
		\node [style=none] (8) at (5.25, -2) {\textcite{Lu2022b}};
		\node [style=none] (9) at (5.75, 0) {\textcite{Kumor2021}};
		\node [style=none] (10) at (5.75, 1) {\textcite{DeHaan2019}};
		\node [style=none] (11) at (6.25, 5) {\textcite{Horner2005}};
		\node [style=none] (13) at (6.25, 4) {\textcite{Pika2020}};
		\node [style=none] (14) at (6.25, 6) {\textcite{Tennie2006, Tennie2010}};
		\node [style=none] (15) at (6.25, 3) {\textcite{Call2005}};
		\node [style=none] (16) at (0, 10) {\textcolor{red}{egocentric causal information}};
		\node [style=none] (17) at (12, -5) {\textcolor{Orange}{social causal information}};
		\node [style=none] (18) at (-12, -5) {\textcolor{ForestGreen}{natural causal information}};
		\node [style=none] (19) at (-16, 12.75) {\textcolor{RoyalBlue}{\textbf{egocentric + natural information}}};
		\node [style=none] (20) at (-8.75, 5) {};
		\node [style=none] (21) at (-16, 12) {};
		\node [style=none] (23) at (7.25, 7) {};
		\node [style=none] (24) at (10.75, 14.25) {};
		\node [style=none] (25) at (10.75, 15) {\textcolor{RoyalBlue}{\textbf{egocentric + social information}}};
		\node [style=none] (27) at (9, 0.75) {};
		\node [style=none] (28) at (15.25, 8.25) {};
		\node [style=none] (29) at (15.25, 9) {\textcolor{RoyalBlue}{\textbf{egocentric + social information}}};
		\node [style=none] (30) at (0, -14.5) {\textcolor{RoyalBlue}{\textbf{social + natural information}}};
		\node [style=none] (31) at (0, -7) {};
		\node [style=none] (32) at (0, -13.75) {};
		\node [style=none] (33) at (16.75, -12.75) {\textcolor{RoyalBlue}{\textbf{complete integration}}};
		\node [style=none] (34) at (0, -4.25) {};
		\node [style=none] (35) at (16.75, -12) {};
		\node [style=none] (36) at (15.25, 10.75) {{\huge \textcolor{TealBlue}{Artificial Agents}}};
		\node [style=none] (37) at (-16, 14.5) {{\huge \textcolor{Salmon}{Natural Agents}}};
		\node [style=none] (38) at (10.75, 16.75) {{\huge \textcolor{Salmon}{Natural Agents}}};
		\node [style=none] (39) at (0, -3.25) {\textcite{Taylor2012}};
		\node [style=none] (40) at (16.75, -14.75) {{\huge \textcolor{Salmon}{Natural Agents}}};
		\node [style=none] (41) at (0, 20) {{\Huge Integration}};
	\end{pgfonlayer}
	\begin{pgfonlayer}{edgelayer}
		\draw [style=thick-blue-dashed-arrow] (20.center) to (21.center);
		\draw [style=thick-blue-dashed-arrow] (23.center) to (24.center);
		\draw [style=thick-blue-dashed-arrow] (27.center) to (28.center);
		\draw [style=thick-blue-dashed-arrow] (31.center) to (32.center);
		\draw [style=thick-blue-dashed-arrow] (34.center) to (35.center);
	\end{pgfonlayer}
\end{tikzpicture}}
    \caption{\textbf{Integration of causal information.}}
    \label{fig:integration}
\end{figure}
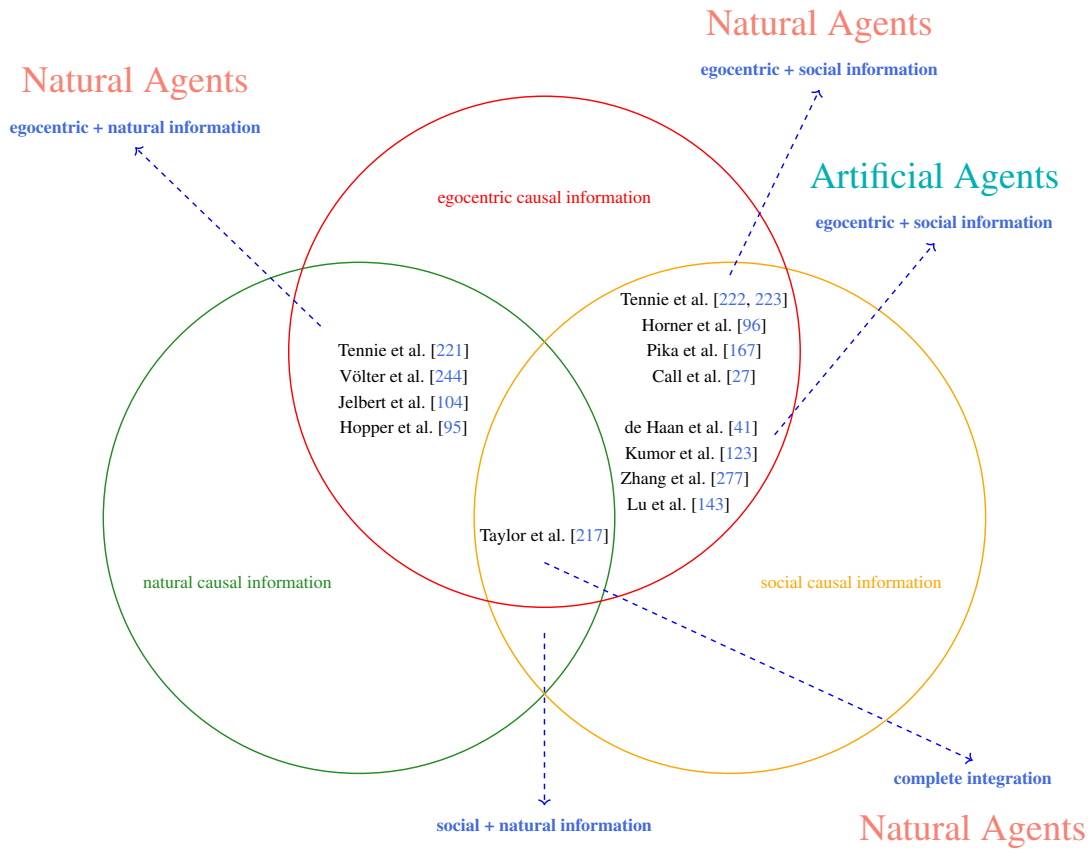

Following \cref{subsec:combining-sources-rl}, integration can be seen, from our perspective, as the process of incorporating and fusing different domains of causal information since, regardless of its source, any experience can be stored in a replay buffer (see also \cref{fig:integration}).
It is however important that different kinds of experience are integrated by concurrently taking into account their different roles, relevance for the given task, and/or potential weights based on the identification of key causal relationships.
The studies in animal cognition we examined so far can give us some clues, in the form of particular behavioural profiles, about what type of causal information integration happens in non-human animals.
However, it is important to keep in mind that behavioural traits are here used as a rudimentary proxy for cognitive operations of integration that remain still largely unknown.

As we saw in \cref{subsec:sources-causal-rl}, different animal cognition studies involve social causal information, e.g. demonstrations of a desired behaviour by an expert \parencite{Call2005, Horner2005, Tennie2006, Tennie2010, Pika2020}.
Despite the existence of negative results (e.g. \textcite{Renner2021}), these studies provide supporting evidence for the claim that non-human animals are capable of \textbf{egocentric + social integration}, see \cref{fig:integration}.
Within the same group, in RL, recent approaches have started to tease out the impact of certain causal relationships in imitation learning \parencite{Zhang2020a, Kumor2021, DeHaan2019, Lu2022b}.
These works represent a first step towards a better understanding of what integrating egocentric and social causal information might mean and especially entail, e.g. looking for invariances, confounders, direct causes of an expert actions.
Yet, it remains to be seen whether these approaches can be successfully extended and/or combined with methods to deal with high-dimensional, partially-observable scenarios (POMDPs) as their counterpart, natural causal learners, can integrate causal information starting from observations alone \parencite{Call2005, Horner2005, Tennie2006, Tennie2010, Pika2020}.

On the other hand, the fact that an animal's behaviour is influenced by the observation of certain causal relationships can be explained by invoking cognitive operations of integration that combine natural and egocentric causal information.
Observational learning experiments suggest that basic forms of \textbf{egocentric + natural integration} are present in non-human animals \parencite{Hopper2008, Volter2016, Jelbert2019, Tennie2019}.
For instance, inferring that the presence of a causal factor (e.g. the weight of an object) has an impact on what an action can accomplish (e.g. whether one can get food from a dispenser with a certain object or not), and behaving accordingly, can be considered as an example of this type of integration.

Conversely, there is almost no evidence for two other forms of integration in non-human animals, which are likely to require causal reasoning abilities about natural events and other agents that we know are present only in adult humans.
Integrating social and natural causal information, \textbf{social + natural integration}, might entail scenarios where an agent of choice, say Agent 1, perceives a causal relationship present in the natural world as exploitable by \emph{another} agent for its own purposes, say Agent 2 (for example, an expert in an imitation learning setup), but not for itself.
In other words, Agent 2 can combine pieces of causal information involving social and natural facts but at the same time Agent 1 is unable to relate the finding to its own circumstances, preventing a further integration with egocentric causal information.
What is hard to even conceive here is the very nature of this form of integration, by which effects on Agent 1 in some context are not directly and immediately measurable, because there is no behavioural evidence for this form of integration (this might come later when the agent is eventually able to capitalise on what it has observed and understood).

While in principle easier to detect for its benefits on a subject's egocentric perspective, integration of causal information from all sources, \textbf{complete integration}, can only be exposed by the most flexible and adaptive kind of behavioural responses and is thus hard to test in real setups.
Work in this direction can be found for instance in \textcite{Taylor2012}, where crows can explore an environment and retrieve food from a box (egocentric causal information).
The animal subjects learn through observation that the area near the box opening may be unsafe because it faces a curtain with a hole, from which a stick may appear and move, due to potentially natural forces in principle unknown to the subjects, to bother them (natural causal information).
Crucially, the crows can also witness that at times a human being enters the testing room and goes behind the curtain, suggesting that the movement of the stick may be caused by another agent (social causal information), posing thus a different kind of threat.
The question here is precisely whether the crows can ``reason'' about the opportunities of exploring the environment to get food based on the perceived risk of facing an aversive stimulus (the moving stick) in relation to the inferred presence/absence of a hidden agent. 
Since the crows were more hesitant to search the food box when an hidden agent was present, there is thus some evidence that the crows could integrate all those pieces of information (egocentric + social + observational) to tackle the task.
It is however debatable whether the crows could effectively recognise the humans agents as the causes responsible for the stick's motion.
In fact, it has been pointed out that non-human animals seem to lack a sophisticated understanding of causality in the psychological domain \parencite{Visalberghi1998}: in this case that the hidden agents could have the intention to move the stick when in the room.

\section{Discussion}
\label{sec:discussion}

Our comparative analysis so far has highlighted areas where animal cognition and causal reinforcement learning share some evident common ground in their otherwise different approaches to the study of causality in cognition and decision making.
In this final section we look in more detail at some of the opportunities offered by our unifying formal account of causal cognition, showcasing ways to make a more synergistic use of its strengths, and speculating on areas we believe will be of particular interest for future explorations.

\subsection{Computational interpretations of studies of natural agents}
\label{subsec:comparing-ai-nat-agents}

Modelling approaches in the literature of animal cognition are mainly concerned with capturing the cognitive and psychological processes of subjects exposed to tasks that are assumed to require an understanding of causal information \parencite{Dickinson2000, Blaisdell2006, Waldmann2008}.
However, simply providing evidence that subjects can learn complex causal structures from patterns of conditional (in)dependence shown to them and reason about interventions and counterfactuals in the world, leaves open the more fundamental question we asked in the introduction.
Specifically, \emph{how} does an adaptive agent interacting with and receiving feedback from an environment become sensitive to certain causal information and process it in ways that are conducive to reaching its goals?
To address this, we propose to use current models and algorithms developed in the fields of causal RL, and provide next some more specific examples.

\subsubsection{Measuring explicitness in natural agents}
In this work, we considered the question raised by \textcite{Starzak2021} of how to define and measure explicitness in animal cognition studies and proposed to think of it as disentanglement (see \cref{subsec:explicit-in-crl-agents}), roughly the degree of causal factorisation of a representation, to gain access to a relevant class of candidate metrics.
While a widely-accepted measure of disentanglement is still missing, different proposals have been put forward, providing thus multiple options that could be considered for the modelling and testing of explicitness in animal cognition \parencite{Higgins2018, Zhang2023, Wang2022}.

As a first step, we believe that a setup based on the AnimalAI Olympics framework \parencite{Crosby2019, Crosby2020a, Voudouris2023} could be used to introduce an experimental pipeline involving training artificial agents on the same class of causal tasks used in the animal cognition literature, to compare their performance with that of animals.
If the performance of two agents (artificial and natural) were to be comparable according to some appropriate success metric (e.g. solving a task, behavioural similarity), one could then measure the degree of disentanglement of the artificial agent's representation and use that to gain some understanding of a possible computational theory reflecting the natural agent's modelling capabilities.
Furthermore, we believe that this approach has the potential to become a standard benchmark for causal AI research to test if, and what kind of, causal representations can unlock the necessary skills to tackle problems of different complexity, showcasing causal learning abilities akin to the ones that appear to be present in natural agents \parencite{Crosby2020, Lake2017}.

\subsubsection{Zero-shot learning for high(er) explicitness}
A second example of the type of formalisation work afforded by causal RL implementations revolves around the fact that high explicitness is assigned to those animal agents that are capable of solving a task without much visual or sensorimotor feedback from the task at hand.
Despite contrasting empirical evidence \parencite{Ebel2019}, there are in fact suggestions that some non-human animals are capable of finding solutions to a certain problem ``in their head'' \parencite{Tennie2010, Sebastian-Enesco2022}, without the need for extended trial and error learning, which has been regarded as an indication of causal understanding \parencite{Limongelli1995}.

Similarly, in causal RL, zero-shot, offline, and continual learning describe models of agents capable of solving certain tasks with extremely limited training data, in virtue of having a causal representation of the environment usually supported by a process of planning that reuses and transfers previously acquired causal knowledge.
These agents are at the forefront of RL research, with problem definitions, formalisations, and benchmarks in constant evolution \parencite{Kirk2023, Touati2023, Khetarpal2022, Abel2023} and we believe they are likely to provide another ideal baseline for the development of computational theories of causal reasoning in animals.

\subsubsection{Emulation as inverse RL}

A large number of approaches in RL that make use of social causal information, i.e. causal information derived from observing the behaviour of other agents (see \cref{subsec:sources-causal-info}), can be said to implement a form of imitation learning.
Computationally, this form of learning can be described as behavioural cloning: trying to copy the policy of an expert agent as opposed to the outcomes of that policy \parencite{Ross2010}.
There is however a growing interest in different approaches, including offline and inverse RL, that we believe have the potential to provide a computational account of emulation as opposed to imitation learning (see \cref{subsubsec:social-causal-info-examples}), where the former describes natural agents that learn to reproduce outcomes of expert demonstrations, e.g.\ as in \textcite{Tennie2010}.

In a typical inverse RL setup for instance, the goal is to learn the reward map of another agent, say the expert.
This (non-unique) map can in general entail different optimal action policies, and thus by learning the expert's goal itself, an agent is not bound to only mimic the actions of the expert.
In turn, this could allow estimating intentions underlying other agents goal-directed behaviours based on a notion of causality in the psychological domain where intentions can be interpreted as causes of behaviour \parencite{Woodward2011, Visalberghi1998}.
A first step in this direction could include, for example, testing RL agents in realistic settings with imitation vs.\ emulation tasks inspired by the animal cognition literature, such as floating-reward tasks \parencite{Jelbert2014, Logan2014a, Mendes2007, Hanus2011, Sebastian-Enesco2022}, which could help to determine whether some degree of emulation is possible with current causal imitation learning techniques.

\subsection{Causal cognition inspired RL}
Looking then at figs.~\ref{fig:explicitness},~\ref{fig:sources} and~\ref{fig:integration} we can also identify potential areas where current causal RL frameworks can take inspiration from ideas developed in animal cognition.
In particular, we refer to areas where works and standard theories of causal understanding in animals have currently no counterpart in RL:\ causal insight in \cref{fig:explicitness}, learning from natural causal information in \cref{fig:sources} and complete integration in \cref{fig:integration}.

\subsubsection{Causal insight for causal RL}
In \cref{subsec:explicitness-causal-rep}, causal insight was described as the (1) capacity to produce adaptive responses as a result of reorganising one's causal knowledge, encoded in (2) highly explicit causal representations that lead to (3) an innovative solution to a problem.

The first defining element (1), involving flexible reuse of information and past knowledge in new tasks and/or environments is a long-standing challenge in AI research \parencite{Schmidhuber1996, Thrun1996, Caruana1997, Yu2020, Geisa2022}.
The second aspect (2), based on the relevance of causal representations for \emph{strong generalisation}, the ability to generalise out-of-distribution, encompasses transfer/meta/multi-task learning paradigms and has been noted in several causal machine learning works, with consensus that disentangled, structured, modular, causal representations can provide several benefits \parencite{Arjovsky2020a, Scholkopf2021, Scherrer2022, Goyal2022, Ahmed2020, Ahuja2021, Wenzel2022}.
The third feature (3), suggesting that solutions that innovative, in the sense that they give a new take on an existing problems or can be used for a new and unseen problem, is not fully captured or does not clearly emerge from the generalisation approaches just mentioned. 
In fact, systematic studies of, e.g. out-of-distribution learning as currently found in the literature are mostly limited to synthetic datasets and/or toy problems characterised by narrow task distributions, neglecting more realistic and ecological settings \parencite{Dittadi2021, Trauble2022, Ke2021, Goyal2020}.

More specifically, these sorts of investigations have not been carried out by means of evaluation methods and benchmarking that can take advantage of work found in the animal cognition literature. 
As suggested in some recent works \parencite{Crosby2020a, Crosby2020, Shanahan2020}, using training and testing protocols from animal cognition experiments has the potential to improve current architectures towards the goal of reproducing common sense abilities of different non-human animals (e.g.\ understanding of everyday physical notions like objecthood, containers, obstructions, and the related sets of affordances).
Since common sense abilities are deeply intertwined with an understanding of causality, studies of this kind could help to ascertain the extent to which causal RL can truly capture the manifestations of causal cognition in natural agents.

More generally, embracing a learning paradigm in which the central question is how an agent should gather and store key causal information for the purpose of subsequently extrapolating a strategy for tasks never seen before, seems \emph{essential} for causal insight to develop in artificial agents\footnote{Among the main achievements of deep RL in the last decade, there are for instance successes again top-level players in Go and other online games \parencite{Silver2016, Silver2018a, Silver2017, Vinyals2019} (see \textcite{Arulkumaran2017} for a review).
Some of the moves in these games have been described as creative and insightful (moves that no human player would have made at the time). 
However, note that in these cases the artificial agents in question are overly specialised in a single domain so we cannot talk of causal insight.}.
Further to the requirement of different training procedures and/or more computational power and data, tackling these kinds of questions will help to overcome persisting limitations, e.g. by introducing a more formal notion of causal insight connected with strong generalisation.

\subsubsection{Natural causal information in causal RL}

Offline reinforcement learning appears to be one of the closest available formalisations of a decision-making problem like the one posed by the tree-branch example \parencite{Tomasello1997}\footnote{We speculate that transformer-based architectures \parencite{Vaswani2017}, especially large language models and vision-language models for RL \parencite{Zhou2023, Open-XEmbodimentCollaboration2024, Wang2023, Gupta2022, Fan2022, Ghosh2024}, could provide an equally interesting proposal, whose in-depth discussion is however left for future work.} where a successful learning agent ought to be able to conduct a causal analysis of the natural scene it was part of, and then proceed to shake a fruit-bearing branch, just on the basis of having witnessed a fruit falling due to the wind shaking the tree.
One of the crucial aspect of this decision problem is the requirement of acting in an optimal manner immediately, based on the (natural) experience collected, i.e. without the ability to take advantage of further trial and error learning, which is precisely the setting of offline reinforcement learning.

However, while \textcite{Tomasello1997} highlights the importance of causal concepts to deal with this decision-making challenge, current approaches to deal with offline learning instead pursue strategies that try to mitigate the degree of distributional shift without any reference to causality.
In a nutshell, some methods introduce constraints on the policy being learned so as to minimise its divergence from the policy that collected the transitions stored in the replay buffer, while others use uncertainty measures to learn more conservative value functions in order to avoid catastrophic mistakes due to distributional shifts, see \textcite{Levine2020} for a review.

It is also important to note that, while the state transitions stored in the replay buffer could come from any policy, e.g. even those employed by agents with different bodily configurations or ``nature'' itself, to the best of our knowledge there are still no techniques to infer trajectory information from visual data in offline RL.
More specifically, we refer to the ability to process natural happenings, i.e. physical phenomena that don't involve any particular agent (the ``ghost'' conditions from the animal cognition literature, see \cref{subsubsec:natural-causal-info-examples}), through a causal lens. 
In practice, this would correspond to extracting state transitions tuples, \((\states_{t}, \states_{t+1}, \actions_{t}, \reward_{t})\), from high-dimensional visual input where a crucial step is to come up with a physicals interpretation of some external event as an impersonal action, producing an environmental state transition (the tree shaken by the wind).
Extending the offline framework in causal RL to include mechanisms to learn from nature has the potential to uncover further aspects of causality not yet understood and might spur a series of novel algorithmic solutions, getting closer to the design of an observational causal agent, where the ``ghost'' conditions from the animal cognition literature \parencite{Hopper2008, Hopper2010, Tennie2006} could be used as a test bed for this new generation of agents.

\subsubsection{Interventions in causal RL Agent}
The ability to reason about and perform interventions, which in a technical sense can be described as local perturbations of a system that set one or more causal factors/mechanisms to certain fixed values \parencite{Pearl2009a, Scholkopf2022}, has often been described as one of the hallmarks of causal reasoning agents \parencite{Blaisdell2006, Hagmayer2009, Gopnik2000, Gopnik2004, Pearl2018c} and has been used to draw a possible distinction between acting and intervening.
In this view, the former only implies an appreciation of the consequences of one's bodily movements (e.g. locomotion, reaching, grasping) leading to changes in perception and conditions for achieving certain goals.
The latter additionally involves an intentional modification of a certain aspect of the environment, exploiting an existing or induced causal relationship, to elicit a desired effect \parencite{Goddu2024, Leising2008}.
For instance, using a stick to make a fruit fall from a tree branch (intervening) is in many ways different from climbing the same tree to grab the fruit (acting), even though the outcome is ultimately the same (eating the fruit).
This suggests that not all actions of an agent qualify as interventions but all interventions are actions, whether realised or only imagined.
Importantly however, while an agent might be regarded as performing an intervention from the perspective of an external observer, it does not follow that the agent itself conceives of its actions as interventions.

By examining the animal cognition literature, at least two markers for interventional-aware agents can be identified.
One is the ability of certain agents to infer that an environmental causal path cannot be activated if an action is directed at producing an effect along that path.
If the tone is produced by the rat through a lever press, the cause (the light) that usually elicits the tone and makes the food available is likely absent and so will be the food \parencite{Leising2008, Blaisdell2006}.
The second one is the capacity to implement an innovative behavioural response simply following the exposure to certain observational patterns, i.e. what we called causal insight \parencite{Taylor2014, Jacobs2015}.
For instance, the primates solving floating-reward tasks seem to showcase a capacity to intervene, i.e. manipulating the environment (water level in the tube) to obtain a desired effect (reward is closer to the surface level) from observations alone, sometimes even without the need for visual feedback \parencite{Sebastian-Enesco2022}.

On the other hand, artificial RL agents can be hardly regarded as intervention-aware insofar as they are mostly engaged in acting in the sense described above (when the agent is embodied, either in a simulation or in the real world), and certainly they do not regard some of their actions as interventions.
It is yet unclear what the most promising approach to develop intervention-aware causal agents will require.
On a basic level, it might be crucial to revisit in causal terms some of the existing RL machinery.
Recent works \parencite{Pan2022, Pan2024} for instance show how the advantage function (see \cref{eqn:advantage}) can be interpreted as the causal effect of an action on the return (see \cref{eqn:return}) that displays typical properties of a causal representation (e.g. disentanglement), and discuss ways to estimate such a function directly from experience.
New architectures might also be needed, and among the necessary computational/algorithmic components of intervention-aware agents there might be an intervention model that converts low-level motor actions into interventions on causal factors \parencite{Goyal2022}.
Perhaps more drastically, causal counterparts of traditional machine learning (and/or RL) concepts will have to be developed and incorporated into the current theory and practice of RL, e.g. a causal/interventional version of the KL divergence \parencite{Wildberger2023, Simoes2024a}.

\section{Conclusion}

In this work, we introduced a unifying theoretical and computational framework for causal cognition, connecting different strands of research, from the classical literature on animal cognition to modern accounts of causal reinforcement learning in AI.
While traditionally presented in antithesis to associative learning, causal cognition has more recently been taken to span a wider spectrum of cognitive abilities all the way from associative learning to complex tasks such as tool use, emulation/imitation and observational learning.
A key aspect of this integrative view stems from recognising different levels of causal understanding as a key component of a framework for causal cognition.

At the same time, the lack of operational definitions for causal understanding, causal information and other similar notions has severely constrained the development of this field.
Recognising the crucial role played by the concept of causal learning and understanding in several influential works \parencite{Woodward2021, Woodward2012, Woodward2011, Woodward2007} (see also \textcite{Goddu2024} for a recent review), \cite{Starzak2021} outlined a conceptual space for causal cognition characterised by three dimensions: explicitness, sources and integration of causal information.
In the present work, we introduced a formal framework that provides more rigorous and clear underpinnings for those dimensions, and offers some precise coordinates to study various aspects of causal cognition.

More specifically, levels of explicitness were defined in terms of degrees of disentanglement \parencite{Bengio2013, Higgins2017, Higgins2018, Burgess2019, Zhang2023}, i.e. degrees of factorisation of a representation, and grouped in macro categories that we introduced under the names of weak and strong disentanglement, based on how much causal structure they can represent.
Sources of causal information were instead classified in terms of where this information originates from, egocentric (an agent's own experience), social (from other agents) and natural sources (the ``physics'' of an agent's environment) \parencite{Woodward2007, Woodward2011}.
Using the idea that in causal RL this information is usually stored on a replay buffer, or experience replay \parencite{Mnih2015}, we then operationalised integration as the ability to fuse pairs of different sources, or even all three of them at the same time.

We then used this framework to conduct a comparative study of causality as seen through the lenses of animal cognition and reinforcement learning, with the former exploring areas that could inspire the latter, and the latter showcasing concrete proposals for computational and process theories of causal cognition missing in the former.
Future work will aim to turn some of our suggestions into practical proposals with applications to both animal cognition (new computational models for causal learning) and reinforcement learning (algorithms implementing more powerful forms of causality) with the goal of showcasing the advancements that we believe can only be derived once an integrated, unifying approach based on the work we presented here can be thoroughly implemented in practice.

\section*{Acknowledgements}

Part of this work involves research conducted by F.T.\ during his PhD, supported by a Leverhulme Doctoral Scholarship under the doctoral programme ``From sensation and perception to awareness'' at the University of Sussex (\url{https://www.sussex.ac.uk/sensation/}).







\printbibliography

\end{document}